\RequirePackage{fix-cm}
\documentclass[twocolumn,final]{svjour3}          
\smartqed  

\usepackage[numbers,sort&compress]{natbib}
\usepackage{rotating}
\usepackage{amssymb}
\usepackage{pifont}
\usepackage{amsmath}
\usepackage{multirow} 
\usepackage{multicol} 
\usepackage{hhline} 
\usepackage{array}
\usepackage{booktabs}
\usepackage{times}
\usepackage{epsfig}
\usepackage{graphicx}
\usepackage{amsmath}
\usepackage{amssymb}
\DeclareGraphicsExtensions{.pdf,.jpeg,.png,.eps}
\usepackage{epstopdf}
\usepackage{color}
\usepackage{algorithm}
\usepackage{algorithmic}
\usepackage{float}
\usepackage{xspace}
\usepackage[flushleft]{threeparttable}
\usepackage{tablefootnote}
\DeclareMathAlphabet{\mathpzc}{OT1}{pzc}{m}{it}
\usepackage{url}
\usepackage{cases}
\usepackage[colorlinks=false]{hyperref}
\hypersetup{colorlinks,linkcolor={blue},citecolor={blue},urlcolor={blue}}  
\usepackage{soul}
\usepackage[subrefformat=parens,labelformat=parens]{subfig}

\usepackage{pbox}
\usepackage{caption}
\usepackage{comment,enumitem}
\usepackage{float}
\usepackage[misc,geometry]{ifsym}
\newcommand{\cmark}{\ding{51}}%
\journalname{}
\begin{document}

\title{Text Detection and Recognition in the Wild: A Review} 

\author{Zobeir Raisi$^\textrm{1}$ \and
Mohamed A. Naiel$^\textrm{1}$ \and
Paul Fieguth$^\textrm{1}$ \and \\
Steven Wardell$^\textrm{2}$ \and
John Zelek$^\textrm{1}$* \thanks{* Corresponding author.}
}


\institute{Z. Raisi \at
              \email{zraisi@uwaterloo.ca}           
           \and
            M. A. Naiel \at
              \email{mohamed.naiel@uwaterloo.ca}
         \and
            \space J. Zelek \at
              \email{jzelek@uwaterloo.ca}
           \and
            P. Fieguth \at
             \email{ pfieguth@uwaterloo.ca}
         \and
          S. Wardell \at
             \email{swardell@atsautomation.com}
             \and
             $^\textrm{1}$ {Vision and Image Processing Lab and the Department of Systems Design Engineering, University of Waterloo, ON, N2L 3G1, Canada} \\
             $^\textrm{2}$ {ATS Automation Tooling Systems Inc., Cambridge, ON, N3H 4R7, Canada}
}

\date{Received: date / Accepted: date}

\maketitle

\begin{abstract}
%
Detection and recognition of text in natural images are two main problems in the field of computer vision that have a wide variety of applications in analysis of sports videos, autonomous driving, industrial automation, to name a few. 
They face common challenging problems that are factors in how text is represented and affected by several environmental conditions.
The current state-of-the-art scene text detection and/or recognition methods have exploited the witnessed advancement in deep learning architectures and reported a superior accuracy on benchmark datasets when tackling multi-resolution and multi-oriented text. 
However, there are still several remaining challenges affecting text in the wild images that cause existing methods to underperform due to there models are not able to generalize to unseen data and the insufficient labeled data.
Thus, unlike previous surveys in this field, the objectives of this survey are as follows: first, offering the reader not only a review on the recent advancement in scene text detection and recognition, but also presenting the results of conducting extensive experiments using a unified evaluation framework that assesses pre-trained models of the selected methods on challenging cases, and applies the same evaluation criteria on these techniques.
Second, identifying several existing challenges for detecting or recognizing text in the wild images, namely, in-plane-rotation, multi-oriented and multi-resolution text, perspective distortion, illumination reflection, partial occlusion, complex fonts, and special characters. 
Finally, the paper also presents insight into the potential research directions in this field to address some of the mentioned challenges that are still encountering scene text detection and recognition techniques.
\end{abstract}
\section{Introduction}
\label{Introduction}
\begin{figure}[t]
	\centering
	\includegraphics[width = \linewidth]{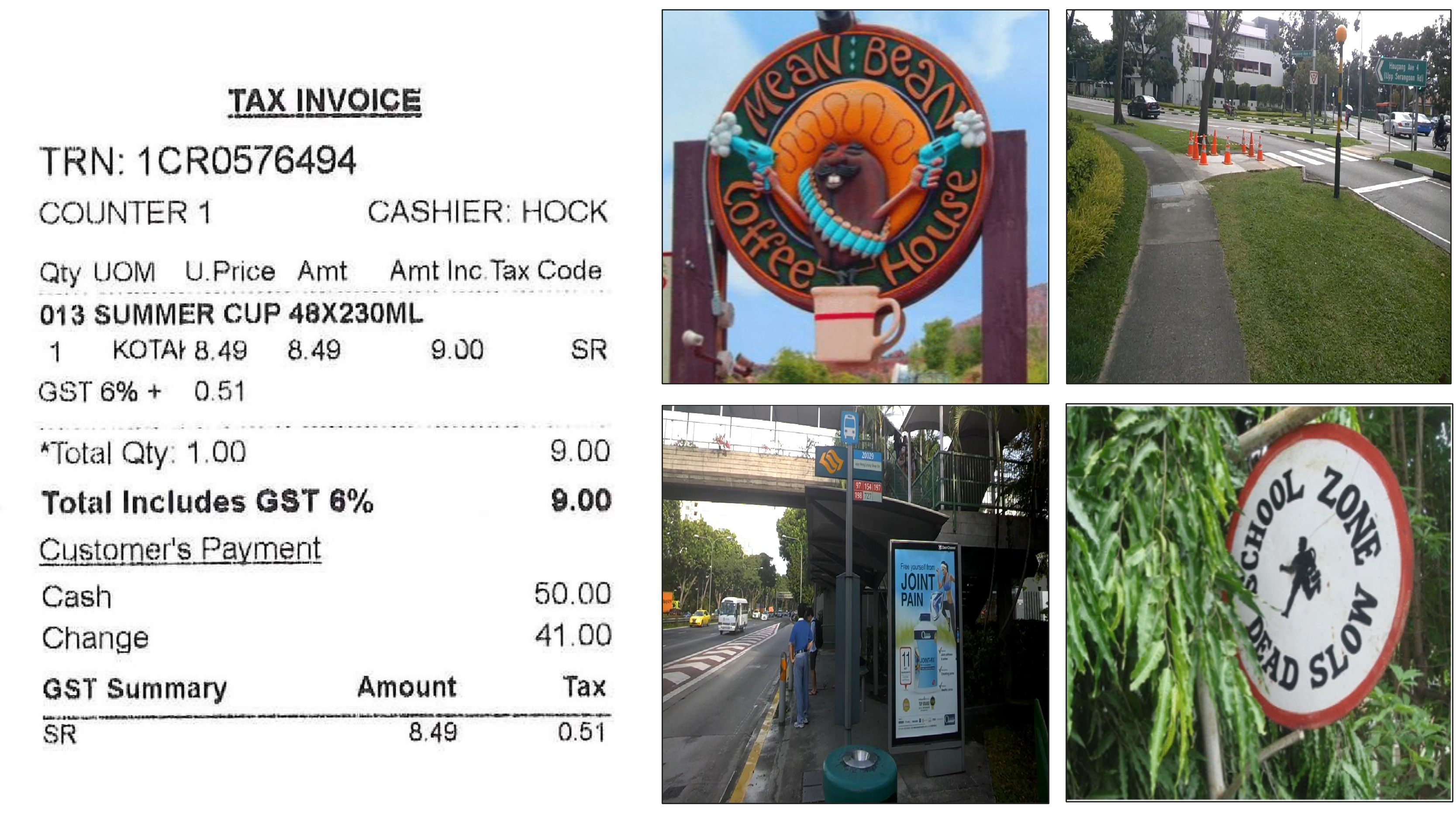}
	\caption{Examples for two main types of text in images: text in a printed document (left column) and text captured in the wild (right column), where sample images are from the public datasets in \cite{huang2019icdar2019,karatzas2015icdar,ch2017total}.}
	\label{fig:typeOfOCR}
\end{figure}

Text is a vital tool for communications and plays an important role in our lives. 
It can be embedded into documents or scenes as a mean of conveying information \cite{survey2015, survey2018, survey2019}. 
Identifying text can be considered as a main building block for a variety of computer vision-based applications, 
such as robotics \cite{case2011autonomous, kostavelis2015semantic}, 
industrial automation \cite{ham1995recognition}, 
image search \cite{chandrasekhar2011stanford,tsai2011mobile}, 
instant translation \cite{ma2000mobile,cheung2008system}, 
automotive assistance \cite{wu2005detection} and 
analysis of sports videos \cite{Messelodi2013}. 
Generally, the area of text identification can be categorized into two main categories: 
identifying text of \textit{scanned printed documents} and text captured for daily scenes (e.g., images with text of more complex shapes captured on urban, rural, highway, indoor / outdoor of buildings, and subject to various geometric distortions, illumination and environmental conditions), where the latter is called \textit{text in the wild} or \textit{scene text}. 
Figure \ref{fig:typeOfOCR} illustrates examples for these two types of text-images.
For identifying text of scanned printed documents, Optical Character Recognition (OCR) methods have been widely used \cite{somerville1991method,chen2003text,survey2015,book_2017}, which achieved superior performances for reading printed documents with satisfactory resolution; 
However, these traditional OCR methods face many complex challenges when used to detect and recognize text in images captured in the wild that cause them to fail in most of the cases \cite{survey2015,survey2018, shi2018aster}.

The challenges of detecting and/or recognizing text in images captured in the wild can be categorized as follows:
\begin{itemize}
\item \textbf{Text diversity:} text can exist in a wide variety of colors, fonts, orientations and languages.
\item  \textbf{Scene complexity:} scene elements of similar appearance to text, such as signs, bricks and symbols. 
\item  \textbf{Distortion factors:} the effect of image distortion due to several contributing factors such as motion blurriness, insufficient camera resolution, capturing angle and partial occlusion \cite{survey2015,survey2018,survey2019}.
\end{itemize} 

In the literature, many techniques have been proposed to address the challenges of scene text detection and/or recognition. 
These schemes can be categorized into \textit{classical machine learning-based}, as in \cite{kim2003texture,chen2004detecting,hanif2009text,wang2011end,lee2011adaboost,bissacco2013photoocr,wang2003character,mancas2006spatial,mancas2007color,song2008novel,kim2008new,de2009character, pan2009text}, and \textit{deep learning-based}, as in \cite{huang2014robust,zhang2015symmetry,jaderberg2016reading,jaderberg2014deep,tian2016detecting,zhou2017east,liao2017textboxes,liao2018textboxes++,ma2018arbitrary,zhang2016multi,yao2016scene,wu2017_self,long2018textsnake,deng2018pixellink,lyu2018multi,qin2019algorithm,baek2019craft,pmtd_liu_2019,lyu2018mask,liu2018fots,he2018end,busta2017deep,shi2016end,shi2016robust,STARNet_2016,rosetta2018,baek2019STR}, approaches. 
A classical approach is often based on combining a feature extraction technique with a machine learning model to detect or recognize text in scene images \cite{wang2011end,matas2004robust,epshtein2010detecting}. 
Although some of these methods \cite{matas2004robust,epshtein2010detecting} achieved good performance on detecting or recognizing horizontal text \cite{survey2015,survey2019}, these methods typically fail to handle images that contains multi-oriented or curved text \cite{survey2018,survey2019}. 
On the other hand, for text captured under adverse situations deep-learning based methods have shown effectiveness in detecting text \cite{jaderberg2014deep,tian2016detecting,zhang2016multi,yao2016scene,liao2017textboxes,zhou2017east,shi2017detecting,liao2018textboxes++,he2018end, ma2018arbitrary,long2018textsnake,deng2018pixellink,pmtd_liu_2019,baek2019craft}, recognizing text \cite{jaderberg2014synthetic, shi2016end, jaderberg2016reading,shi2016robust, lee2016recursive,  STARNet_2016, wang2017gated, yang2017learning,cheng2017fan, liu2018char-net,  cheng2018aon, bai2018ep, rosetta2018, liu2018synthetically, xie2019ACE, zhan2019esir, Wang2019ASA,baek2019STR}, and end-to-end detection and recognition of text  \cite{busta2017deep,lyu2018mask,liu2018fots,he2018end}.

Earlier surveys on scene text detection and recognition methods \cite{survey2015, survey_yin2016} have performed a comprehensive review on classical methods that mostly introduced before the deep-learning era. 
%
%
While more recent surveys \cite{survey2018,survey2019,survey_2019_liu} have focused more on the advancement occurred in scene text detection and recognition schemes in the deep learning era.
Although these two groups cover an overview of the progress made in both the classical and deep-learning based methods,
they concentrated mainly on summarizing and comparing the results reported in the witnessed papers.  

This paper aims to address the gap in the literature by not only reviewing the recent advances in scene text detection and recognition, with a focus on the deep learning-based methods, but also using the the same evaluation methodology to assess the performance of some of the best state-of-the-art methods on challenging benchmark datasets. 
Further, this paper studies the shortcomings of the existing techniques through conducting an extensive set of experiments followed by results analysis and discussions. 
Finally, the paper proposes potential future research directions and best practices, which potentially would lead to designing better models that are able to handle scene text detection and recognition under adverse situations.


\section{Literature Review}
\label{sec:literatureReview}



\begin{figure}[t]
    \centering
    \includegraphics[width=\linewidth]{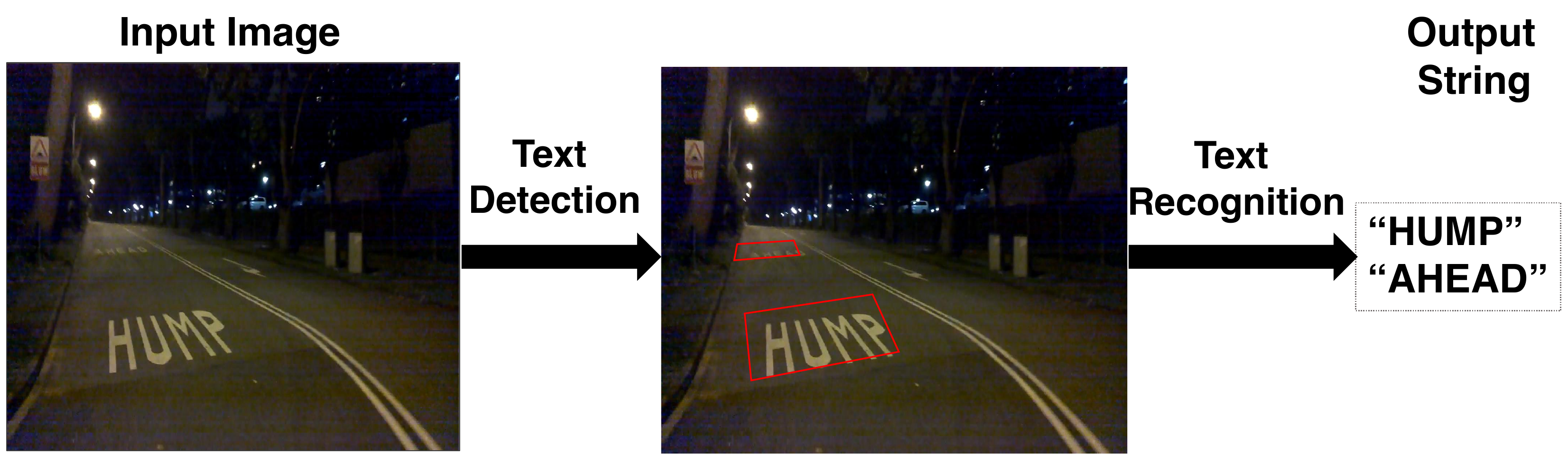}
    \caption{General schematic diagram of scene text detection and recognition, where sample image is from the public dataset in \cite{veit2016coco}.}
    \label{fig:Three_task_OCR}
\end{figure}

During the past decade, researcher proposed many techniques  for reading text in images captured in the wild \cite{epshtein2010detecting, wang2010word, liao2017textboxes, shi2016robust,jaderberg2016reading}. 
These techniques, first localize text regions in images by predicting bounding boxes for every possible text region, and then recognize the contents of every detected region.
Thus, the process of interpreting text from images can be divided into two subsequent tasks, namely, \textit{text detection} and \textit{text recognition} tasks.  
As shown in Fig. \ref{fig:Three_task_OCR}, text detection aims detecting or localizing text regions from  images. On the other hand, text recognition task only focuses on the process of converting the detected text regions into computer-readable and editable characters, words, or text-line. In this section, the conventional and recent algorithms for text detection and recognition will be discussed.

\subsection{Text Detection}
\begin{figure*}
    \centering
    \includegraphics[width=\linewidth]{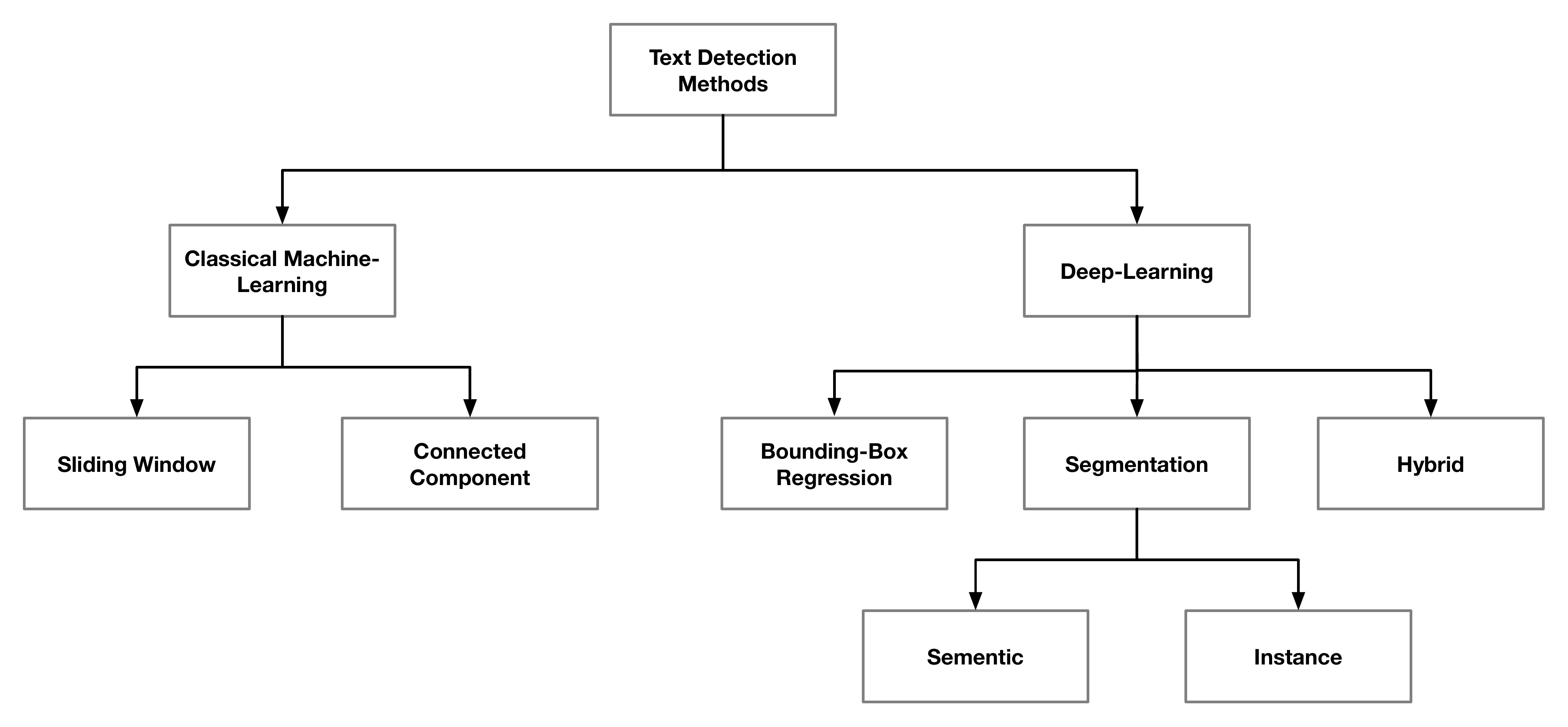}
    \caption{General taxonomy for the various text detection approaches.}
    \label{fig:taxnomaydetection}
\end{figure*}

 As illustrated in Figure \ref{fig:taxnomaydetection}, scene text detection methods can be categorized into \textit{classical machine learning-based} \cite{chen2004detecting,neumann2010method,epshtein2010detecting, wang2011end,lee2011adaboost,yi2011_cc,wang2012end,mishra2012top,yao2012detecting,bissacco2013photoocr,huang2014robust,yin2014robust} and \textit{deep learning-based} \cite{jaderberg2014deep,tian2016detecting,zhang2016multi,yao2016scene,liao2017textboxes,zhou2017east,shi2017detecting,liao2018textboxes++,he2018end, ma2018arbitrary,long2018textsnake,deng2018pixellink,pmtd_liu_2019} methods. In this section, we will review the methods related to each of these categories. 

{}

\subsubsection{Classical Machine Learning-based Methods}
This section summarizes the traditional methods used for scene text detection, which can be categorized into two main approaches, namely, 
\textit{sliding-window} and \textit{connected-component} based approaches.  

In \textit{sliding window-based methods}, such as  \cite{kim2003texture,chen2004detecting,hanif2009text,wang2011end,lee2011adaboost,bissacco2013photoocr}, a given test image is used to construct an image pyramid to be scanned over all the possible text locations and scales by using a sliding window of certain size. 
Then, a certain type of image features (such as mean difference and standard deviation as in \cite{hanif2009text}, histogram of oriented gradients (HOG) \cite{dalal_hog} as in \cite{pan2010hybrid,wang2011end,tian2015text} and edge regions as in \cite{lee2011adaboost}) are obtained from each window and classified by a classical classifier (such as random ferns \cite{bosch2007_random_Ferns} as in \cite{wang2011end}, and adaptive boosting (AdaBoost) \cite{AdaBoost_schapire_1999} with multiple weak classifiers, for instance, decision trees \cite{lee2011adaboost}, log-likelihood  \cite{chen2004detecting}, likelihood ratio test \cite{hanif2009text}) to detect text in each window. 
%
For example, in an early work by Chen and Yuille \cite{chen2004detecting},
intensity histograms, intensity gradients and gradient directions features were obtained at each sliding window location within a test image. 
Next, several weak log-likelihood classifiers, trained on text represented by using the same type of features, were used  to construct a strong classifier using the AdaBoost framework for text detection.
In \cite{wang2011end}, HOG features were extracted at every sliding window location and a Random Fern  classifier \cite{ozuysal2007fast} was used for multi-scale character detection, where the non-maximal suppression (NMS) in \cite{felzenszwalb2009object} was performed to detect each character separately.
However, these methods \cite{chen2004detecting,lee2011adaboost,wang2011end} are only applicable to detect horizontal text and have a low detection performance on scene images, which have arbitrary orientation of text \cite{survey_zhu2016scene}.
%
 %
%

\textit{Connected-component based methods} aim to extract image regions of similar properties (such as color \cite{wang2003character,mancas2006spatial,mancas2007color,song2008novel,kim2008new}, texture \cite{ye2005fast}, boundary \cite{li2008adaptive,shivakumara2009gradient,neumann2012real,cho2016canny}, and corner points \cite{zhao2010text}) to create candidate components that can be categorized into text or non-text class by using a traditional classifier (such as support vector machine (SVM)  \cite{neumann2010method}, Random Forest \cite{yao2012detecting} and nearest-neighbor  \cite{neumann2013scene}). 
These methods detect characters of a given image and then combine the extracted characters into a word \cite{epshtein2010detecting,neumann2010method,yi2011_cc} or a text-line \cite{chen2011robust}.
Unlike sliding-window based methods, connected-component based methods are more efficient and robust, and they offer usually a lower false positive rate, which is crucial in scene text detection \cite{survey_2019_liu}. 

Maximally stable extremal regions (MSER) \cite{matas2004robust} and stroke width transform (SWT) \cite{epshtein2010detecting} are the two main representative connected-component based methods that constitute the basis of many subsequent text detection works \cite{neumann2010method,chen2011robust,yi2011_cc, yao2012detecting,huang2013text,huang2014robust,yin2014robust,busta2015fastext,cho2016canny,survey_yin2016}.
%
However, the mentioned classical methods aim to detect individual characters or components that may easily cause discarding regions with ambiguous characters or generate a large number of false detection that reduce their detection performance \cite{he2017single}.
Furthermore, they require multiple complicated sequential steps, which lead to easily propagating errors to later steps. 
In addition, these methods might fail in some difficult situations, such as detecting text under non-uniform illumination, and text with multiple connected characters \cite{zhang_2016_CC}.
%

\subsubsection{Deep Learning-based Methods}

\begin{table*}[]
    \centering
    \caption{Deep learning text detection methods, where W: Word, T: Text-line, C: Character, D: Detection, R: Recognition, BB: Bounding Box Regression-based, SB: Segmentation-based, ST: Synthetic Text, IC13: ICDAR13, IC15: ICDAR15,  M500: MSRA-TD500, IC17: ICDAR17MLT, TOT: TotalText, CTW:CTW-1500 and the rest of the abbreviations used in this table are presented in Table \ref{tab:attribute}.}
    \scalebox{0.9}{
     \setlength{\tabcolsep}{2 pt}
    \begin{tabular}{lcccccccccccclr}
    \toprule
    \multirow{2}{*}{Method} & \multirow{2}{*}{Year} & \multicolumn{3}{c}{IF} &\multicolumn{2}{c}{Neural Network}& \multirow{2}{*}{\parbox{1.1cm}{\centering Detection Target}}&\multicolumn{2}{c}{Challenges} & \multirow{2}{*}{Task} & \multirow{2}{*}{\parbox{0.3cm}{\centering Code}}&\multirow{2}{*}{\parbox{0.9cm}{\centering Model Name}}&\multicolumn{2}{c}{Training Datasets} \\ \cline{3-5} \cline{6-7} \cline{9-10} \cline{14-15}
       & & BB&SB&Hy & Architecture  & Backbone &  &Quad&Curved & & & &First-Stage&Fine-Tune\\ \toprule
    Jaderberg\textit{et al.}\cite{jaderberg2014deep}   &2014  &  --   &  --     & & CNN               & --                     &W      &  --   &  --               &D,R    & --      &DSOL             & MJSynth                 & --  \\    
    Huang    \textit{et al.} \cite{huang2014robust}    &2014  &  --   &  --     & & CNN               & --                     &W      &  --   &  --               &D      & --      &RSTD             & --                      & IC11 or IC15  \\    
    Tian    \textit{et al.} \cite{tian2016detecting}   &2016  &\cmark &  --     & & Faster R-CNN      &VGG-16                  &T,W    &  --   &  --               &D      &\cmark   &CTPN             & PD                      & IC13  \\    
    Zhang \textit{et al.} \cite{zhang2016multi}        &2016  &  --   &\cmark   & & FCN               &VGG-16                  &W      &\cmark &  --               &D      &\cmark   &MOTD             & --                      & IC13, IC15 or M500  \\    
    Yao \textit{et al.} \cite{yao2016scene}            &2016  &  --   &\cmark   & & FCN               &VGG-16                  &W      &\cmark &  --               &D      &\cmark   &STDH             & --                      & IC13, IC15 or M500  \\    
    Shi \textit{et al.} \cite{shi2017detecting}        &2017  &\cmark &  --     & &SSD                & VGG-16                 &C,W    &\cmark &  --               &D      &\cmark   &SegLink          & ST                      & IC13, IC15 or M500  \\    
    He \textit{et al.}  \cite{he2017single}.           &2017  &  --   &\cmark   & &SSD                &VGG-16                  &W      &\cmark &  --               &D      &\cmark   &SSTD             & --                      & IC13 or IC15  \\    
    Hu \textit{et al.} \cite{hu2017wordsup}            &2017  &  --   &\cmark   & & FCN               & VGG-16                 &C      &\cmark &  --               &D      &  --     &Wordsup          & ST                      & IC15 or COCO  \\    
    Zhou \textit{et al.} \cite{zhou2017east}           &2017  &\cmark &  --     & & FCN               &VGG-16                  &W,T    &\cmark &  --               &D      &\cmark   &EAST             & --                      & IC15*, COCO or M500  \\    
    He \textit{et al.}  \cite{he2017deep}              &2017  &\cmark &  --     & &DenseBox           &--                      &W,T    &\cmark &  --               &D      &  --     &DDR              & --                      & IC13, IC15 \& PD  \\    
    Ma \textit{et al.}  \cite{ma2018arbitrary}         &2018  &\cmark &  --     & &Faster R-CNN       &VGG-16                  &W      &\cmark &  --               &D      &\cmark   &RRPN             & M500                    & IC13 or IC15  \\    
    Jiang \textit{et al.} \cite{jiang2017r2cnn}        &2018  &\cmark &  --     & &Faster R-CNN       &VGG-16                  &W      &\cmark &  --               &D      &\cmark   &R2CNN            & IC15 \& PD              & --  \\    
    Long \textit{et al.} \cite{long2018textsnake}      &2018  &  --   &\cmark   & &U-Net              & VGG-16                 &W      &\cmark &\cmark             &D      &\cmark   &TextSnake        & ST                      & IC15, M500, TOT or CTW  \\    
    Liao \textit{et al.} \cite{liao2018textboxes++}    &2018  &\cmark &  --     & &SSD                &VGG-16                  &W      &\cmark &  --               &D,R    &\cmark   &TextBoxes++      & ST                      & IC15  \\    
    He \textit{et al.} \cite{he2018end}                &2018  &  --   &\cmark   & &FCN                &PVA                     &C,W    &\cmark &  --               &D,R    &\cmark   &E2ET             & ST                      & IC13 or IC15  \\    
    Lyu \textit{et al.} \cite{lyu2018mask}             &2018  &  --   &\cmark   & &Mask-RCNN          &ResNet-50               &W      &\cmark &  --               &D,R    &\cmark   &MTSpotter        & ST                      & IC13, IC15 or TOT  \\    
    Liao \textit{et al.} \cite{liao2018rotation}       &2018  &\cmark &  --     & &SSD                &VGG-16                  &W      &\cmark &  --               &D      &\cmark   &RRD              & ST                      & IC13, IC15, COCO or M500  \\    
    Lyu \textit{et al.} \cite{lyu2018}                 &2018  &  --   &\cmark   & &FCN                &VGG-16                  &W      &\cmark &  --               &D      &\cmark   &MOSTD            & ST                      & IC13 or IC15 \\   
    Deng \textit{et al.}*\cite{deng2018pixellink}      &2018  &\cmark &  --     & &FCN                &VGG-16                  &W      &\cmark &  --               &D      &\cmark   &Pixellink*        & IC15                    & IC13, IC15* or M500  \\   
    Liu \textit{et al.}\cite{liu2018fots}             &2018  &\cmark &  --     & &CNN                &ResNet-50               &W      &\cmark &  --               &D,R    &\cmark   &FOTS              & ST                      & IC13, IC15 or IC17  \\   
    Baek \textit{et al.}*\cite{baek2019craft}          &2019  &  --   &\cmark   & &U-Net              &VGG-16                  &C,W,T  &\cmark &\cmark             & D     &\cmark   &CRAFT*            & ST                      & IC13, IC15* or IC17  \\   
    Wang \textit{et al.}*\cite{wang2019_PAN}           &2019  &  --   &\cmark   & &FPEM+FFM          &ResNet-18                &W      &\cmark &\cmark             & D     &\cmark   &PAN*              & ST                      & IC15*, M500, TOT or CTW  \\   
    Liu \textit{et al.}*\cite{pmtd_liu_2019}           &2019  &-- & --  &\cmark &Mask-RCNN          &ResNet-50                 &W      &\cmark &\cmark             & D     &\cmark   &PMTD*             & IC17                    & IC13 or IC15*            \\   
    Xu \textit{et al.} \cite{xu_2019_textfield}        &2019  &  --   &\cmark   & &FCN                &VGG-16                  &W      &\cmark &\cmark             & D     &\cmark   &Textfield        & ST                      & IC15, M500, TOT or CTW   \\   
    Liu \textit{et al.}* \cite{liu2019_BOX}             &2019  &  --   &\cmark   & &Mask-RCNN          &ResNet-101              &W      &\cmark &\cmark             & D     &\cmark   &MB*              & ST                      & IC15*, IC17 or M500       \\   
    Wang  \textit{et al.}* \cite{PSENet_wang2019}      &2019  &  --   &\cmark   & &FPN                &ResNet                  &W      &\cmark &\cmark             & D     &\cmark   &PSENet*           & IC17                    & IC13 or IC15*                 \\   

    \bottomrule
    \end{tabular}    }
    \begin{tablenotes}
      \small
      \item Note: * The method has been considered for evaluation in this paper, where all the selected methods have been trained 
      on ICDAR15 (IC15) dataset to compare there results in a unified framework.
    \end{tablenotes}
    \label{tab:compare_deepMethods}
\end{table*}

\begin{table}[]
    \centering
    \caption{Supplementary table of abbreviations.}
    \scalebox{1}{
    \setlength{\tabcolsep}{2 pt}
    \begin{tabular}{l|p{6.5cm}}
    \toprule
         Attribution & Description  \\
    \toprule
         FCN  \cite{long2015fully} & Fully Convolutional Neural Network \\
         FPN \cite{Lin_2017_FPN} & Feature pyramid networks\\
         PVA-Net \cite{kim2016pvanet} & Deep but Lightweight Neural Networks for Real-time Object Detection\\
         RPN \cite{fasterrcnn_ren2015} &  Region Proposal Network \\         
         SSD \cite{liu2016ssd} & Single shot detector\\
         U-Net \cite{Unet_2015} & Convolutional Networks developed for Biomedical Image Segmentation\\
         FPEM \cite{wang2019_PAN} & Feature Pyramid Enhancement Module \\
         FFM \cite{wang2019_PAN} & Feature Fusion Module \\
    \bottomrule
    \end{tabular}}
    \label{tab:attribute}
\end{table}

The emergence of deep learning \cite{krizhevsky2012imagenet} has changed the way researchers approached the text detection task and has enlarged the scope of research in this field by far.
Since deep learning-based techniques have many advantageous over the classical machine learning-based ones (such as faster and simpler pipeline \cite{zhou2017oriented}, detecting text of various aspect ratios \cite{shi2017detecting}, and offering the ability to be trained better on synthetic data \cite{jaderberg2016reading}) they have been widely used \cite{zhang2016multi,he2017deep,ma2018arbitrary}. 
In this section, we present a review on the recent advancement in deep learning-based text detection methods; 
Table \ref{tab:compare_deepMethods} summarizes a comparison among some of the current state-of-the-art techniques in this field.

Earlier deep learning-based text detection methods \cite{jaderberg2014deep,huang2014robust,zhang2015symmetry,jaderberg2016reading} usually consist of multiple stages. 
For instance, Jaderberg \textit{et al.} \cite{jaderberg2014deep} extended the architecture of a convolutional neural network (CNN) to train a supervised learning model in order to produce text saliency map, then combined bounding boxes at multiple scales by undergoing filtering and NMS. 
Huang \textit{et al.} \cite{huang2014robust} utilized both conventional connected component-based approach and deep learning for improving the precision of the final text detector. 
In this technique, the classical MSER \cite{matas2004robust} high contrast regions detector was employed on the input image to seek character candidates; then, a CNN classifier was utilized to filter-out non-text candidates by generating a confidence map that was later used for obtaining the detection results.
Later in \cite{jaderberg2016reading} the aggregate channel feature (ACF) detector \cite{dollar2014fast} was used to generate text candidates, and then a CNN was utilized for bounding box regression to reduce the false-positive candidates.
However, these earlier deep learning methods \cite{jaderberg2014deep,huang2014robust,zhang2015symmetry} aim mainly to detect characters; thus, their performance may decline when characters present within a complicated background, i.e., when elements of the background are similar in appearance to characters, or characters affected by geometric variations \cite{zhang2016multi}.

Recent deep learning-based text detection methods \cite{tian2016detecting,zhou2017east,liao2017textboxes,shi2017detecting,liao2018textboxes++,he2018end, ma2018arbitrary} inspired by object detection pipelines \cite{fasterrcnn_ren2015,YOLO_2016,liu2016ssd,long2015fully,Maskrcnn_He2017}
 can be categorized into \textit{bounding-box regression based}, \textit{segmentation-based} and  \textit{hybrid} approaches as illustrated in Figure \ref{fig:taxnomaydetection}.
%

\textit{Bounding-box regression based methods} for text-detection \cite{jaderberg2014deep,tian2016detecting,zhou2017east,liao2017textboxes,liao2018textboxes++,ma2018arbitrary} regard text as an object and aim to predict the candidate bounding boxes directly.
For example, TextBoxes in  \cite{liao2017textboxes} modified the single-shot descriptor (SSD) \cite{liu2016ssd} kernels by applying long default anchors and filters to handle the significant variation of aspect ratios within text instances. 
In \cite{shi2017detecting}, Shi \textit{et al.} have utilized an architecture inherited from SSD \cite{liu2016ssd} to decompose text into smaller segments and then link them into text instances, so called SegLink, by using spatial relationships or linking predictions between neighboring text segments, which enabled SegLink to detect long lines of Latin and non-Latin text that have large aspect ratios.
The Connectionist Text Proposal Network (CTPN) \cite{tian2016detecting}, a modified version of Faster-RCNN \cite{fasterrcnn_ren2015}, used an anchor mechanism to predict the location and score of each fixed-width proposal simultaneously, and then connected the sequential proposals by a recurrent neural network (RNN). 
Gupta \textit{et al.} \cite{gupta2016synthetic} proposed a fully-convolutional regression network inspired by the YOLO network \cite{YOLO_2016}, while to reduce the false-positive text in images a random-forest classifier was utilized as well.
However, these methods \cite{tian2016detecting,gupta2016synthetic,liao2017textboxes},  which inspired from the general object detection problem, 
may fail to handle multi-orientated text and require further steps to group text components into text lines to produce an oriented text box;  because unlike the general object detection problem, detecting word or text regions require bounding boxes of larger aspect ratio \cite{he2017deep,shi2017detecting}.

With considering that scene text generally appears in arbitrary shapes, several works have tried to improve the performance of detecting multi-orientated text \cite{he2017deep,zhou2017east,shi2017detecting,ma2018arbitrary,liao2018textboxes++}. 
For instance, He \textit{et al.} \cite{he2017deep} proposed a multi-oriented text detection based on direct regression to generate arbitrary quadrilaterals text by calculating offsets between every point of text region and vertex coordinates. This method is particularly beneficial to localize quadrilateral boundaries of scene text, which are hard to identify the constitute characters and have significant variations in scales and perspective distortions.
In EAST \cite{zhou2017east}, FCN is applied to detect text regions directly without using the steps of candidate aggregation and word partition, and then NMS is used to detect word or line text. This method  predicts the rotated boxes or quadrangles of words or text-lines at each point in the text region.
Ma \textit{et al.} \cite{ma2018arbitrary} introduced Rotation Region Proposal Networks (RRPN), based on Faster-RCNN \cite{fasterrcnn_ren2015}, to detect arbitrary-oriented text in scene images. 
Later, Liao \textit{et al.} \cite{liao2018textboxes++} extended TextBoxes to TextBoxes++ by improving the network structure and the training process. Textboxes++ replaced the rectangle bounding boxes of text to quadrilateral to detect arbitrary-oriented text.
Although bounding-box based methods \cite{tian2016detecting,he2017deep,zhou2017east,shi2017detecting,ma2018arbitrary,liao2018textboxes++} have simple architecture, they require complex anchor design, hard to tune during training, and may fail to deal with detecting 
curved text.

\begin{figure}[t]
    \centering
    \includegraphics[width = \linewidth]{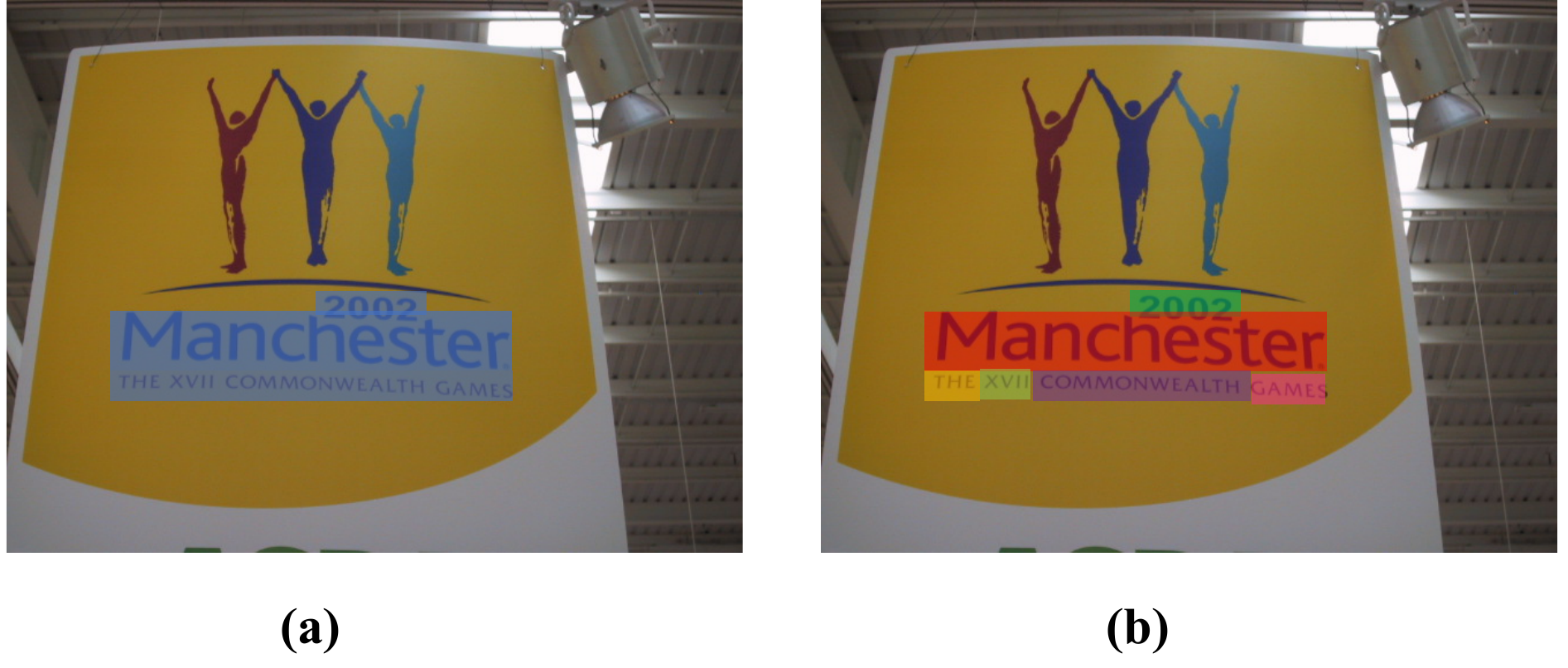}
    \caption{Illustrative example for semantic vs. instance segmentation. Groundtruth annotations for (a) semantic segmentation, where very close characters are linked together, and (b) instance segmentation. The image comes from the public dataset in \cite{karatzas2013icdar}. Note, this figure is best viewed in color format.}
    \label{fig:semantic_problem}
\end{figure}

\textit{Segmentation-based methods}  in \cite{zhang2016multi,yao2016scene,wu2017_self,long2018textsnake,deng2018pixellink,lyu2018multi,qin2019algorithm,pmtd_liu_2019} cast text detection as a \textit{semantic segmentation problem}, which aim to classify text regions in images at the pixel level as shown in Fig. \ref{fig:semantic_problem}(a). 
These methods, first extract text blocks from the segmentation map generated by a FCN \cite{long2015fully}
and then obtain bounding boxes of the text by post-processing.
For example, Zhang \textit{et al.} \cite{zhang2016multi} adopted FCN to predict the salient map of text regions, as well as for predicting the center of each character in a given image.
Yao \textit{et al.} \cite{yao2016scene}, modified FCN to produce three kind of score maps: text/non-text regions, character classes, and character linking orientations of the input images. Then a word partition post-processing method is applied to obtain word bounding boxes with the segmentation maps.
Although these segmentation-based methods \cite{zhang2016multi,yao2016scene} perform well on rotated and irregular text, they might fail to accurately separate the adjacent-word instances that tend to connect. 
%

To address the problem of linked neighbour characters, Pixellinks \cite{deng2018pixellink} leveraged 8-directional information for each pixel to highlight the text margin, and Lyu \cite{lyu2018multi} proposed corner detection method to produce position-sensitive score map. 
In \cite{long2018textsnake}, TextSnake was proposed to detect text instances by predicting the text regions and the center-line together with geometry attributes. 
This method does not require character-level annotation and is capable of reconstructing the precise shape and regional strike of text instances.
%
Inspired by \cite{hu2017wordsup}, character affinity maps were used in \cite{baek2019craft} to connect detected characters into a single word and a weakly supervised framework was used to train a character-level detector.
%
%
To better detection adjacent text instances, in \cite{PSENet_wang2019} a progressive scale expansion network (PSENet) was introduced to find kernels with multiple scales and separate text instances close to each other accurately.
However, the methods in \cite{PSENet_wang2019,baek2019craft} require large number of images for training, which increase the run-time and can present difficulties on platforms with limited resources.

Recently, several works \cite{lyu2018mask,huang2019mask,xie2019scene,pmtd_liu_2019} have treated scene text detection as an \textit{instance segmentation problem}, an example is shown in Fig. \ref{fig:semantic_problem}(b), and many of them have applied Mask R-CNN \cite{Maskrcnn_He2017} framework to improve the performance of scene text detection, which 
is useful for detecting text instances of arbitrary shapes.
For example, inspired by Mask R-CNN, 
SPCNET \cite{xie2019scene} uses a text context module to detect text of arbitrary shapes and a re-score mechanism to suppress false positive detections.
However, the methods in \cite{lyu2018mask,huang2019mask,xie2019scene} have some drawbacks, which may decline their performance:
Firstly, they suffer from the errors of bounding box handling in a complicated background, where the predicted bounding box fails to cover the whole text image. 
Secondly, these methods \cite{lyu2018mask,huang2019mask,xie2019scene} aim at separating text pixels from the background ones that can lead to many mislabeled pixels at the text borders
\cite{pmtd_liu_2019}.

\textit{Hybrid} methods \cite{he2017single,liao2018rotation,lyu2018,zhang2019look} use 
segmentation-based approach to predict score maps of text and aim at the same time to obtain text bounding-boxes through regression.
For example, single-shot text detector (SSTD) \cite{he2017single} used an attention mechanism to enhance text regions in image and reduce background interference on the feature level.
Liao \textit{et al.} \cite{liao2018rotation} proposed rotation-sensitive regression for oriented scene text detection, which makes full use of rotation-invariant features by actively rotating the convolutional filters. However, this method is incapable of capturing all the other possible text shapes that exist in scene images \cite{baek2019craft}.
Lyu \textit{et al.} \cite{lyu2018} presented a method that 
detects and groups corner points of text regions to generate text boxes. 
Beside detecting long oriented text and handling considerable variation in aspect ratio, this method also requires simple post-processing. 
Liu \textit{et al.} \cite{pmtd_liu_2019} proposed a new Mask R-CNN-based framework, namely, pyramid mask text detector (PMTD) that assigns a soft pyramid label, $l \in [0,1]$, 
for each pixel in text instance, and then reinterprets the obtained 2D soft mask into the 3D space. 
Then, a novel plane clustering algorithm is employed on the soft pyramid to infer the optimal text box that helped this method to achieve the state-of-the-art performance on several recent datasets \cite{icdar2019,karatzas2015icdar,icdar2017}. 
However, due to PMTD framework is designed explicitly for handling multi-oriented text, it is still underperfoming on curved-text datasets \cite{ch2017total,CTW_1500_yuliang2017}.


\subsection{Text Recognition}
\label{sec:recognition}
The scene text recognition task aims to convert detected text regions into characters or words. Case sensitive character classes often consist of: 10 digits, 26 lowercase letters, 26 uppercase letters, 32 ASCII punctuation marks, and the end of sentences (EOS) symbol. 
However, text recognition models proposed in the literature have used different choices of character classes, which Table \ref{tab:recognition_comp} provides their numbers.

Since the properties of scene text images are  different from that of scanned documents, it is difficult to develop an effective text recognition method based on a classical OCR or handwriting recognition method, such as \cite{bunke1997handbook,zhou1997extracting, sawaki2000automatic, arica2001overview, lucas2003icdar, ICDAR2005}. 
As we mentioned in Section \ref{Introduction}, this is because images captured in the wild tend to include text under various challenging conditions such as images of low resolution \cite{Mishra_12_IIIT, wang2010word}, lightning extreme \cite{Mishra_12_IIIT,wang2010word}, environmental conditions \cite{karatzas2015icdar,karatzas2013icdar}, and have different number of fonts \cite{karatzas2015icdar,karatzas2013icdar, cut80_2014}, orientation angles \cite{cut80_2014, ch2017total}, languages \cite{icdar2017} and lexicons \cite{wang2010word,Mishra_12_IIIT}. Researchers proposed different techniques to address these challenging issues, which can be categorized into \textit{classical machine learning-based} \cite{sawaki2000automatic, de2009character, pan2009text, wang2011end, wang2010word, neumann2012real} and \textit{deep learning-based} \cite{jaderberg2014synthetic, shi2016end, jaderberg2016reading,shi2016robust, lee2016recursive,  STARNet_2016, wang2017gated, yang2017learning,cheng2017fan, liu2018char-net,  cheng2018aon, bai2018ep, rosetta2018, rosetta2018, liu2018synthetically, xie2019ACE, zhan2019esir, Wang2019ASA} text recognition methods, which are discussed in the rest of this section.

\subsubsection{Classical Machine Learning-based Methods}
In the past two decades, traditional scene text recognition methods \cite{sawaki2000automatic, de2009character, pan2009text} have used standard image features, such as HOG \cite{dalal_hog} and SIFT \cite{lowe2004_sift}, with a \textit{classical machine learning} classifier, such as SVM \cite{suykens1999_svm} or k-nearest neighbors \cite{K_NN}, then a statistical language model or visual structure prediction is applied to prune-out mis-classified characters \cite{survey2015, survey_zhu2016scene}.
%

Most classical machine learning-based methods follow a \textit{bottom-up} approach that classified \textit{characters} are linked up into words.
For example, in \cite{wang2011end, wang2010word} HOG features are first extracted from each sliding window, and then a pre-trained nearest neighbor or SVM  classifier is applied to classify the characters of the input word image. 
Neumann and Matas \cite{neumann2012real} proposed a set of handcrafted features,  which include aspect and hole area ratios, used with an SVM classifier for text recognition.
However, these methods \cite{wang2011end, wang2010word, neumann2012real,bissacco2013photoocr} cannot achieve either an effective recognition accuracy, due to the low representation capability of handcrafted features, or building models that are able to handle text recognition in the wild. 
%
Other works adopted a \textit{top-down} approach, where the \textit{word} is directly recognized from the entire input images, rather than detecting and recognizing individual characters \cite{almazan2014word}. 
For example,
Almazan \textit{et al.} \cite{almazan2014word} treated word recognition as a content-based image retrieval problem, where word images and word labels are embedded into an Euclidean space and the embedding vectors are used to match images and labels.
One of the main problems of using these methods \cite{almazan2014word,gordo_2015,rodriguez_2013} is that they fail in recognizing input word images outside of the word-dictionary dataset.


\subsubsection{Deep Learning-based Methods}
\label{sec:recognition_deep}
With the recent advances in deep neural network architectures \cite{VGG,ResNet_He2015L,long2015fully,RCNN_Lee2016}, many researchers proposed \textit{deep learning-based} methods \cite{wang2012end, bissacco2013photoocr, jaderberg2014synthetic} to tackle the challenges of recognizing text in the wild. Table \ref{tab:recognition_comp} illustrates a comparison among some of the recent state-of-the-art deep learning-based text recognition methods \cite{he2016recog,shi2016robust,lee2016recursive,STARNet_2016,shi2016end,wang2017gated,yang2017learning,cheng2017fan,liu2018char-net,cheng2018aon,bai2018ep,Liao2018STR_CAFCN,rosetta2018,shi2018aster,liu2018synthetically,xie2019ACE,zhan2019esir,Wang2019ASA,Wan2019_2DCTC}.
For example, Wang \textit{et al.} \cite{wang2012end} proposed a CNN-based feature extraction framework for character recognition, then applied the NMS technique of \cite{neubeck2006_NMS} to obtain the final word predictions. 
Bissacco \textit{et al.} \cite{bissacco2013photoocr} employed a fully connected network (FCN) for character feature representation, then to recognize characters an n-gram approach was used. 
Similarly, \cite{jaderberg2014synthetic} designed a deep CNN framework with multiple softmax classifiers, trained on a new synthetic text dataset, which each character in the word images predicted with these independent classifiers.
These early deep CNN-based character recognition methods \cite{jaderberg2014synthetic, wang2012end,bissacco2013photoocr} require localizing each character, which may be challenging due to the complex background, irrelevant symbols, and the short distance between adjacent characters in scene text images.

\begin{table*}[]
    \centering
    \caption{Comparison among some of the state-of-the-art of the deep learning-based text recognition methods, where TL: Text-line, C: Character, Seq: Sequence Recognition, PD: Private Dataset, HAM: Hierarchical Attention Mechanism, ACE: Aggregation Cross-Entropy, and the rest of the abbreviations are introduced in Table \ref{tab:recognition_Atrribution}. }
    \scalebox{0.91}{
 \setlength{\tabcolsep}{2.5 pt}
\begin{tabular}{lcccccccccc}
\toprule
Method           & Model & Year  & Feature Extraction & Sequence modeling & Prediction & Training Dataset$^{\dagger}$ & Irregular recognition & Task & \# classes & Code \\
\toprule
Wang \textit{et al.} \cite{wang2012end}                &E2ER    & 2012  &CNN             & --     &SVM    & PD        &   --  & C     &62 &  --  \\
Bissacco \textit{et al.} \cite{bissacco2013photoocr}   &PhotoOCR& 2013  &HOG,CNN         & --     &  --   & PD        &   --  & C     &99 & --   \\
Jaderberg \textit{et al.} \cite{jaderberg2014synthetic}&SYNTR   & 2014  &CNN             & --     &  --   & MJ        &   --  & C     &36 &\cmark \\
Jaderberg \textit{et al.} \cite{jaderberg2014synthetic}&SYNTR   & 2014  &CNN             & --     &  --   & MJ        &   --  & W     &90k &\cmark \\
He \textit{et al.} \cite{he2016recog}                  &DTRN    & 2015  &DCNN            & LSTM   & CTC   & MJ        &   --  & Seq   &37 &  --  \\
Shi \textit{et al.}* \cite{shi2016robust}              &RARE    & 2016  &STN+VGG16       & BLSTM  & Attn  & MJ        & \cmark& Seq   &37 & \cmark\\
Lee \textit{et al.} \cite{lee2016recursive}            &R2AM    & 2016  & Recursive CNN  & LTSM   &Attn   & MJ        &   --  & C     &37 &  --   \\
Liu \textit{et al.}* \cite{STARNet_2016}               &STARNet & 2016  &STN+RSB         & BLSTM  & CTC   & MJ+PD     & \cmark& Seq   &37 & \cmark\\
Shi \textit{et al.}* \cite{shi2016end}                 &CRNN    & 2017  &VGG16           & BLSTM  & CTC   & MJ        &  --   &Seq    &37 &\cmark \\
Wang \textit{et al.} \cite{wang2017gated}              &GRCNN   & 2017  & GRCNN          & BLSTM  & CTC   & MJ        &  --   &Seq    &62 &  --   \\
Yang \textit{et al.} \cite{yang2017learning}           &L2RI    & 2017  & VGG16          & RNN    & Attn  & PD+CL     &\cmark &Seq    &-- &  --   \\
Cheng \textit{et al.} \cite{cheng2017fan}              &FAN     & 2017  & ResNet         & BLSTM  & Attn  & MJ+ST+CL  &  --   &Seq    &37 & --   \\
Liu \textit{et al.}  \cite{liu2018char-net}            &Char-Net& 2018  &CNN             & LTSM   & Att   & MJ        &\cmark &C      &37 & --   \\
Cheng \textit{et al.} \cite{cheng2018aon}              &AON     & 2018  &AON+VGG16       & BLSTM  & Attn  & MJ+ST     &\cmark &Seq    &37 & --   \\
Bai \textit{et al.} \cite{bai2018ep}                   &EP      & 2018  &ResNet          & --     & Attn  & MJ+ST     &  --   &Seq    &37 &  --  \\
Liao \textit{et al.} \cite{Liao2018STR_CAFCN}          &CAFCN   & 2018  & VGG            & --     & --    & ST        &\cmark &C      &37 &  --  \\
Borisyuk \textit{et al.}*  \cite{rosetta2018}          &ROSETTA & 2018  &ResNet          & --     & CTC   & PD        &  --   &Seq    & --&   --  \\
Shi \textit{et al.}* \cite{shi2018aster}               &ASTER   & 2018  &STN+ResNet      & BLSTM  & Attn  & MJ+ST     &\cmark &Seq    &94 & \cmark\\
Liu \textit{et al.} \cite{liu2018synthetically}        &SSEF    & 2018  &VGG16           & BLSTM  & CTC   & MJ        &\cmark &Seq    &37 &   --  \\
Baek \textit{et al.}* \cite{baek2019STR}               &CLOVA   & 2018  &STN+ResNet      & BLSTM  & Attn  & MJ+ST     &\cmark &Seq    &36 & \cmark\\
Xie \textit{et al.} \cite{xie2019ACE}                  &ACE     & 2019  & ResNet         & --     & ACE   & ST+MJ     &\cmark &Seq    &37 & \cmark \\
Zhan \textit{et al.} \cite{zhan2019esir}               &ESIR    & 2019  &IRN+ResNet,VGG  & BLSTM  & Attn  & ST+MJ     &\cmark &Seq    &68 &  --    \\
Wang \textit{et al.} \cite{Wang2019ASA}                &SSCAN   & 2019  & ResNet,VGG     & --     &Attn   & ST        &\cmark &Seq    &94 &  --    \\
Wang \textit{et al.} \cite{Wan2019_2DCTC}              &2D-CTC  & 2019  & PSPNet         & --     &2D-CTC & ST+MJ     &\cmark &Seq    &36 &  --    \\

\bottomrule

\end{tabular}}

    \begin{tablenotes}
      \small
      \item[] Note: * This method has been considered for evaluation.
      \item[] $\dagger$ Trained dataset/s used in the original paper. We used a pre-trained model of MJ+ST datasets for evaluation to compare the results in a unified framework.
    \end{tablenotes}

\label{tab:recognition_comp}
\end{table*}

\begin{table}[]
    \centering
    \caption{The description of abbreviations.}
    \begin{tabular}{l|l}
    \toprule
    Attribution    & Description  \\
    \toprule
    Attn	&  attention-based sequence prediction \\
    BLSTM 	& Bidirectional LTSM\\
    CTC	    & Connectionist temporal classification\\
    CL	    & Character-labeled\\
    MJ	    & MJSynth\\
    ST	    & SynthText \\
    PD	    & Private Data\\
    STN	    & Spatial Transformation Network \cite{Jaderberg_STN} \\
    TPS	    & Thin-Plate Spline \\
    PSPNet  &Pyramid Scene Parsing Network \cite{zhao2017_PSPNet}\\
    
    \bottomrule

    \end{tabular}
    \label{tab:recognition_Atrribution}
\end{table}{}

For word recognition, Jaderberg \textit{et al.} \cite{jaderberg2016reading} conducted a 90k English word classiﬁcation task with a CNN architecture. 
Although this method \cite{jaderberg2016reading} showed a better word recognition performance compared to just the individual character recognition methods \cite{wang2012end,bissacco2013photoocr,jaderberg2014synthetic}, it has two main drawbacks: (1) this method can not recognize out-of-vocabulary words, and (2) deformation of long word images may affect its recognition rate.

\begin{figure}[t]
    \centering
    \includegraphics[width = \linewidth]{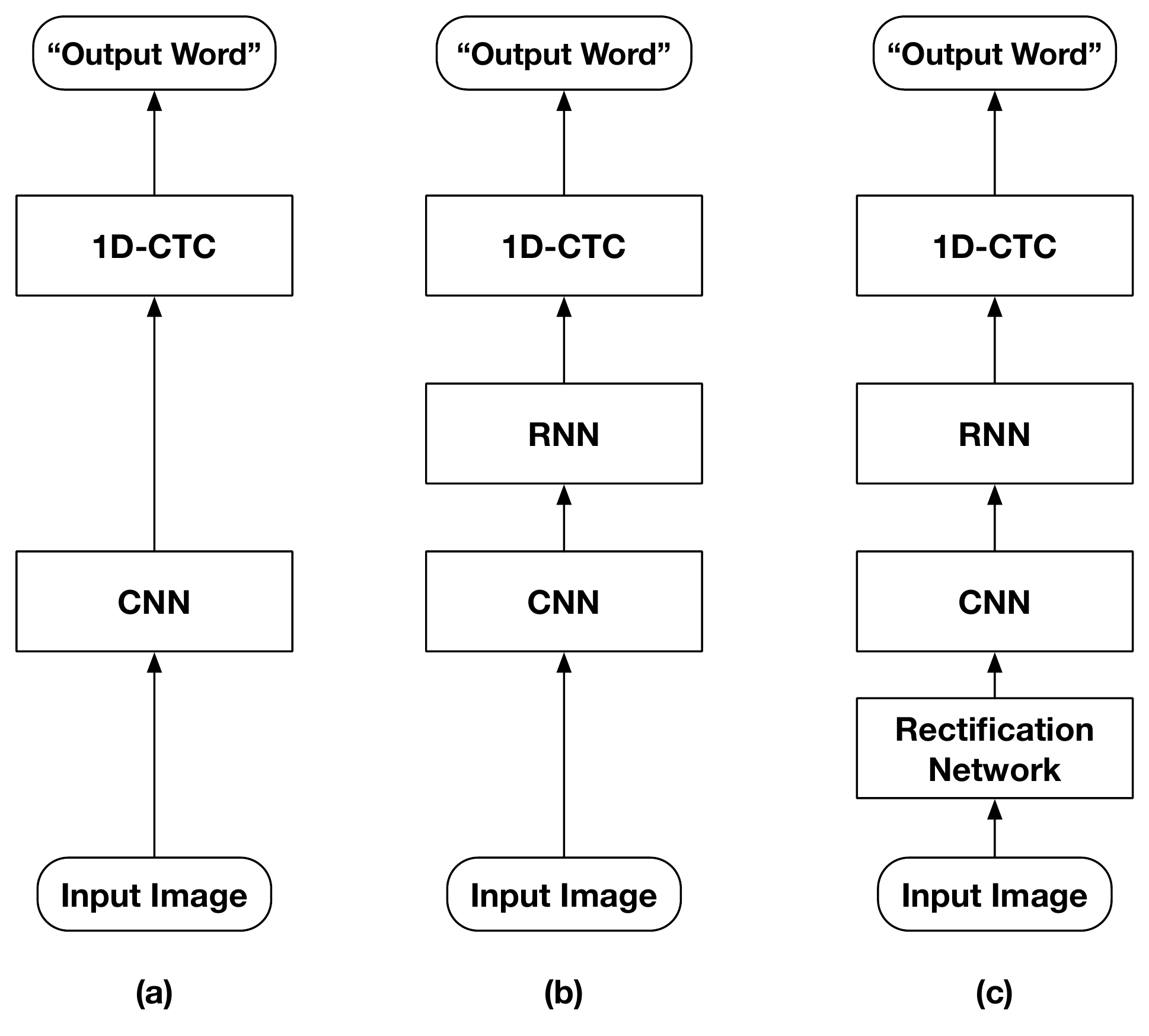}
    \caption{Comparison among some of the recent 1D CTC-based scene text recognition frameworks, where 
    (a) baseline frame of CNN with 1D-CTC \cite{rosetta2018},
    (b) adding RNN to the baseline frame \cite{shi2016end}, 
    and (c) using a Rectification Network before the framework of (b) \cite{STARNet_2016}.}
    \label{fig:ctc_based}
\end{figure}

With considering that scene text generally appears in the form of a \textit{sequence} of characters, many recent works \cite{he2016recog,shi2016robust,STARNet_2016,shi2016end,cheng2017fan,cheng2018aon,bai2018ep,liu2018synthetically,xie2019ACE,zhan2019esir,Wang2019ASA} have mapped every input sequence into an output sequence of variable length.
%
%
%
%
Inspired by the speech recognition problem, several sequence-based text recognition methods \cite{shi2016end, he2016recog, STARNet_2016, wang2017gated, lee2016recursive, rosetta2018, liu2018synthetically} have used \textit{connectionist temporal classification} (CTC) \cite{graves2006_ctc} for prediction of character sequences. Fig. \ref{fig:ctc_based} illustrates three main CTC-based text recognition frameworks that have been used in the literature. 
In the first category \cite{yin2017scene,rosetta2018}, CNN models (such as VGG \cite{VGG}, RCNN \cite{RCNN_Lee2016} and ResNet \cite{ResNet_He2015L}) have been used with CTC as shown in Fig. \ref{fig:ctc_based}(a). 
For instance, in \cite{yin2017scene}, a sliding window is first applied to the text-line image in order to effectively capture contextual information, and then a CTC prediction is used to predict the output words. 
Rosetta \cite{rosetta2018}  used only the extracted features from convolutional neural network by applying a ResNet model as a backbone to predict the feature sequences. 
Despite reducing the computational complexity, these methods \cite{yin2017scene,rosetta2018} suffered the lack of contextual information and showed a low recognition accuracy.

\begin{figure}[t]
    \centering
    \includegraphics[width = \linewidth,trim=0.18in 0.2in 0.15in 0.175in, clip]{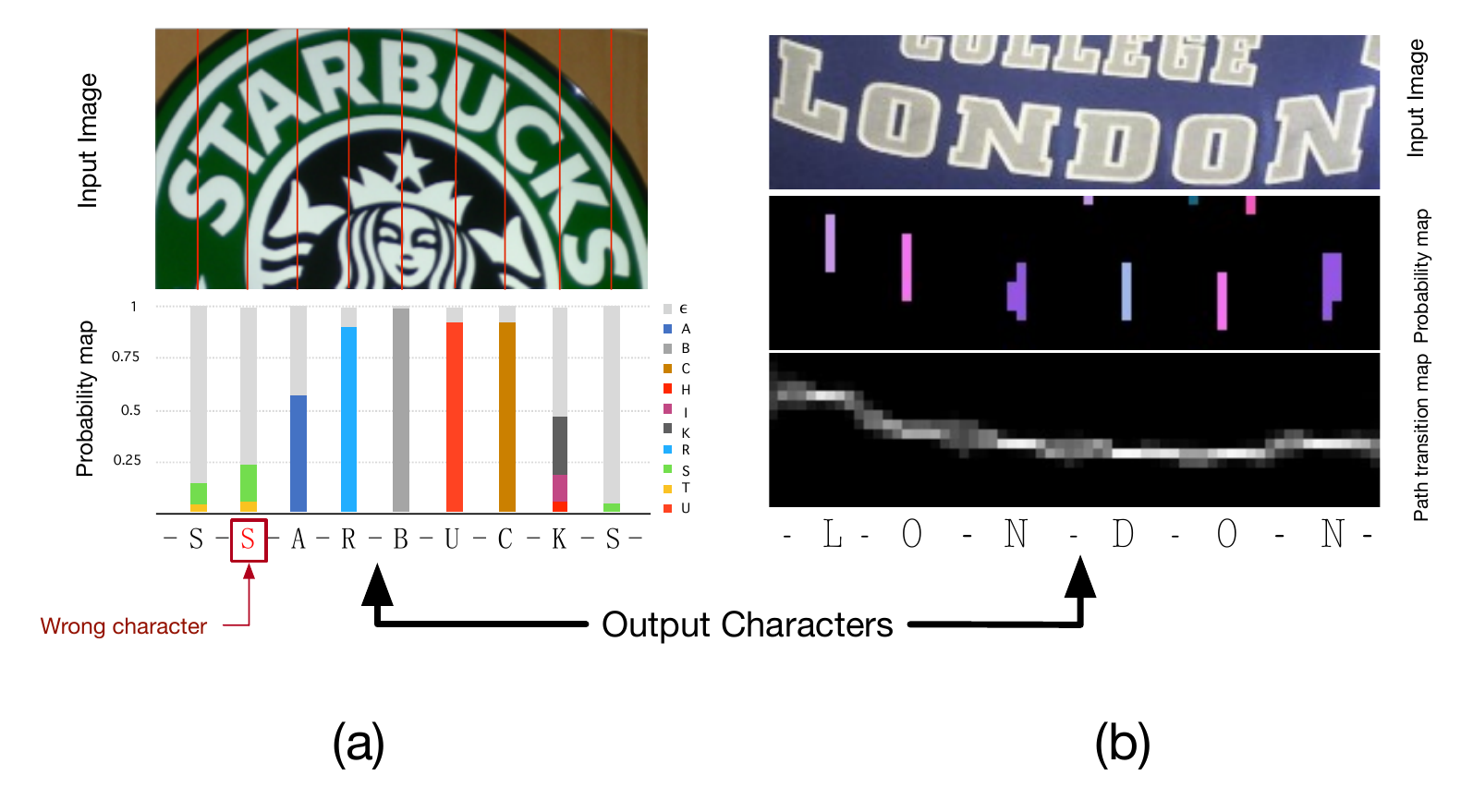}
    \caption{Comparing the processing steps for tackling the character recognition problem using (a) 1D-CTC \cite{graves2006_ctc}, and (b) \mbox{2D-CTC} \cite{Wan2019_2DCTC}.}
    \label{fig:1DCTCvs2DCTC}
\end{figure}

For better extracting contextual information, several works \cite{shi2016end,he2016recog,wang2017gated} have used RNN \cite{yang2017learning} combined with CTC to identify the conditional probability between the predicted and the target sequences (Fig. \ref{fig:ctc_based}(b)).
For example, in \cite{shi2016end} a VGG model \cite{VGG_su2014} is employed as a backbone to extract features of input image followed by a bidirectional long-short-term-memory (BLSTM) \cite{LSTM_1997} for extraction of contextual information and then a CTC loss is applied to identify sequence of characters.
Later, Wang \textit{et al.} \cite{wang2017gated} proposed a new architecture based on recurrent convolutional neural network (RCNN), namely gated RCNN (GRCNN), which used a gate to modulate recurrent connections in a previous model RCNN. 
However, as illustrated in Fig. \ref{fig:1DCTCvs2DCTC}(a) these techniques \cite{shi2016end, he2016recog, wang2017gated} are insufficient to recognize irregular text \cite{xie2019ACE} as characters are arranged on a 2-dimensional (2D) image plane and the CTC-based methods are only designed for 1-dimensional (1D) sequence to sequence alignment,  
therefore these methods require converting 2D image features into 1D features, which may lead to loss of relevant information \cite{Wan2019_2DCTC}.

To handle irregular input text images, Liu \textit{et al.} \cite{STARNet_2016} proposed a spatial-attention residue Network (STAR-Net) that leveraged a spatial transform network (STN) \cite{Jaderberg_STN} for tackling text distortions. 
It is shown in \cite{STARNet_2016} that the usage of STN within the residue
convolutional blocks, BLSTM and CTC framework, shown in Fig. \ref{fig:ctc_based}(c), allowed performing scene text recognition under various distortions. 
Recently, Wang \textit{et al.} \cite{Wan2019_2DCTC} introduced a 2D-CTC technique 
in order to overcome the limitations of 1D-CTC based methods. This method \cite{Wan2019_2DCTC} can be directly applied on 2D probability distributions to produce more precise recognition.
For this purpose, as shown in Fig. \ref{fig:1DCTCvs2DCTC}(b),
beside the time step, an extra height dimension is also added for path searching to consider all the possible paths over the height dimensions in order to better align the search space and focus on relevant features.

{}

%

\begin{figure*}[t]
    \centering
    \includegraphics[width=\linewidth]{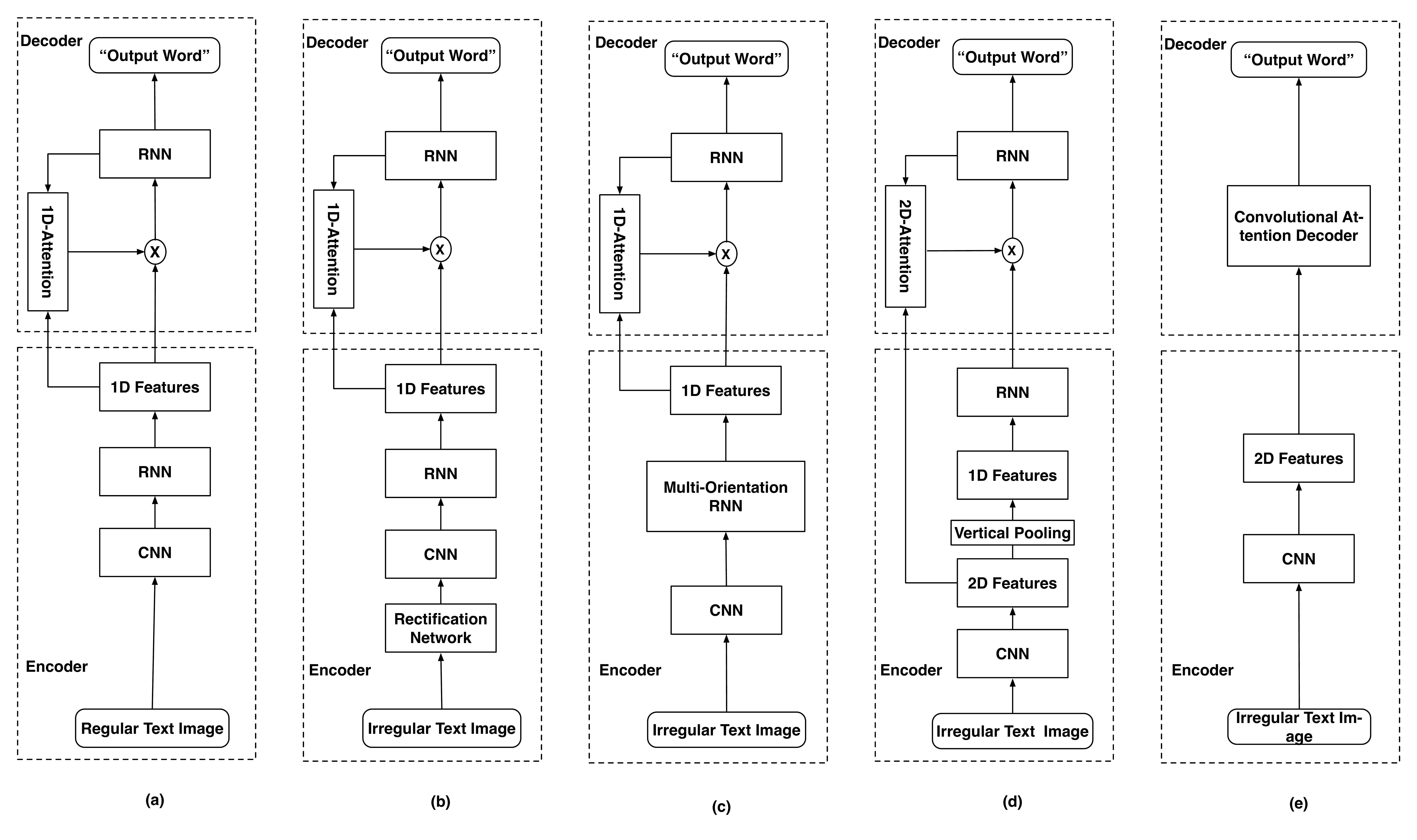}
    \caption{Comparison among some of the recent attention-based scene text recognition frameworks,  
    where (a), (b) and (c) are 1D-attention-based frameworks used in a basic model \cite{lee2016recursive}, rectification network of ASTER \cite{shi2018aster}, and multi-orientation encoding of AON \cite{cheng2018aon}, respectively, 
    (d) 2D-attention-based decoding used in \cite{li2019_SAR},
    (e) convolutional attention-based decoding used in SRCAN \cite{Wang2019ASA} and FACLSTM \cite{wang2019faclstm}.}
    \label{fig:Attentib_BD}
\end{figure*}

The \textit{attention mechanism} that first used for machine translation in \cite{bahdanau2014neural} has been also adopted for scene text recognition \cite{shi2016robust,lee2016recursive,STARNet_2016,yang2017learning,liu2018char-net,shi2018aster, cheng2018aon,zhan2019esir}, where an implicit attention is learned automatically to enhance deep features in the decoding process. Fig. \ref{fig:Attentib_BD} illustrates  five main attention-based text recognition frameworks that have been used in the literature.
For regular text recognition, a basic 1D-attention-based encoder and decoder framework, as presented in Fig. \ref{fig:Attentib_BD}(a), is used to recognize text images in \cite{lee2016recursive,wojna_2017,deng2016you}. 
For example, Lee and Osindero \cite{lee2016recursive} proposed a recursive recurrent neural network with attention modeling (R2AM), where a recursive CNN is used for image encoding in order to learn broader contextual information, then a 1D attention-based decoder is applied for sequence generation. However, directly training R2AM on irregular text is difficult due to the on-horizontal character placement \cite{yang2017recog}.

Similar to CTC-based recognition methods in handling irregular text, many attention-based methods \cite{shi2018aster,liu2018char-net,zhan2019esir,baek2019STR,luo_2019_moran} have used image rectification modules to control distorted text images as shown in Fig. \ref{fig:Attentib_BD}(b). 
For instance, Shi \textit{et al.} \cite{shi2016robust,shi2018aster} proposed a text recognition system that combined attention-based sequence and a STN module to rectify 
irregular text (\textit{e.g.} curved or perceptively distorted), then the text within the rectified image is recognized by a RNN network. 
However, training a STN-based method without considering human-designed geometric ground truth is difficult, especially, in complicated arbitrary-oriented or strong-curved text images. 

Recently, many methods \cite{liu2018char-net,zhan2019esir,luo_2019_moran} proposed several techniques to rectify irregular text. 
For instance, instead of rectifying the entire word image, Liu \textit{et al.} \cite{liu2018char-net} presented a Character-Aware Neural Network (Char-Net) for recognizing distorted scene characters. Char-Net includes a word-level encoder, a character-level encoder, and a LSTM-based decoder. Unlike STN, Char-Net can detect and rectify individual characters using a simple local spatial transformer. 
This leads to the detection of more complex forms of distorted text, which cannot be recognized easily by a global STN. 
However, Char-Net fails where the images contain sever blurry text. 
In \cite{zhan2019esir}, a robust line-fitting transformation is proposed to correct the prospective and curvature distortion of scene text images in an iterative manner. 
For this purpose, an iterative rectification network using the thin plate spline (TPS) transformation is applied in order to increase the rectification of curved images, and thus improved the performance of recognition. However, the main drawback of this method is the high computational cost due to the multiple rectification steps.
Luo \textit{et al.} \cite{luo_2019_moran} proposed two mechanisms to improve the performance of text recognition, a new multi-object rectified attention
network (MORAN) to rectify irregular text images and a fractional pickup mechanism to enhance the sensitivity of the attention-based network in the decoder. 
However, this method fails on complicated backgrounds, where the curve angle in text image is too large.

%


In order to handle oriented text images, Cheng \textit{et al.} \cite{cheng2018aon} proposed an arbitrary orientation network (AON) to extract deep features of images in four orientation directions, then a designed filter gate is applied to generate the integrated sequence of features. Finally, a 1D attention-based decoder is applied to generate character sequences. The overall architecture of this method is shown in Fig. \ref{fig:Attentib_BD}(c). 
Although AON can be trained by using  word-level annotations, it leads to redundant representations due to using this complex four directional network.

%
    


The performance of attention-based methods may decline in more challenging conditions, such as images of low-quality and sever distorted text, text affected by these conditions may lead to misalignment and attention drift problems \cite{Wan2019_2DCTC}.
To reduce the severity of these problems, Cheng \textit{et al.} \cite{cheng2017fan} proposed a focusing attention network (FAN) that consists of an attention network (AN) for character recognition and a focusing network (FN) for adjusting the attention of AN. It is shown in \cite{cheng2017fan} that FAN is able to correct the drifted attention automatically, and hence, improve the regular text recognition performance. 
%
%
%

Some methods \cite{yang2017learning,li2019_SAR,Liao2018STR_CAFCN} used 2D attention \cite{2DAttention_xu2015}, as presented in Fig. \ref{fig:Attentib_BD}(d), to overcome the drawbacks of 1D attention.
These methods can learn to focus on individual character features in the 2D space during decoding, which can be trained using either character-level \cite{yang2017learning} or word-level \cite{li2019_SAR} annotations.
For example, Yang \textit{et al.} \cite{yang2017learning} introduced an auxiliary dense character detection task using a fully convolutional network (FCN) for encouraging the learning of visual representations to improve the recognition of irregular scene text. 
Later, Liao \textit{et al.} \cite{Liao2018STR_CAFCN} proposed a framework called Character Attention FCN (CA-FCN), 
which models the irregular scene text recognition problem in a 2D space instead of the 1D space as well. In this network, a character attention module \cite{Wang_Attention} is used to predict multi-orientation characters in an arbitrary shape of an image. Nevertheless, this framework requires character-level annotations and cannot be
trained end-to-end \cite{lyu2018mask}.
In contrast, Li \textit{et al.} \cite{li2019_SAR} proposed a model that used word-level annotations, which enables this model to utilize both real and synthetic data for training without using character-level annotations. 
However, 2-layer RNNs are adopted respectively in both encoder and
decoder, which precludes computation parallelization and suffers from heavy computational burden.


To address these computational cost issue of 2D-attention-based techniques \cite{yang2017learning,li2019_SAR,Liao2018STR_CAFCN}, in \cite{wang2019faclstm} and \cite{Wang2019ASA} the RNN stage of 2D-attentions techniques were eliminated, and a convolution-attention network \cite{ConvLSTM_shi} was used instead, enabling irregular text recognition, as well as fully parallel computation that accelerates the processing speed. 
Fig. \ref{fig:Attentib_BD}(e) shows a general block diagram of this attention-based category.
For example, Wang \textit{et al.} \cite{Wang2019ASA} proposed a simple and robust convolutional-attention network (SRACN), where convolutional attention network decoder is directly applied into 2D CNN features. 
SRACN does not require to convert input images to sequence representations and directly can map text images into character sequences. 
Meanwhile, Wang \textit{et al.} \cite{wang2019faclstm} considered the scene text recognition as a spatio-temporal prediction problem and proposed the focus attention convolution LSTM (FACLSTM) network for scene text recognition. 
It is shown in \cite{wang2019faclstm} that FACLSTM is more effective for text recognition, specifically for curved scene text datasets, such as CUT80 \cite{cut80_2014} and SVT-P \cite{svtp_2013}.

\section{Experimental Results}
\label{sec:ExpResults}
In this section, we present an extensive evaluation for some selected state-of-the-art scene text detection \cite{pmtd_liu_2019,baek2019craft,zhou2017east,wang2019_PAN,PSENet_wang2019,liu2019_BOX,deng2018pixellink} and recognition \cite{baek2019STR,shi2018aster,shi2016end,rosetta2018,STARNet_2016,shi2016robust} techniques on recent public datasets \cite{Mishra_12_IIIT,svtp_2013,ICDRA2003,karatzas2013icdar,karatzas2015icdar,cut80_2014,veit2016coco} that include wide variety of challenges. 
%
One of the important characteristics of a scene text detection or recognition scheme  is to be generalizable, which shows how a trained model on one dataset is capable of detecting or recognizing challenging text instances on other datasets. 
This evaluation strategy is an attempt to close the gap in evaluating text detection and recognition methods that are used to be mainly trained and evaluated on a specific dataset. 
Therefore, to evaluate the generalization ability for the methods under consideration, we propose to compare both detection and recognition models on unseen datasets. 

Specifically, we selected the following methods for evaluation of the recent advances in the deep learning-based schemes for scene text detection: PMTD\footnote{\url{https://github.com/jjprincess/PMTD}} \cite{pmtd_liu_2019}, CRAFT\footnote{\url{https://github.com/clovaai/CRAFT-pytorch}} \cite{baek2019craft}, EAST\footnote{\url{https://github.com/ZJULearning/pixel_link}} \cite{zhou2017east},
PAN\footnote{\url{https://github.com/WenmuZhou/PAN.pytorch}} \cite{wang2019_PAN},
MB\footnote{\url{https://github.com/Yuliang-Liu/Box_Discretization_Network}} \cite{liu2019_BOX}, 
PSENET\footnote{\url{https://github.com/WenmuZhou/PSENet.pytorch}} \cite{PSENet_wang2019}, and 
Pixellink\footnote{\url{https://github.com/argman/EAST}} \cite{deng2018pixellink}.
For each method except MB \cite{liu2019_BOX}, we used the corresponding pre-trained model directly from the authors' GitHub page that was trained on ICDAR15 \cite{karatzas2015icdar} dataset. While for MB \cite{liu2019_BOX}, we trained the algorithm on ICDAR15 according to the code that was provided by the authors.  
For testing the detectors in consideration, the ICDAR13 \cite{karatzas2013icdar}, ICDAR15 \cite{karatzas2015icdar}, and COCO-Text \cite{lin2014microsoft} datasets have been used.
This evaluation strategy avoids an unbiased evaluation and allows assessment for the generalizability of these techniques.
Table \ref{tab:Text_dataset} illustrates the number of test images for each of these datasets.

For conducing evaluation among the scene text recognition schemes the following deep-learning based techniques have been selected: CLOVA \cite{baek2019STR}, ASTER \cite{shi2018aster}, CRNN \cite{shi2016end}, ROSETTA \cite{rosetta2018}, STAR-Net \cite{STARNet_2016} and RARE \cite{shi2016robust}. 
Since recently the SynthText (ST) \cite{gupta2016synthetic} and MJSynth (MJ) \cite{jaderberg2014synthetic} synthetic datasets have been used extensively for building recognition models, we aim to compare the state-of-the-arts methods when using these synthetic datasets.
All recognition models have been trained on combination of SynthText \cite{gupta2016synthetic} and MJSynth \cite{jaderberg2014synthetic} datasets, while for evaluation we have used ICDAR13 \cite{karatzas2013icdar}, ICDAR15 \cite{karatzas2015icdar}, and COCO-Text \cite{lin2014microsoft} datasets, in addition to four mostly used datasets, namely, III5k \cite{Mishra_12_IIIT}, CUT80 \cite{cut80_2014}, SVT \cite{wang2010word}, and SVT-P \cite{svtp_2013} datasets.
 As shown in Table \ref{tab:Text_dataset}, the selected datasets cover datasets that mainly contain regular or horizontal text images and other datasets that include curved, rotated and distorted, or so called irregular, text images.
 Throughout this evaluation also, we used 36 classes of alphanumeric characters, 10 digits (0-9) + 26 capital English characters (A-Z) = 36.

%

In the remaining part of this section, we will start by summarizing the challenges within each of the utilized datasets (Section \ref{sec:OCRDatasets}) and then presenting the evaluation metrics (Section \ref{sec:OCRmetrics}). Next, we present the quantitative and qualitative analysis, as well as discussion on scene text detection methods (Section \ref{sec:EvalDetectionTechniques}), and on scene text recognition methods (Section \ref{sec:EvalRecognitionTechniques}).

\begin{table*}[h]
\centering
\caption{Comparison among some of the recent text detection and recognition datasets.}
\label{tab:Text_dataset}
\resizebox{\linewidth}{!}{%
\begin{tabular}{l c c c c c c c c c c c c c c c c c c} \toprule
    
	 \multirow{2}{*}{Dataset}& \multirow{2}{*}{Year} & \multicolumn{3}{c}{\# Detection Images} &\multicolumn{2}{c}{\# Recognition words}  & \multicolumn{3}{c}{Orientation}  & \multicolumn{3}{c}{Properties}  & \multicolumn{2}{c}{Task}  \\ \cline{3-5} \cline{6-7} \cline{8-10} \cline{11-13}\cline{14-15}
	 &                                              &Train & Test  &Total  &Train  &Test   & H & MO & Cu &Language & Annotation &  & D&R\\ \toprule
	IC03* \cite{lucas2003icdar}                     & 2003    &258   &251    & 509   &1156  & 1110   & \cmark &  --   &  --    & EN      &W,C  &   &\cmark  &\cmark\\ 
	SVT* \cite{wang2010word}                        & 2010    &100   &250    & 350   &--    & 647    & \cmark &  --   &  --    & EN      &W    &   &\cmark  &\cmark\\ 
    IC11 \cite{shahab2011icdar}                     & 2011    &100   &250    &350    &211   &514     & \cmark &  --   &  --    & EN      &W,C  &   &\cmark  &\cmark\\ 
	IIIT 5K-words* \cite{Mishra_12_IIIT}            & 2012    &--    &--     & --    &2000  &3000    & \cmark &  --   &  --    & EN      &W    &   &--      &\cmark\\
	MSRA-TD500 \cite{yao2012detecting}              & 2012    &300   &200    & 500   &--    & --      & \cmark &\cmark &  --    & EN, CN &TL   &   &\cmark  &-- \\
	SVT-P* \cite{svtp_2013}                         & 2013    & --   &238    & 238   &--    &639     & \cmark &\cmark &  --    & EN      &W    &   &\cmark  &\cmark\\
	ICDAR13* \cite{karatzas2013icdar}               & 2013    &229   &233    & 462   &848   &1095    & \cmark &  --   &  --    & EN      &W    &   &\cmark  &\cmark\\
	CUT80* \cite{cut80_2014}                        & 2014    &--    &80  & 80    &--    &280     & \cmark &\cmark &\cmark  & EN      &W    &   &\cmark  &\cmark\\
	COCO-Text* \cite{lin2014microsoft}          & 2014    &43686 &20000  &63686  &118309 &27550  & \cmark &\cmark &\cmark  &EN       &W    &   &\cmark  &\cmark\\
	ICDAR15* \cite{karatzas2015icdar}               & 2015    &1000  &500    &1500   &4468  &2077    & \cmark &\cmark &  --    & EN      &W    &   &\cmark  &\cmark\\
	ICDAR17 \cite{icdar2017}                        & 2017    &7200  &9000   &18000  &68613 &--      & \cmark &\cmark &\cmark  & ML      & W   &   &\cmark  &\cmark \\
	TotalText \cite{ch2017total}                    & 2017    &1255  &300    &1555   &--    &11459   & \cmark &\cmark &\cmark  & EN      &W    &   &\cmark  &\cmark \\
	CTW-1500 \cite{CTW_1500_yuliang2017}            & 2017    &1000  &500    &1500   &--    & --     & \cmark &\cmark &\cmark  & CN      & W   &   &\cmark  &\cmark \\
	SynthText \cite{gupta2016synthetic}             & 2016    &800k  &--     &800k   &8M    & --     & \cmark &\cmark &\cmark  & EN      & W   &   &\cmark  &\cmark \\
	MJSynth \cite{jaderberg2014synthetic}           & 2014    &--    &--     &--     &8.9M  & --     & \cmark &\cmark &\cmark  & EN      & W   &   &  --    &\cmark \\
	
	\bottomrule
	
\end{tabular}
}
    \begin{tablenotes}
      \small
      \item[] Note: * This dataset has been considered for evaluation. H: Horizontal, MO: Multi-Oriented, Cu: Curved, EN: English, CN: Chinese, ML:Multi-Language, W: Word, C: Character, TL: Textline D: Detection, R: Recognition.
    \end{tablenotes}
\end{table*}






\subsection{Datasets}
\label{sec:OCRDatasets}
There exist several datasets that have been introduced for scene text detection and recognition \cite{lucas2003icdar,wang2010word,Mishra_12_IIIT,yao2012detecting,svtp_2013,karatzas2013icdar,cut80_2014,jaderberg2014synthetic,lin2014microsoft,karatzas2015icdar,gupta2016synthetic,icdar2017,ch2017total,CTW_1500_yuliang2017}. These datasets can be categorized into synthetic datasets that are used mainly for training purposes, such as \cite{Gupta16} and \cite{jaderberg2014synthetic}, and real-word datasets that have been utilized extensively for evaluating the performance of detection and evaluation schemes, such as \cite{yao2012detecting, karatzas2013icdar, karatzas2015icdar, lin2014microsoft, CTW_1500_yuliang2017, wang2010word}. 
Table \ref{tab:Text_dataset} compares some of the recent text detection and recognition datasets, and the rest of this section presents a summary of each of these datasets.



\subsubsection{MJSynth}
The \textit{MJSynth} \cite{jaderberg2014synthetic} dataset is a synthetic dataset that specifically designed for scene text recognition. Fig. \subref*{fig:MJ} shows some examples of this dataset.  This dataset includes about 8.9 million word-box gray synthesized images, which have been generated from the Google fonts and the images of ICDAR03 \cite{ICDRA2003} and SVT \cite{wang2010word} datasets. All the images in this dataset have annotated in word-level ground-truth and 90k common English words have been used for generating of these text images.  

\subsubsection{SynthText}
\begin{figure}[t]
\centering
\subfloat[MJ word boxes]{\label{fig:MJ} \includegraphics[width=0.45\linewidth]{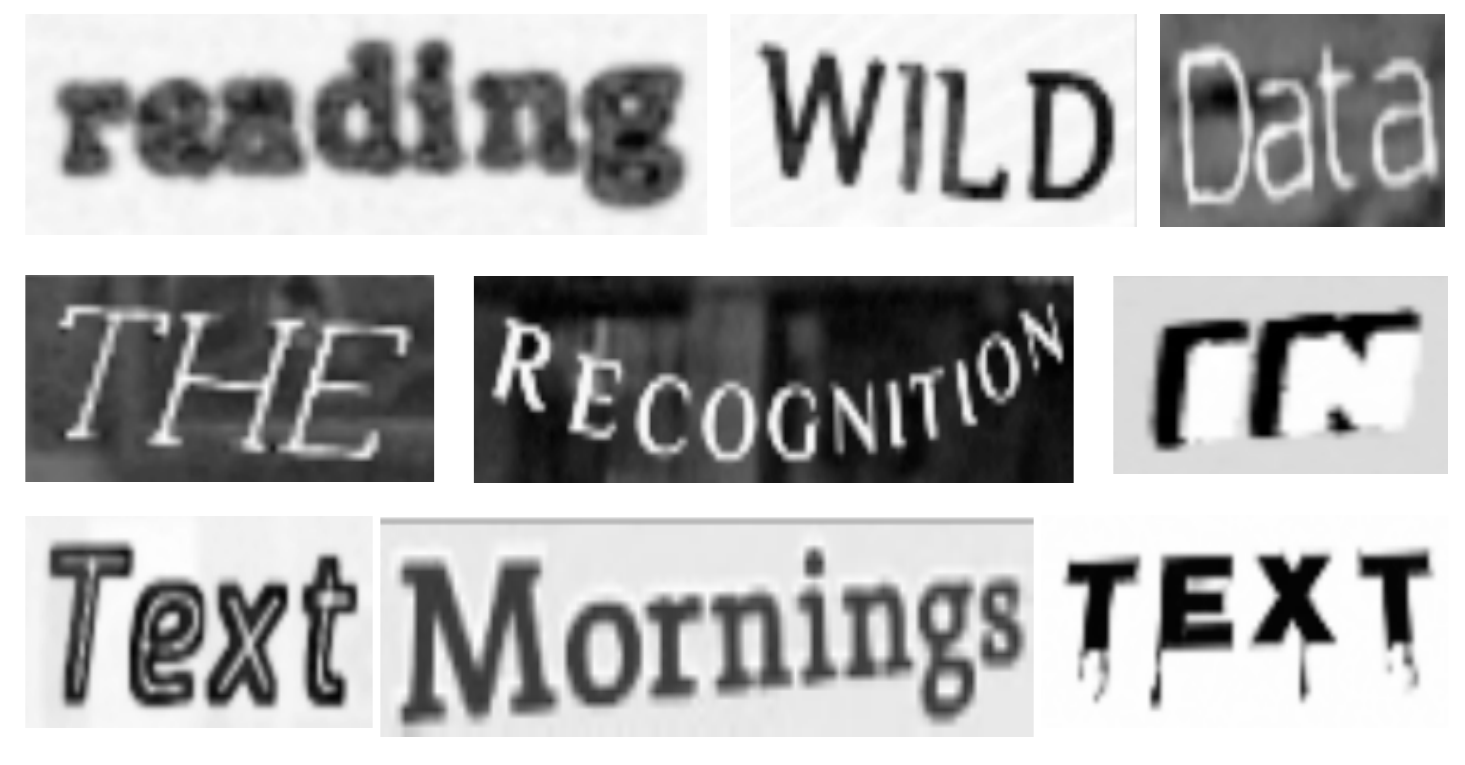}}
\centering
\subfloat[SynthText]{ \label{fig:SynthText} \includegraphics[width=0.45\linewidth]{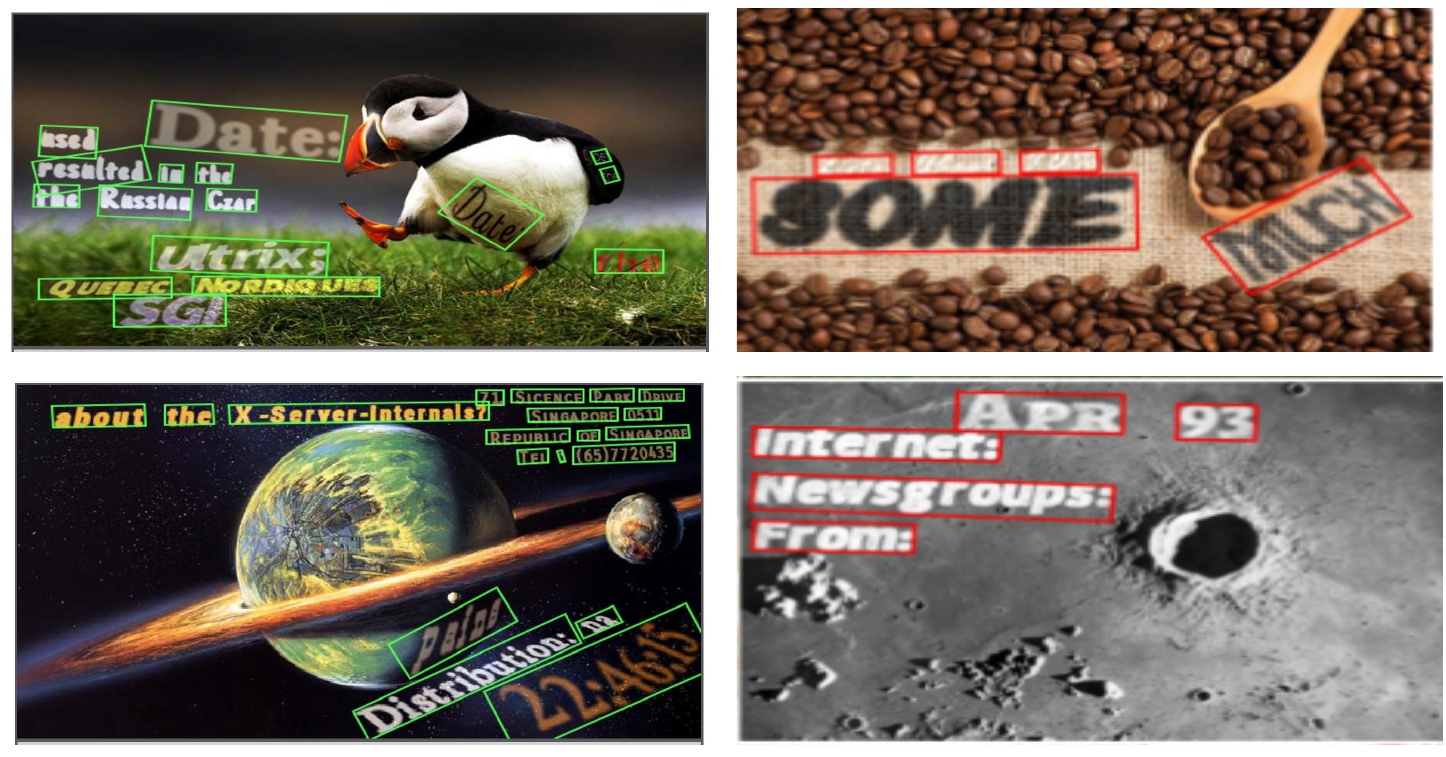}}
\caption{Sample images of synthetic datasets used for training in scene text detection and recognition \cite{gupta2016synthetic,jaderberg2014synthetic}.}
\label{fig:synthetic dataset}
\end{figure}

The \textit{SynthText in the Wild} dataset \cite{Gupta16} contains 858,750 synthetic scene images with 7,266,866 word-instances, and 28,971,487 characters. 
Most of the text instances in this dataset are multi-oriented and annotated with word and character-level rotated bounding boxes, as well as text sequences (see Fig. \subref*{fig:SynthText}). 
They are created by blending natural images with text rendered with different fonts, sizes, orientations and colors. 
This dataset has been originally designed for evaluating scene text detection \cite{Gupta16}, and leveraged in training several detection pipelines \cite{baek2019craft}. 
However, many recent text recognition methods \cite{cheng2018aon,shi2018aster,xie2019ACE,zhan2019esir,Wan2019_2DCTC} have also combined the cropped word images of the mentioned dataset with the MJSynth dataset \cite{jaderberg2014synthetic} for improving their recognition performance.

\subsubsection{ICDAR03}
The \emph{ICDAR03} dataset \cite{ICDRA2003} contains horizontal camera-captured scene text images. 
This dataset has been mainly used by recent text recognition methods, which consists of 1,156 and 110 text instances for training and testing, respectively. 
In this paper, we have used the same test images of \cite{baek2019STR} for evaluating the state-of-the-art text recognition methods.

\subsubsection{ICDAR13}
The \textit{ICDAR13} dataset \cite{karatzas2013icdar} includes images of horizontal text (the $i$th groundtruth annotation is represented by the indices of the top left corner associated with the width and height of a given bounding box as $G_i=[x_1^i,y_1^i,x_2^i,y_2^i]^{\top}$ that have been used in ICDAR 2013 competition and it is one of the benchmark datasets that used in many detection and recognition methods \cite{shi2016end,shi2016robust,STARNet_2016,zhou2017east,deng2018pixellink,liu2018fots,baek2019craft,pmtd_liu_2019,baek2019STR,luo_2019_moran}. The detection part of this dataset consists of 229 images for training and 233 images for testing, recognition part consists of 848 word-image for training and 1095 word-images for testing.
All text images of this dataset have good quality and text regions are typically centered in the images.

\subsubsection{ICDAR15}
The \textit{ICDAR15} dataset \cite{karatzas2015icdar} can be used for assessment of text detection or recognition schemes. The detection part has 1,500 images in total that consists of 1,000 training and 500 testing images for detection, and the recognition part consists of 4468 images for training and 2077 images for testing. 
This dataset includes text at the word-level of various orientation, and captured under different illumination and complex backgrounds conditions than that included in ICDAR13 dataset \cite{karatzas2013icdar}. However, most of the images in this dataset are captured for indoors environment.
In scene text detection, rectangular ground-truth used in the ICDAR13 \cite{karatzas2013icdar} are not adequate for the representation of multi-oriented text because: (1), they cause unnecessary overlap. (2), they can not precisely localize marginal text, and (3) they provide unnecessary noise of background \cite{liu2017deep}. Therefore to tackle the mentioned issues,  
the annotations of this dataset are represented using quadrilateral boxes (the $i$th groundtruth annotation can be expressed as  $G_i=[x_1^i,y_1^i,x_2^i,y_2^i,x_3^i,y_3^i,x_4^i,y_4^i]^{\top}$ for four corner vertices of the text).


\subsubsection{COCO-Text}
This dataset firstly was introduced in \cite{lin2014microsoft}, and so far, it is the largest and the most challenging text detection and recognition dataset. 
As shown in Table \ref{tab:Text_dataset}, the dataset includes 63,686 annotated images, where the dataset is partitioned into 43,686 training images, and 20,000 images for validation and testing. In this paper, we use the second version of this dataset, COCO-Text, as it contains 239,506 annotated text instances instead of 173,589 for the same set of images. 
As in ICDAR13, text regions in this dataset are annotated in a word-level using rectangle bounding boxes. 
The text instances of this dataset also are captured from different scenes, such as outdoor scenes, sports fields and grocery stores.
Unlike other datasets, COCO-Text dataset also contains images with low resolution, special characters, and partial occlusion.

%
\subsubsection{SVT}
 The \textit{Street View Text (SVT)} dataset \cite{wang2010word} consists of a collection of outdoor images with scene text of high variability of blurriness and/or resolutions, which were harvested using Google Street View.
 As shown in Table \ref{tab:Text_dataset}, this dataset includes 250 and 647 testing images for evaluation of detection and recognition tasks, respectively. 
 We utilize this dataset for assessing the state of the art recognition schemes. 
\subsubsection{SVT-P} 
The \textit{SVT - Perspective (SVT-P)} dataset \cite{svtp_2013} is specifically designed to evaluate recognition of perspective distorted scene text. 
It consists of 238 images with 645 cropped text instances collected from non-frontal angle snapshot in Google Street View, which many of the images are perspective distorted.

\subsubsection{IIIT 5K-words}
The \textit{IIIT 5K-words} dataset contains 5000 word-cropped scene images \cite{Mishra_12_IIIT},
that is used only for word-recognition tasks, and it is partitioned into 2000 and 3000 word images for training and testing tasks, respectively. 
In this paper, we use only the testing set for assessment. 


\subsubsection{CUT80}
The \textit{Curved Text} (CUT80) dataset is the first dataset that focuses on curved text images \cite{cut80_2014}. 
This dataset contains 80 full and 280 cropped word images for evaluation of text detection and text recognition algorithms, respectively. 
Although CUT80 dataset was originally designed for curved text detection, it has been widely used for scene text recognition \cite{cut80_2014}.

\subsection{Evaluation Metrics}
\label{sec:OCRmetrics}
The ICDAR standard evaluation metrics \cite{lucas2003icdar, wolf2006object, karatzas2013icdar,karatzas2015icdar} are the most commonly used protocols for performing quantitative comparison among the text detection techniques \cite{survey2015,survey2018}. 
\subsubsection{Detection}
In order to quantify the performance of a given text detector, as in \cite{zhou2017east,deng2018pixellink,baek2019craft,pmtd_liu_2019}, we utilize the Precision (P) and Recall (R) metrics that have been used in information retrieval field. In addition, we use the H-mean or F1-score that can be obtained as follows.
\begin{equation}
    \text{H-mean} = 2\times \frac{P \times R}{P+R}
    \label{eq:f_2003}
\end{equation}
where calculating the precision and recall are based on using the ICDAR15 intersection over union (IoU) metric \cite{karatzas2015icdar}, which is obtained for the $j$th ground-truth and $i$th detection bounding box as follow:
\begin{equation}
    \text{IoU} = \frac{\text{Area}(G_j \cap D_i)}{\text{Area}(G_j \cup D_i)}
\end{equation}
and a threshold of IoU $\geq 0.5$ is used for counting a correct detection.

%

\subsubsection{Recognition}
Word recognition accuracy (WRA) is a commonly used evaluation metric, due to its application in our daily life instead of character recognition accuracy, for assessing the text recognition schemes \cite{shi2016end,shi2016robust,STARNet_2016,shi2018aster,baek2019STR}. Given a set of cropped word images, WRA is defined as follow: 

\begin{equation}\label{eq:WRA}
    \text{WRA (\%)} = \frac{\text{No. of Correctly Recognized Words}}{\text{Total Number of Words}} \times 100 
\end{equation}



\subsection{Evaluation of Text Detection Techniques}
\label{sec:EvalDetectionTechniques}
\subsubsection{Quantitative Results}
To evaluate the generalization ability of detection methods, we compare the detection performance on ICDAR13 \cite{karatzas2013icdar}, ICDAR15 \cite{karatzas2015icdar} and COCO-Text \cite{veit2016coco} datasets. 
Table \ref{tab:det_new} illustrates the detection performance of the selected state of the art text detection methods, namely, 
PMTD \cite{pmtd_liu_2019}, 
CRAFT \cite{baek2019craft}, 
PSENet \cite{PSENet_wang2019},
MB \cite{liu2019_BOX},
PAN \cite{wang2019_PAN},
Pixellink \cite{deng2018pixellink} and
EAST \cite{zhou2017east}.
From this table, although the ICDAR13 dataset includes less challenging conditions than that included in the ICDAR15 dataset, the detection performances of all the methods in consideration have been decreased on this dataset. 
Comparing the same method performance on ICDAR15 and ICDAR13, PMTD offered a minimum performance decline of $\sim$ 0.60\% in H-mean, while Pixellink that ranked the second-best on ICDAR15 had the worst H-mean value on ICDAR13 with decline of $\sim$ 20.00\%.
Further, all methods experienced a significant decrease in detection performance when tested on COCO-Text dataset, which indicate that these models do not yet provide a generalization capability on different challenging datasets.

\begin{table*}[]
    \centering
    \caption{Quantitative comparison among some of the recent text detection methods on ICDAR13 \cite{karatzas2013icdar}, ICDAR15 \cite{karatzas2015icdar} and COCO-Text \cite{veit2016coco} datasets using precision (P), recall (R) and H-mean.}
    \begin{tabular}{l|ccc|ccc|ccc}
    \toprule
         \multirow{2}{*}{Method} & \multicolumn{3}{c|}{ICDAR13}& \multicolumn{3}{c|}{ICDAR15}& \multicolumn{3}{c}{COCO-Text}\\ \cline{2-10}
& P       & R       & H-mean         & P     & R     & H-mean     & P     & R     & H-mean     \\ \toprule
EAST \cite{zhou2017east}              & \underline{84.86\%}   & 74.24\%   & \underline{79.20\%}     & 84.64\% & 77.22\% & 80.76\% & 55.48\% & 32.89\% & 41.30\% \\
Pixellink \cite{deng2018pixellink}    & 62.21\% & 62.55\%  & 62.38\%   & 82.89\% & 81.65\% & \underline{82.27\%} & \underline{61.08\%} & 33.45\% & 43.22\%   \\
PAN \cite{wang2019_PAN}               & 83.83\%   & 69.13\%   & 75.77\%    & \underline{85.95\%} & 73.66\% & 79.33\% & 59.07\% & 43.64\% & 50.21\% \\
MB \cite{liu2019_BOX}                 & 72.64\%   & 60.36\%   & 65.93\%     & 85.75\% & 76.50\%  & 80.86\% & 55.98\% & 48.45\% & 51.94\%  \\
PSENet \cite{PSENet_wang2019}         & 81.04\%   & 62.46\%   & 70.55\%    & 84.69\% & 77.51\% & 80.94\% & 60.58\% & 49.39\% & 54.42\%  \\
CRAFT \cite{baek2019craft}            & 72.77\%   & \underline{77.62\%}   & {75.12\%}     & 82.20\%  & \underline{77.85\%} & 79.97\% & 56.73\% & \underline{55.99\%} & \underline{56.36\%} \\
PMTD \cite{pmtd_liu_2019}             & \textbf{92.49\%}   & \textbf{83.29\%}   & \textbf{87.65\%}     & \textbf{92.37\%} &\textbf{84.59\%} & \textbf{88.31\%} & \textbf{61.37\%} & \textbf{59.46\%} & \textbf{60.40\%}\\
    \bottomrule
    \end{tabular}
    \label{tab:det_new}
\end{table*}

\subsubsection{Qualitative Results}
\begin{figure*}
    \centering
    \includegraphics[width=\linewidth]{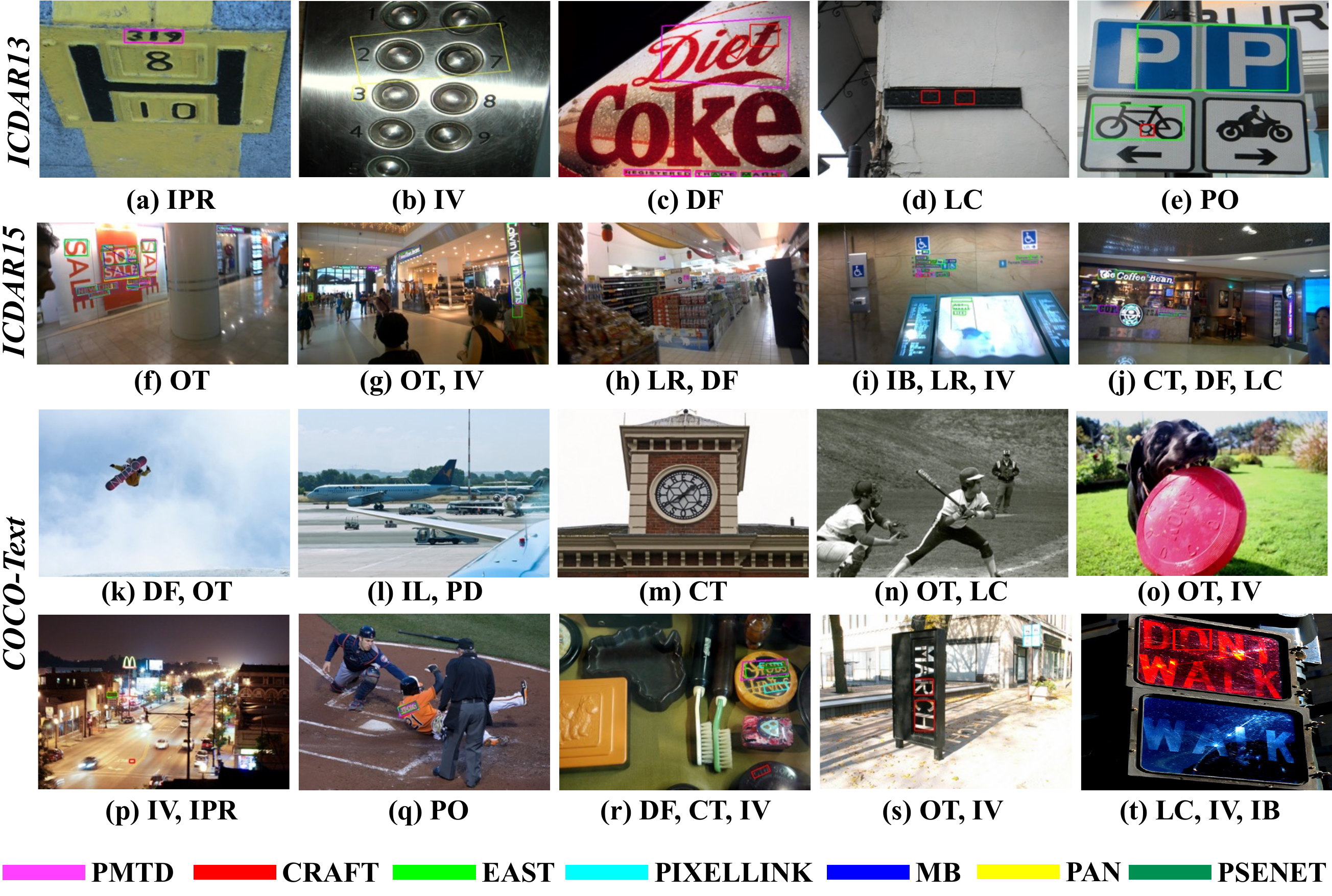}
    \caption{Qualitative detection results comparison among CRAFT \cite{baek2019craft}, PMTD \cite{pmtd_liu_2019}, MB \cite{liu2019_BOX}, PixelLink \cite{deng2018pixellink}, PAN \cite{wang2019_PAN}, EAST \cite{zhou2017east} and PSENET \cite{PSENet_wang2019} on some challenging examples, where 
    PO: Partial Occlusion, DF: Difficult Fonts, LC: Low Contrast, IV: Illumination Variation, IB: Image Blurriness, LR: Low Resolution, PD: Perspective Distortion, IPR: in-plane-rotation, OT: Oriented Text, and CT: Curved Text. Note: since we used pre-trained models on ICDAR15 dataset for all the methods in comparison, the results of some methods may different from those reported in the original papers.}
    \label{fig:qual_det}
\end{figure*}

Figure \ref{fig:qual_det} illustrates sample detection results for the considered methods \cite{pmtd_liu_2019,baek2019craft,PSENet_wang2019,liu2019_BOX,wang2019_PAN,deng2018pixellink,zhou2017east} on some challenging scenarios from ICDAR13, ICDAR15 and COCO-Text datasets. 
Even though, the best text detectors, PMTD and CRAFT detectors, offer better robustness in detecting text under various orientation and partial occlusion levels,
these detection results illustrate that the performances of these methods are still far from perfect. Especially, when text instances are affected by challenging cases like text of difficult fonts, colors, backgrounds, and illumination variation and in-plane rotation, or a combination of challenges.
Now we categorize  the common difficulties in scene text detection as follows:
\paragraph{Diverse Resolutions and Orientations}
Unlike the detection tasks, such as detection of pedestrians \cite{dollar2011pedestrian} or cars \cite{du2017car,ammour2017deep}, text in the wild usually appears on a wider variety of resolutions and orientations, which can easily leads to inadequate detection performances \cite{survey2018,survey2019}. 
For instance on ICDAR13 dataset, as can be seen from the results 
in Fig. \ref{fig:qual_det} (a), all the methods failed to detect the low and high resolutions text using the default parameters of these detectors. The same conclusion can be drawn from the results on 
Figures \ref{fig:qual_det} (h) and (q) 
on ICDAR15 and COCO-Text datasets, respectively.
As well as, this conclusion can also be confirmed from the distribution of word height in pixels on the considered datasets as shown in 
Fig. \ref{fig:height_prob}.
Although the considered detection models have focused on handling multi-orientated text, they still lake the robustness in tackling this challenge as well as facing difficulty in detecting text subjected to in-plan rotation or high curvature. For example, the low detection performance noted as can be seen in Figures \ref{fig:qual_det} (a), (j), and (p) 
on ICDAR13, ICDAR15 and COCO-Text, respectively.
\paragraph{Occlusions} 
Similar to other detection tasks, text can be occluded by itself or other objects, such as text or object superimposed on a text instance as shown in Fig. \ref{fig:qual_det}. 
Thus, it is expected for text-detection algorithms to at least detect partially occluded text. 
However, as we can see from the sample results in Figures \ref{fig:qual_det} (e) and (f),
the studied methods failed in detection of text mainly due to the partial occluded effect.

\paragraph{Degraded Image Quality} 
Text images captured in the wild are usually affected by various illumination conditions (as in Figures \ref{fig:qual_det} (b) and (d)), motion blurriness (as in Figures \ref{fig:qual_det} (g) and (h)), and low contrast text (as in Figures \ref{fig:qual_det} (o) and (t)). 
As we can see from Fig. \ref{fig:qual_det}, the studied methods perform weakly on these type of images. 
This is due to existing text detection techniques have not tackled explicitly these challenges. 

\subsubsection{Discussion}
\begin{figure*} 
    \centering
     \setlength{\tabcolsep}{2 pt}
          \scalebox{1}{
    \begin{tabular}{ccc}
    \resizebox{0.33\linewidth}{!}{\includegraphics*{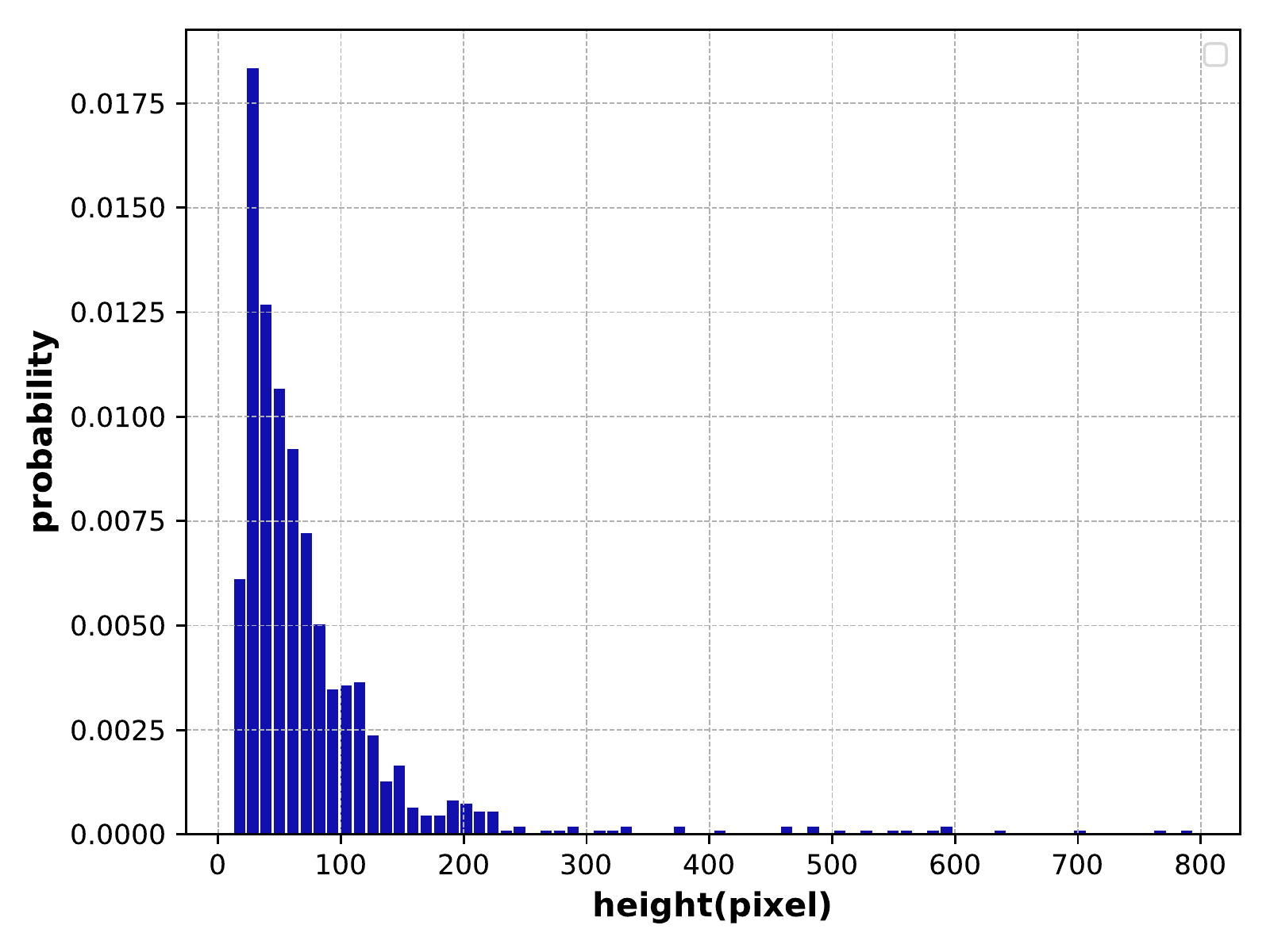}}&  
    \resizebox{0.33\linewidth}{!}{\includegraphics*{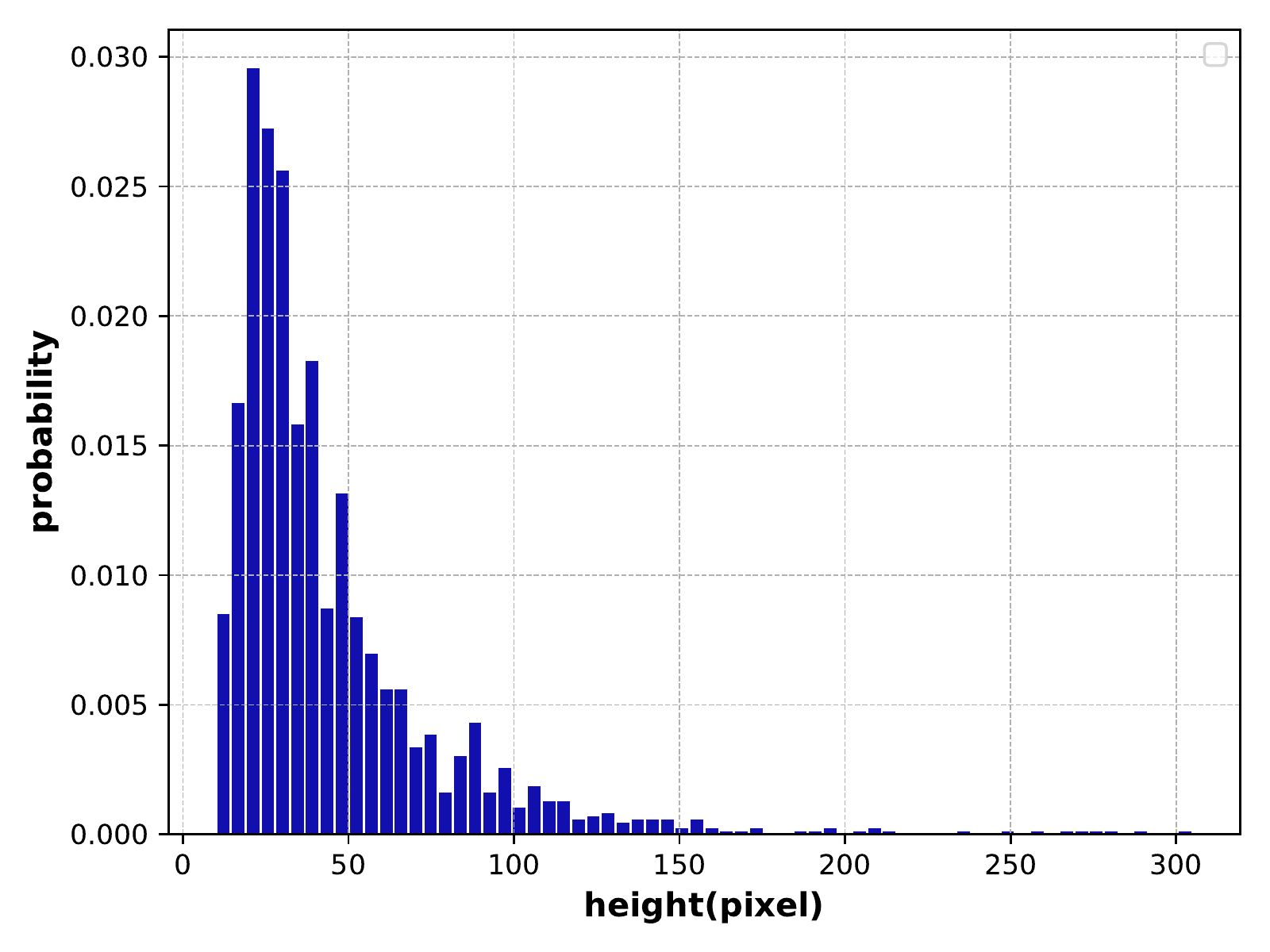}}&
    \resizebox{0.33\linewidth}{!}{\includegraphics*{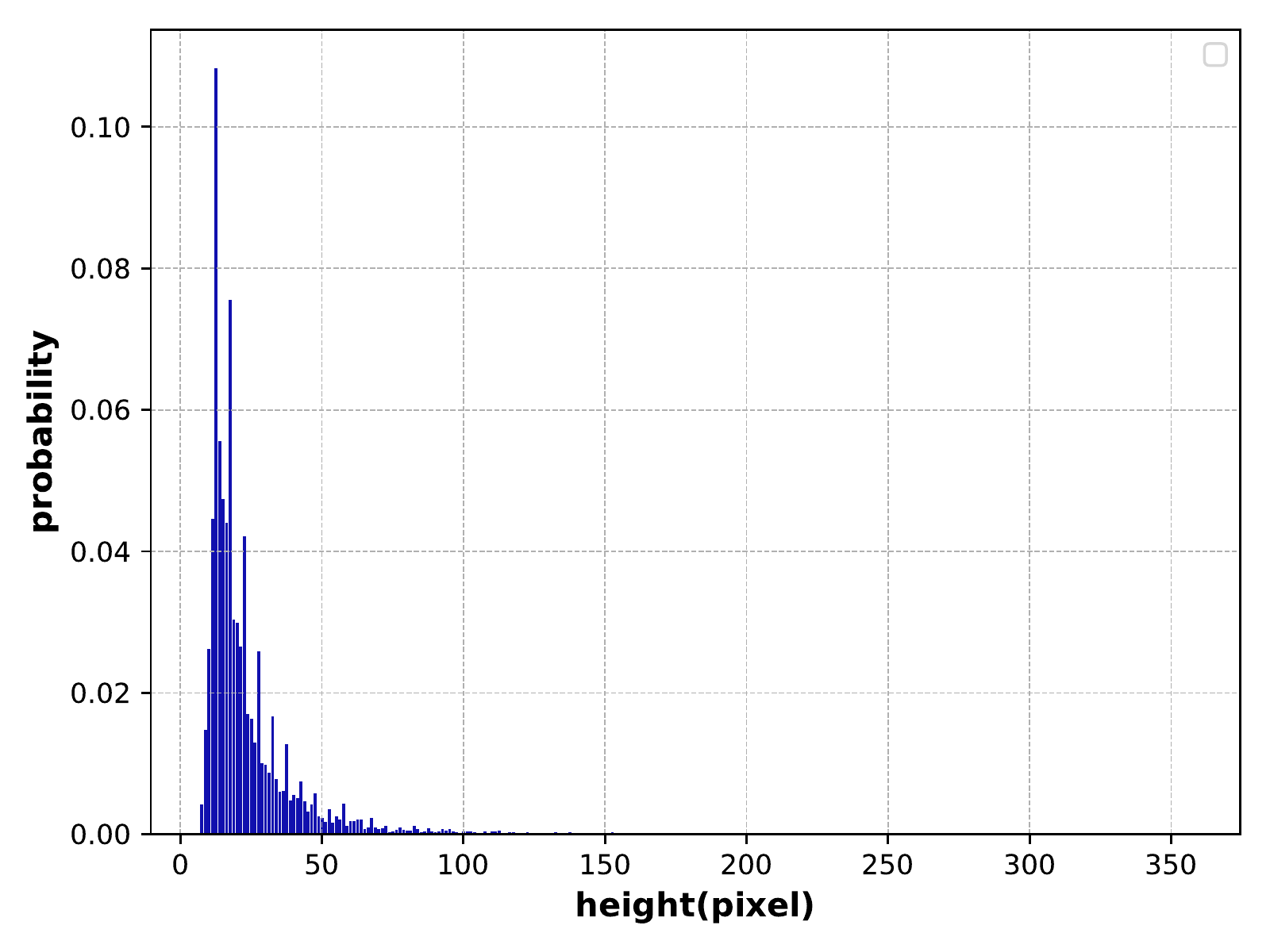}}\\
    (a) ICDAR13 & (b) ICDAR15 &  (c) COCO-Text   
    \end{tabular}}
    \caption{Distribution of word height in pixels
    computed on the test set of (a) ICDAR13, (b) ICDAR15 and (c) COCO-Text detection datasets.}
    \label{fig:height_prob}
\end{figure*}

\begin{figure*} [!t]
    \centering
     \setlength{\tabcolsep}{2 pt}
    \begin{tabular}{ccc}
    \resizebox{0.33\linewidth}{!}{\includegraphics*{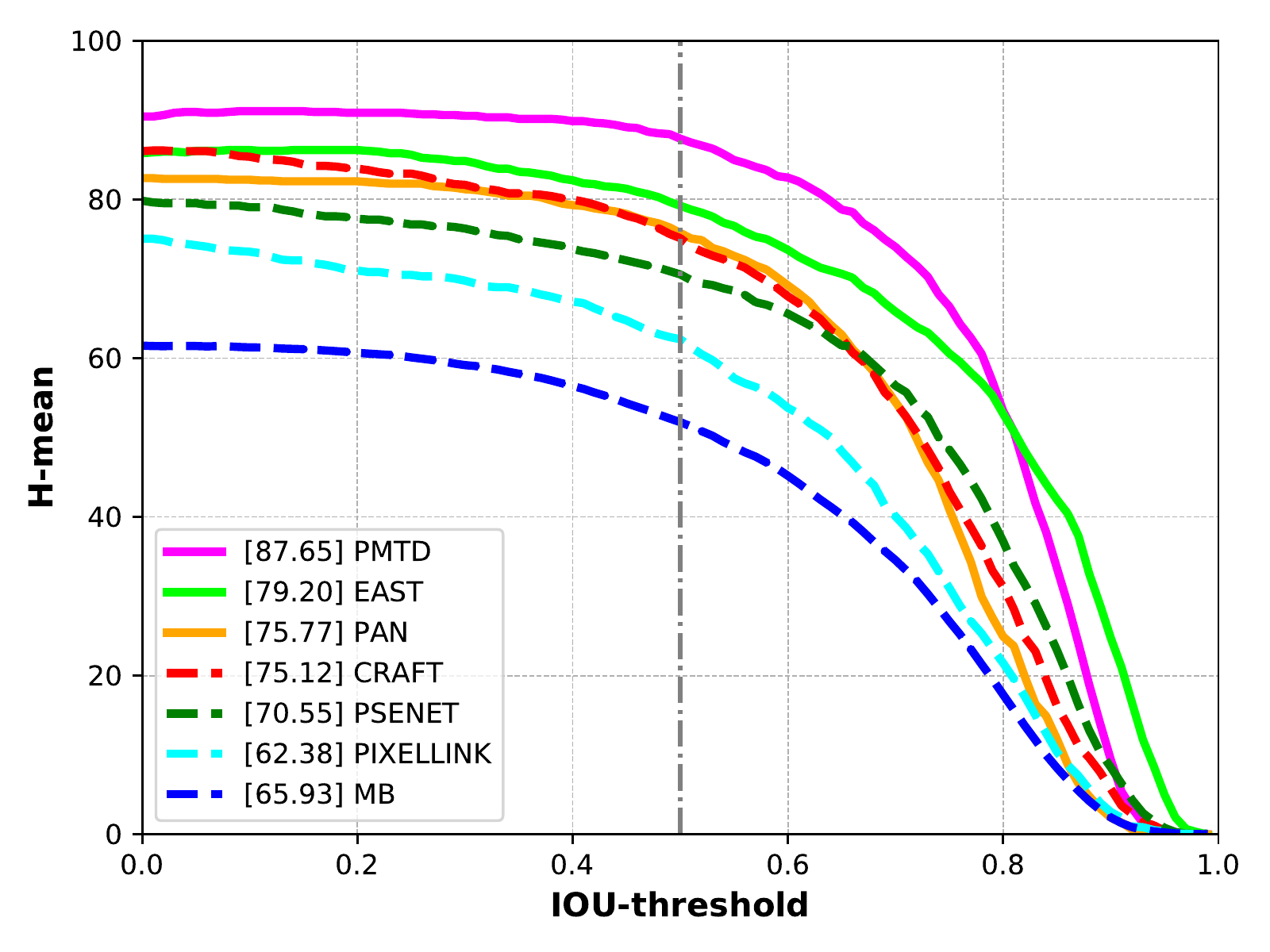}}&  
    \resizebox{0.33\linewidth}{!}{\includegraphics*{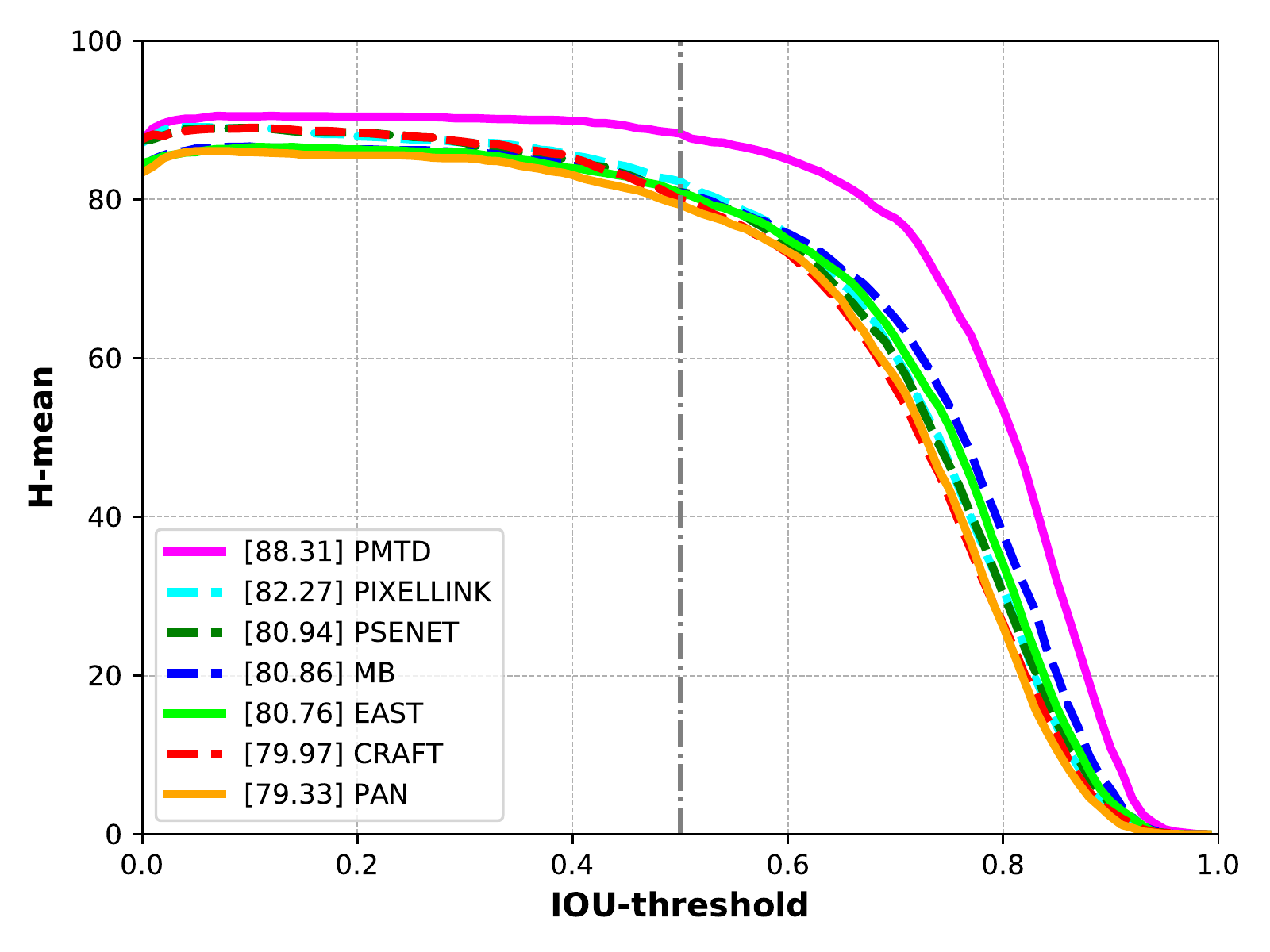}}&
    \resizebox{0.33\linewidth}{!}{\includegraphics*{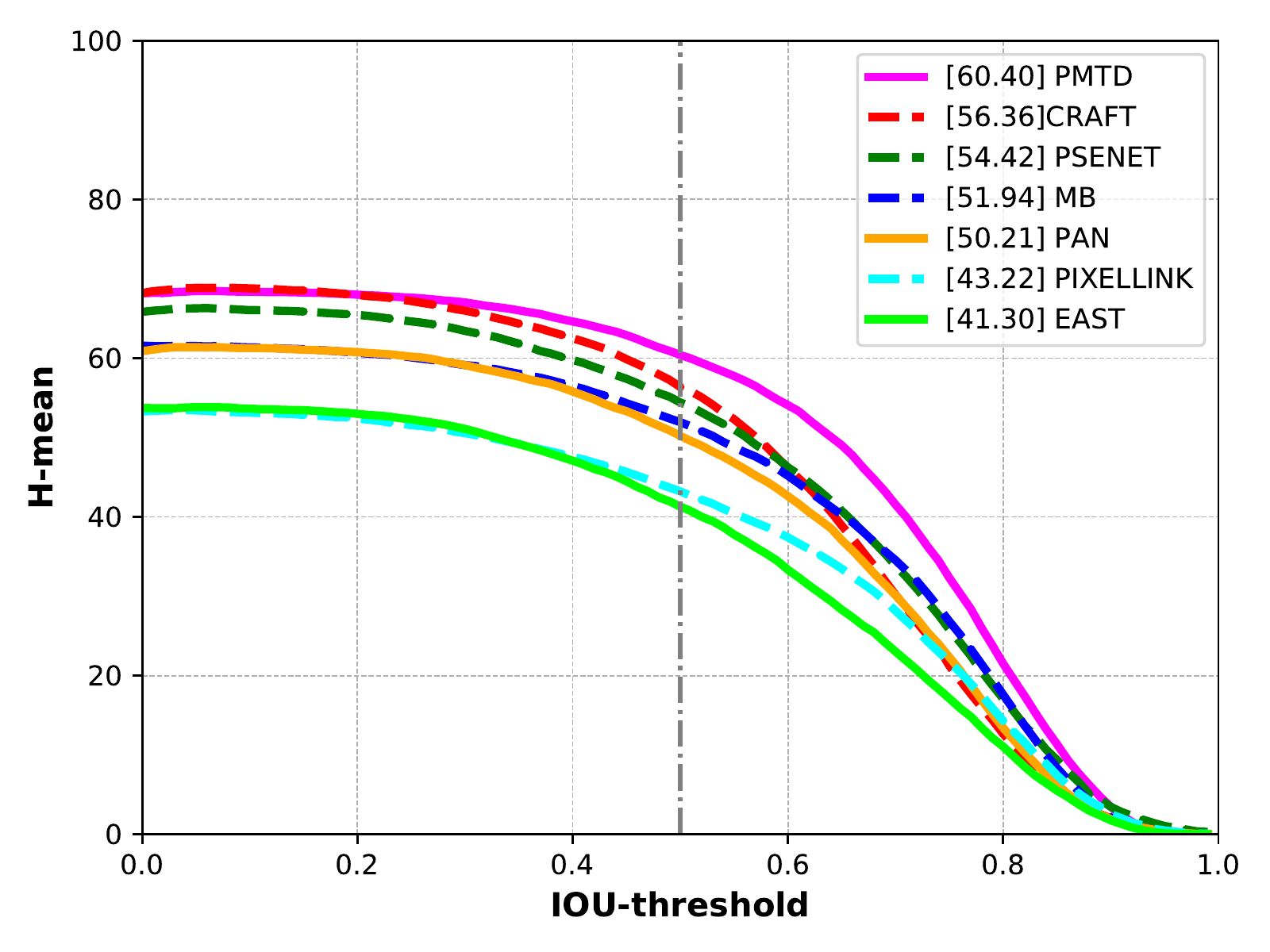}}\\
    (a) ICDAR13 & (b) ICDAR15 &  (c) COCO-Text   
    \end{tabular}
    \caption{Evaluation of the text detection performance for CRAFT \cite{baek2019craft}, PMTD \cite{pmtd_liu_2019}, MB \cite{liu2019_BOX}, PixelLink \cite{deng2018pixellink}, PAN \cite{wang2019_PAN}, EAST \cite{zhou2017east} and PSENET \cite{PSENet_wang2019} using H-mean versus IoU $\in [0, 1]$
    computed on (a) ICDAR13 \cite{karatzas2013icdar}, (b) ICDAR15 \cite{karatzas2015icdar}, and (c) COCO-Text \cite{veit2016coco} datasets.}
    \label{fig:iou_det}
\end{figure*}

In this section, we present an evaluation of the mentioned detection methods with respect to the robustness and speed.
\paragraph{Detection Robustness} 
As we can see from Fig. \ref{fig:height_prob}, most of the target words existed in the three target scene text detection datasets are of low resolutions, which makes the text detection task more challenging.
To compare the robustness of the detectors under the various IoU values, Fig. \ref{fig:iou_det} illustrates the H-mean computed at IoU $\in [0, 1]$ for each of the studied methods.   
From this figure it can be noted that increasing the IoU $>0.5$ causes rapidly reducing the H-mean values achieved by the detectors on all the three datasets, which indicates that the considered schemes are not offering adequate overlap ratios, i.e., IoU, at higher threshold values.

More specifically, on ICDAR13 dataset (Figure \ref{fig:iou_det}a) EAST \cite{zhou2017east} detector outperforms the PMTD \cite{pmtd_liu_2019} for IoU $>0.8$; 
this can be attributed to that EAST detector uses a multi-channel FCN network that allows detecting more accurately text instances at different scales that are abundant in ICDAR13 dataset.
Further, Pixellink \cite{deng2018pixellink} that ranked second on ICDAR15 has the worst detection performance on ICDAR13. 
This poor performance is also can be seen in challenging cases of the qualitative results in Figure \ref{fig:qual_det}. 
For COCO-Text \cite{veit2016coco} dataset, 
all methods offer poor H-mean performance on this dataset (Fig. \ref{fig:iou_det}c). In addition, generally, the H-means of the detectors are declined to the half, from $\sim$ 60\% to below of $\sim$ 30\%, for IoU $\geq0.7$. 


In summary, PMTD and CRAFT show better H-mean values than that of EAST and Pixellink for IoU $<0.7$. 
Since CRAFT is character-based methods, it performed better in localizing difficult-font words with individual characters. Moreover, it can better handles text with different size due to the its property of localizing individual characters, not the whole word detection has been used in the majority of other methods.
Since COCO-Text and ICDAR15 datasets contain multi-oriented and curved text, we can see from Fig. \ref{fig:iou_det} that PMTD shows robustness to imprecise detection compared to other methods for precise detection for IoU $>0.7$, which  means this method can predict better arbitrary shape of text. This is also obvious in challenging cases like curved text, difficult fonts with various orientation, and in-plane rotated text of Fig. \ref{fig:qual_det}.

\begin{figure}
    \centering
    \includegraphics[width=\linewidth]{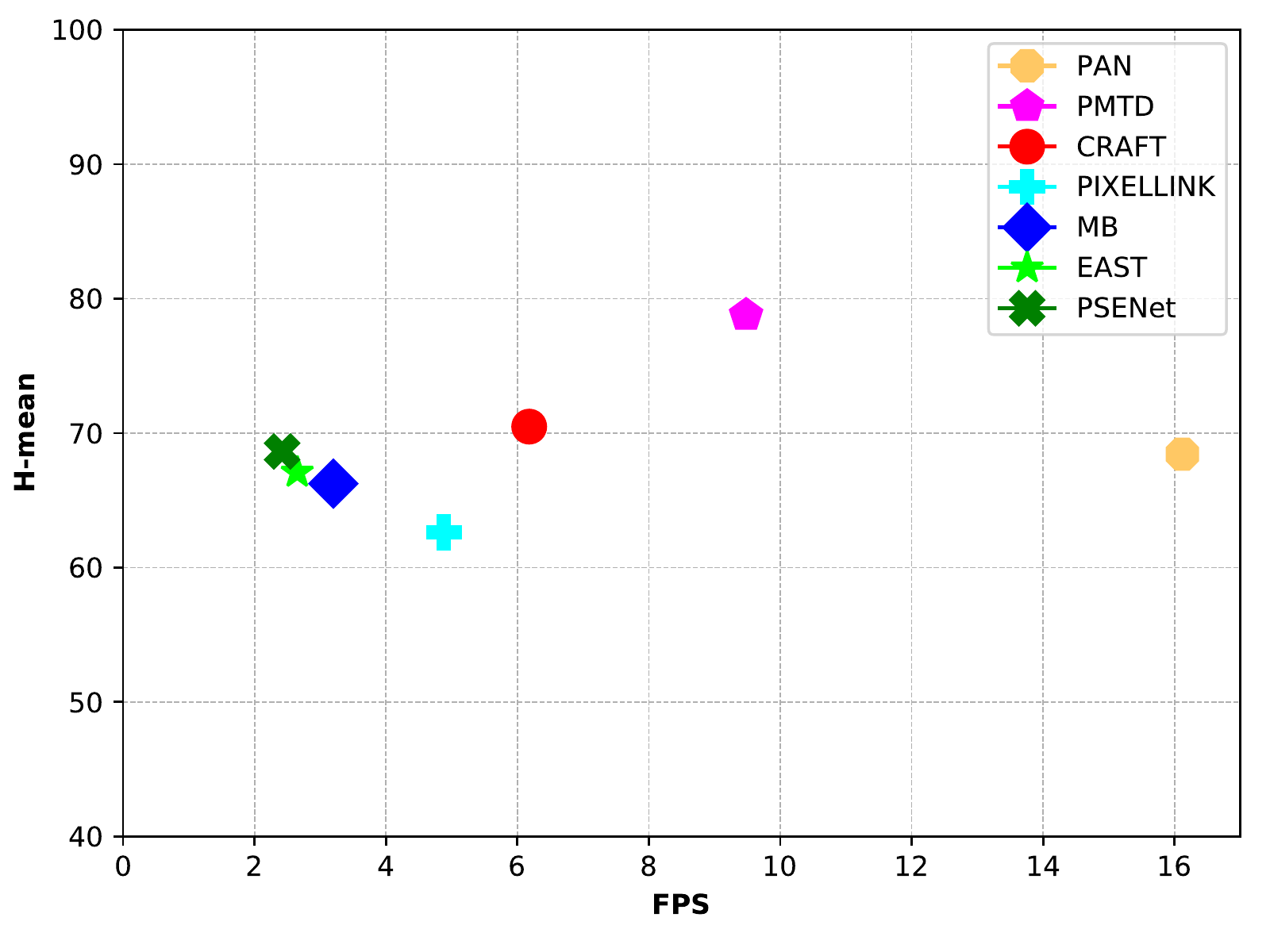}
    \caption{Average H-mean versus frames per second (FPS) computed on ICDAR13 \cite{karatzas2013icdar}, ICDAR15 \cite{karatzas2015icdar}, and COCO-Text \cite{veit2016coco} detection datasets using IoU $\geq 0.5$.}
    \label{fig:fps_det}
\end{figure}

\paragraph{Detection Speed} To evaluate the speed of detection methods, Fig. \ref{fig:fps_det} plots H-mean versus the speed in frame per second (FPS) for each detection algorithm for IoU $\geq 0.5$. The slowest and fastest detectors are PSENet \cite{PSENet_wang2019} and PAN \cite{wang2019_PAN}, respectively. PMTD \cite{pmtd_liu_2019} achieved the second fastest detector with the best H-mean. 
PAN utilizes a segmentation-head with low computational cost using a light-weight model, ResNet \cite{ResNet_He2015L} with 18 layers (ResNet18), as a backbone and a few stages for post-processing that result in an efficient detector. 
On the other hand, PSENet uses multiple scales for predicting of text instances using a deeper (ResNet with 50 layer) model as a backbone, which cause it to be slow during the test time.

\subsection{Evaluation of Text Recognition Techniques}
\label{sec:EvalRecognitionTechniques}
\subsubsection{Quantitative Results}
In this section, we compare the selected scene text recognition methods \cite{shi2016end,shi2016robust,rosetta2018,STARNet_2016,shi2018aster} in terms of the word recognition accuracy (WRA) defined in \eqref{eq:WRA} on datasets with regular \cite{Mishra_12_IIIT,ICDRA2003,karatzas2013icdar} and irregular \cite{karatzas2015icdar,svtp_2013,cut80_2014,veit2016coco} text, and Table \ref{tab:rec_quan} summarizes these quantitative results. 
It can be seen from this table that all methods have generally achieved higher WRA values on datasets with regular-text \cite{Mishra_12_IIIT,ICDRA2003,karatzas2013icdar} than that achieved on datasets with irregular-text \cite{karatzas2015icdar,svtp_2013,cut80_2014,veit2016coco}. 
Furthermore, methods that contain a rectification module in their feature extraction stage  for spatially transforming text images, namely, ASTER \cite{shi2018aster}, CLOVA \cite{baek2019STR} and STAR-Net \cite{STARNet_2016}, have been able to perform better on datasets with irregular-text. 
In addition, attention-based methods, ASTER \cite{shi2018aster} and CLOVA \cite{baek2019STR}, outperformed the CTC-based methods, CRNN \cite{shi2016end}, STAR-Net \cite{STARNet_2016} and ROSETTA \cite{rosetta2018} because attention methods better handles the alignment problem in irregular text compared to CTC-based methods. 

It is worth noting that despite the studied text recognition methods have used only synthetic images for training, as can be seen from Table \ref{tab:recognition_comp}, they have been able to handle recognizing text in the wild images. 
However, for COCO-Text dataset, each of the methods has achieved a much lower WRA values than that obtained on the other datasets, this is can be attributed to the more complex situations exist in this dataset that the studied models are not able to fully encounter. 
In the next section, we will highlight on the challenges that most of the state-of-the-art scene text recognition schemes are currently facing.

\begin{table*}[t]
\caption{Comparing some of the recent text recognition techniques using WRA on IIIT5k \cite{Mishra_12_IIIT}, SVT \cite{wang2010word} , ICDAR03 \cite{ICDRA2003}, ICDAR13 \cite{karatzas2013icdar}, ICDAR15 \cite{karatzas2015icdar}, SVT-P \cite{svtp_2013}, CUT80 \cite{cut80_2014} and COCO-Text \cite{veit2016coco} datasets.}
\centering
\begin{tabular}{l|llllllll}
\toprule
Method                                  & IIIT5k & SVT   & ICDAR03  & ICDAR13  & ICDAR15  & SVT-P & CUT80 & COCO-Text \\ \toprule
CRNN \cite{shi2016end}                  & 82.73\%  & 82.38\% & 93.08\% & 89.26\% & 65.87\% & 70.85\% & 62.72\% & 48.92\%    \\
RARE \cite{shi2016robust}               & 83.83\% & 82.84\% & 92.38\% & 88.28\% & 68.63\% & 71.16\% & 66.89\% & 54.01\%    \\
ROSETTA \cite{rosetta2018}              & 83.96\% & 83.62\% & 92.04\% & 89.16\% & 67.64\% & 74.26\% & 67.25\% & 49.61\%    \\
STAR-Net \cite{STARNet_2016}            & 86.20\% & 86.09\% & 94.35\% & 90.64\% & 72.48\% & 76.59\% & 71.78\% & 55.39\%    \\
CLOVA \cite{baek2019STR}                & \underline{87.40\%} & \underline{87.01\%} & \textbf{94.69\%} & \textbf{92.02\%} & \textbf{75.23\%} & \underline{80.00\%} & \underline{74.21\%} & \underline{57.32\%}    \\
ASTER \cite{shi2018aster}               & \textbf{93.20\%}  & \textbf{89.20\%} & \underline{92.20\%} & \underline{90.90\%} & \underline{74.40\%} & \textbf{80.90\%} & \textbf{81.90\%} &\textbf{60.70\%}    \\
\bottomrule
\end{tabular}
\begin{center}
\begin{tablenotes}      
\centering
      \small
      \item[] Note: Best and second best methods are highlighted by bold and underline, respectively.
\end{tablenotes}
\end{center}
\label{tab:rec_quan}
\end{table*}

\subsubsection{Qualitative Results}\label{sec:RecQualitativeResults}
\begin{figure*}
    \centering
    \includegraphics[width=1.03\linewidth]{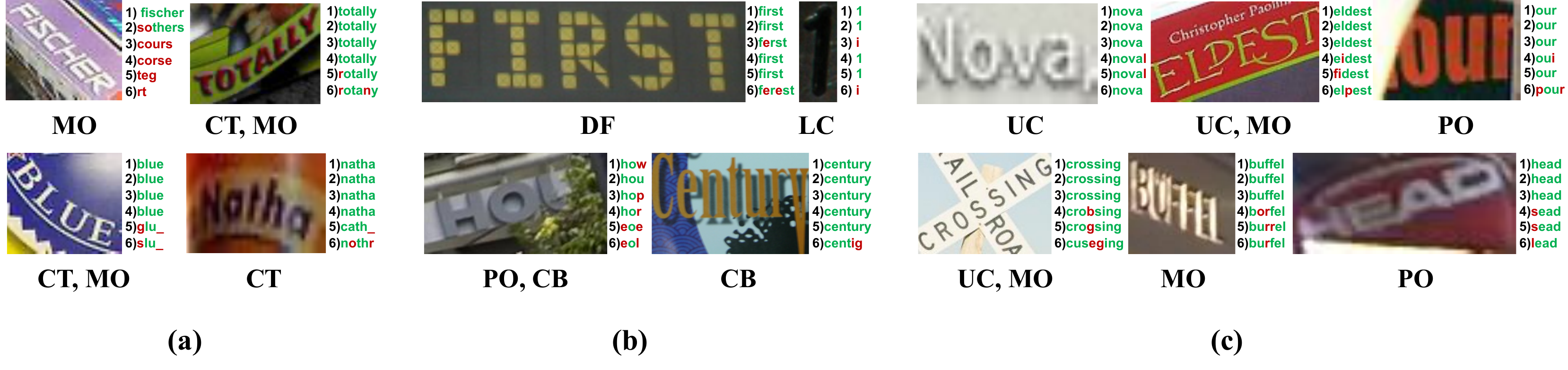}
    \caption{Qualitative results for challenging examples of scene text recognition: 
    (a) Multi-oriented and curved text, 
    (b) complex backgrounds and fonts, and
    (c) text with partial occlusions, sizes, colors or fonts.
    Note: The original target word images and their corresponding recognition output are on the left and right hand sides of every sample results, respectively, where the numbers denote: 
    1) ASTER \cite{shi2018aster}, 
    2) CLOVA \cite{baek2019STR},
    3) RARE \cite{shi2016robust},
    4) STAR-Net \cite{STARNet_2016},
    5) ROSETTA \cite{rosetta2018} and  
    6) CRNN \cite{shi2016end}, and the resulted characters highlighted by green and red  denote correctly and wrongly predicted characters, respectively, where
     MO: Multi-Oriented, VT: Vertical Text, CT: Curved Text, DF: Difficult Font, LC: Low Contrast, CB: Complex Background, PO: Partial Occlusion and UC: Unrelated Characters.
     It should be noted that the results of some methods may different from that reported in the original papers, because we used pre-trained models on MJSynth \cite{jaderberg2014synthetic} and SynthText \cite{gupta2016synthetic} datasets for each specific method. 
}
    \label{fig:qual_rec}
\end{figure*}

In this section, we present a qualitative comparison among the considered text recognition schemes, as well as we conduct an investigation on challenging scenarios that still causing partial or complete failures to the existing techniques.
Fig. \ref{fig:qual_rec} highlights on a sample qualitative performances for the considered text recognition methods on ICDAR13, ICDAR15 and COCO-Text datasets.
As shown in Fig. \ref{fig:qual_rec}(a), methods in \cite{shi2016robust,STARNet_2016,shi2018aster,baek2019STR} performed well on multi-oriented and curved text because these methods adopt TPS as a rectification module in their pipeline that allows rectifying irregular text into a standard format and thus the subsequent CNN can better extract features from this normalized images.
Fig. \ref{fig:qual_rec}(b) illustrates text subject to complex backgrounds or unseen fonts. In these cases, the methods that utilized ResNet, which has a deep CNN architecture, as the backbone for feature extraction, 
as in ASTER, CLOVA, STAR-Net, and ROSETTA, outperformed the methods that used VGG as in RARE and CRNN.

\begin{figure*}
    \centering
    \includegraphics[width=\linewidth]{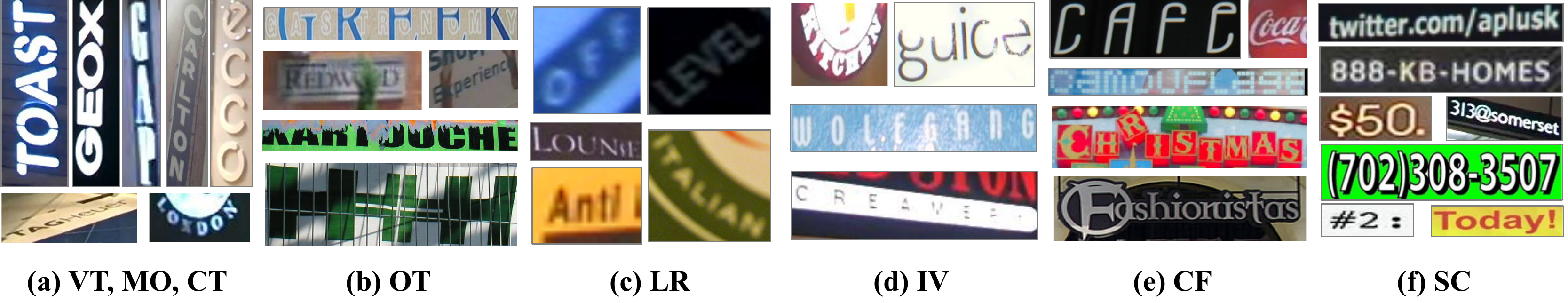}
    \caption{Illustration for challenging cases on scene text recognition that still cause recognition failure, where
    a) vertical text (VT), multi-oriented (MO) text and curved text (CT),
    b) occluded text (OT),
    c) low resolution (LR),
    d) illumination variation (IV),
    e) complex font (CF), and
    f) special characters (SC).}
    \label{fig:qual_fail}
\end{figure*}

Although the considered state-of-the-art methods have shown the ability to recognize text under some challenging examples, as illustrated in Fig. \ref{fig:qual_rec}(c), there are still various challenging cases that these methods do not explicitly handle, such as recognizing text of calligraphic fonts and text subject to heavy occlusion, low resolution and illumination variation.
Fig. \ref{fig:qual_fail} shows some challenging cases from the considered  benchmark datasets that all of the studied recognition methods failed to handle. 
In the rest of this section, we will analyze these failure cases and suggest future works to tackle these challenges.

\paragraph{Oriented Text}
Existing state-of-the-art scene text recognition methods have focused more on recognizing horizontal \cite{shi2016end,lee2016recursive}, multi-oriented  \cite{cheng2018aon,cheng2017fan} and curved text \cite{shi2016robust,shi2018aster,baek2019STR,zhan2019esir,Wan2019_2DCTC,luo_2019_moran}, which leverage a spatial rectification module \cite{Jaderberg_STN,zhan2019esir,luo_2019_moran} and typically use sequence-to-sequence models designed for reading text.
Despite these attempts to solve recognizing text of arbitrary orientation, there are still types of orientated text in the wild images that these methods could not tackle,
such as highly curved text, in-plane rotation text, vertical text, and text stacked from bottom-to-top and top-to-down demonstrated in Fig. \ref{fig:qual_fail}(a). 
%
%
%
In addition, since horizontal and vertical text have different characteristics, researchers have recently attempt \cite{choi2018_vertical,ling2018_vertical} to design techniques for recognizing both types of text in a unified framework. 
Therefore, further research would be required to construct models that are able to simultaneously recognizing of different orientations. 

\paragraph{Occluded Text}
Although existing attention-based methods \cite{shi2016robust,shi2018aster,baek2019STR} have shown the capability of recognizing text subject to partial occlusion, their performance declined on recognizing text with heavy occlusion, as shown in Fig. \ref{fig:qual_fail}(b).
This is because the current methods do not extensively exploit contextual information to overcome occlusion. Thus, future researches may consider superior language models \cite{BERT_devlin2018} to utilize context maximally for predicting the invisible characters due to occluded text.

\paragraph{Degraded Image Quality}
It can be noted also that the state-of-the-art text recognition methods, as in \cite{shi2016end,STARNet_2016,shi2016robust,rosetta2018,shi2018aster,baek2019STR}, did not specifically overcome the effect of degraded image quality, such as low resolution and illumination variation, on the recognition accuracy. 
Thus, inadequate recognition performance can be observed from the sample qualitative results in Figures \ref{fig:qual_fail}(c) and \ref{fig:qual_fail}(d). 
As a suggested future work, it is important to study how image enhancement techniques, such as image super-resolution \cite{lyn2020image}, image denoising \cite{anwar2019real,hou2020nlh}, and learning through obstructions \cite{liu2020learning}, 
can allow text recognition schemes to address these issues.

%

\paragraph{Complex Fonts}
There are several challenging text of graphical fonts (e.g., Spencerian Script, Christmas, Subway Ticker) in the wild images that the current methods do not explicitly handle (see Fig. \ref{fig:qual_fail}(e)). 
Recognizing text of complex fonts in the wild images emphasizes on designing schemes that are able to recognize different fonts by improving the feature extraction step of these schemes or using style transfer techniques \cite{gomez2019selective,karras2019analyzing} for learning the mapping from one font to another.

\paragraph{Special Characters}
In addition to alphanumeric characters, special characters (e.g., the \$, /, -, !, :, @ and \# characters in Fig. \ref{fig:qual_fail}(f)) are also abundant in the wild images, however existing text recognition methods \cite{baek2019STR,Liao2018STR_CAFCN,zhao2017_PSPNet,shi2016robust,shi2016end} have excluded them during training and testing. 
Therefore, these pretrained models suffer from the inability to recognize special characters. 
Recently, CLOVA in \cite{baek2019craft} has shown that training the models on special characters improves the recognition accuracy, which suggests further study in how to incorporate special characters in both training and evaluation of text recognition models.

\subsubsection{Discussion}

In this section, we conduct empirical investigation for the performance of the considered recognition methods \cite{baek2019STR,Liao2018STR_CAFCN,zhao2017_PSPNet,shi2016robust,shi2016end} using ICDAR13, ICDAR15 and COCO-Text datasets, and under various word lengths and aspect ratios. In addition, we compare the recognition speed for these methods.

\begin{figure*}[th] 
    \centering
     \setlength{\tabcolsep}{2 pt}
     \scalebox{1}{
    \begin{tabular}{ccc}
    \resizebox{0.33\linewidth}{!}{\includegraphics*{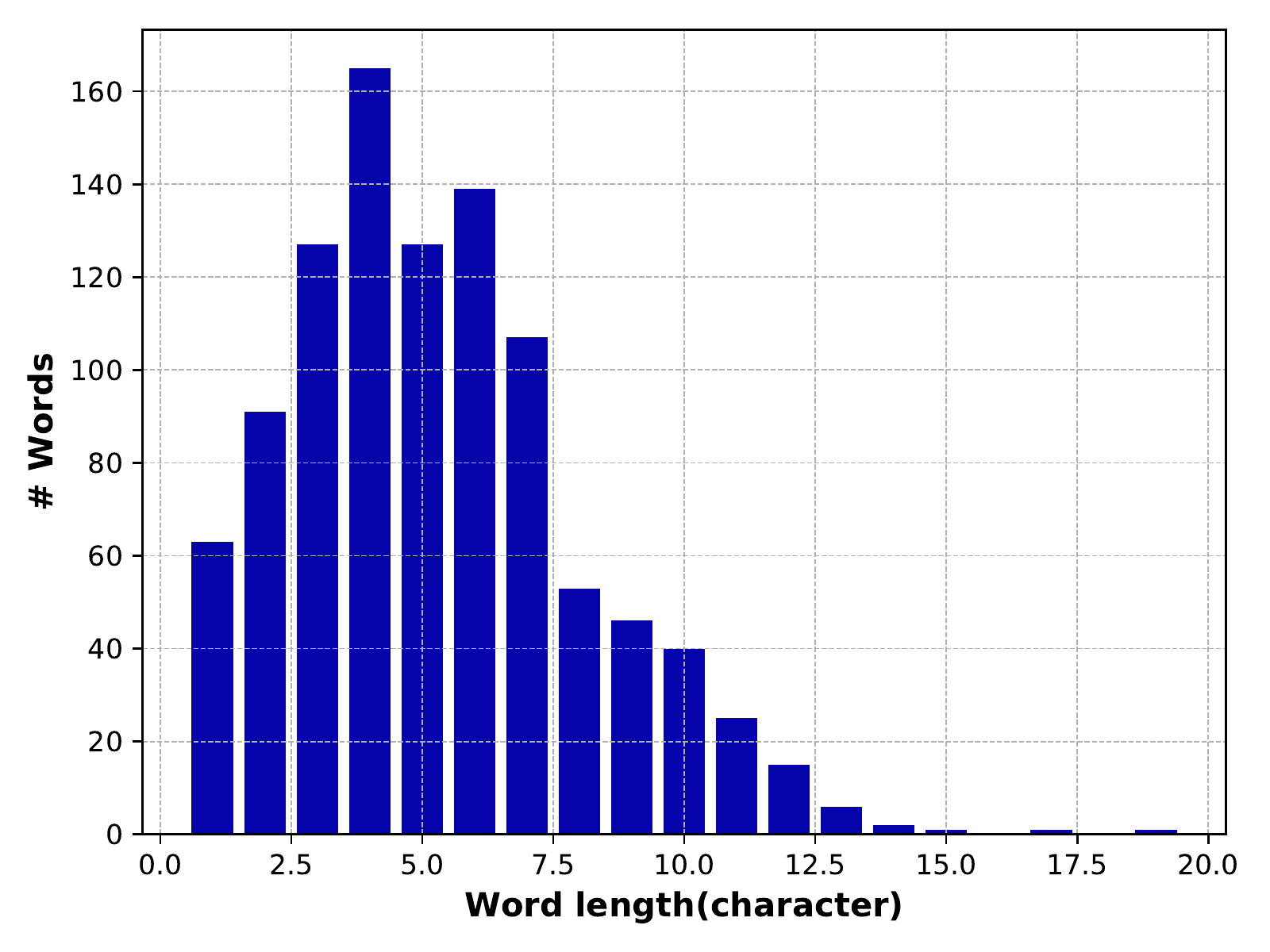}}&  
    \resizebox{0.33\linewidth}{!}{\includegraphics*{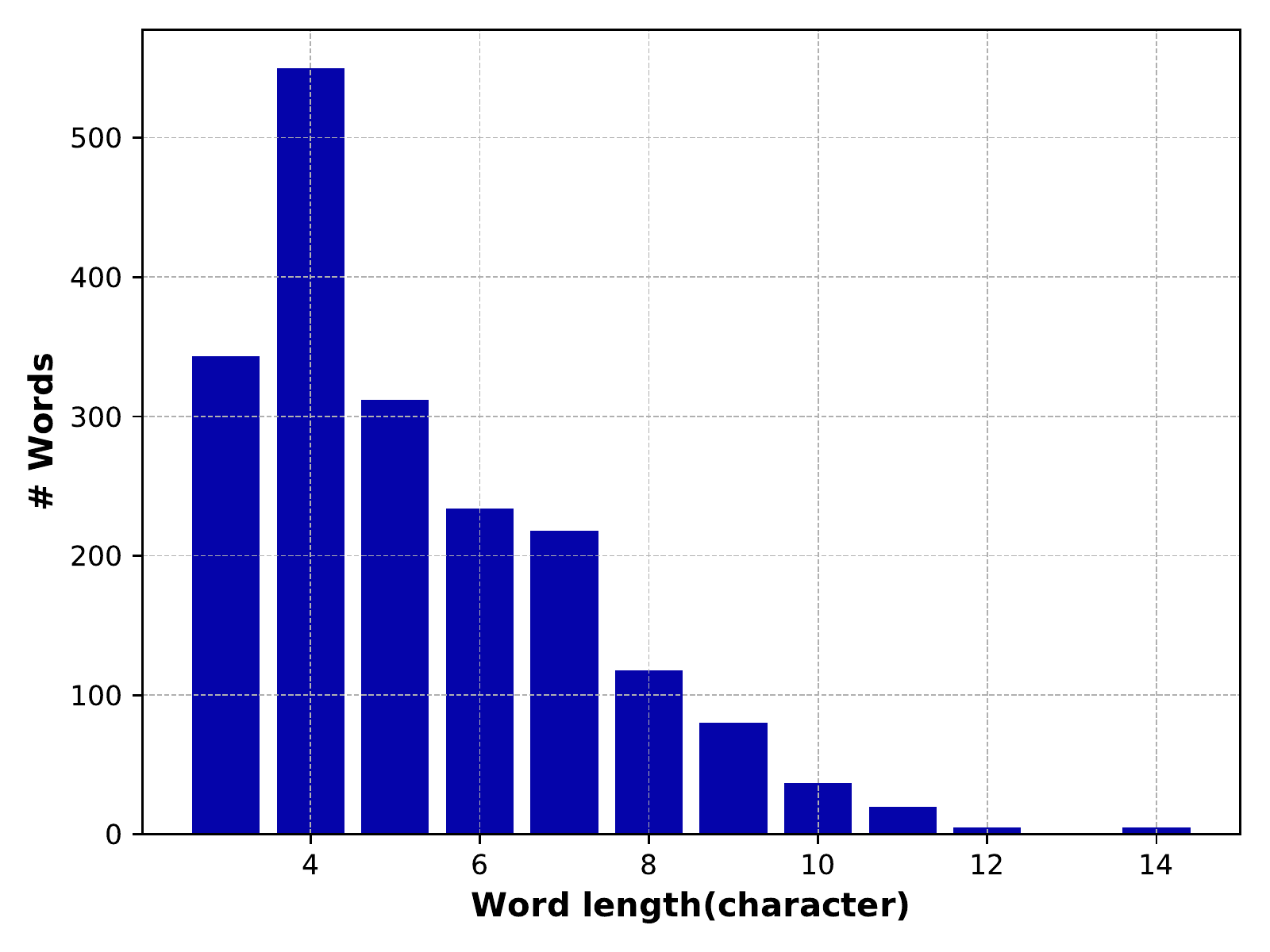}}&
    \resizebox{0.33\linewidth}{!}{\includegraphics*{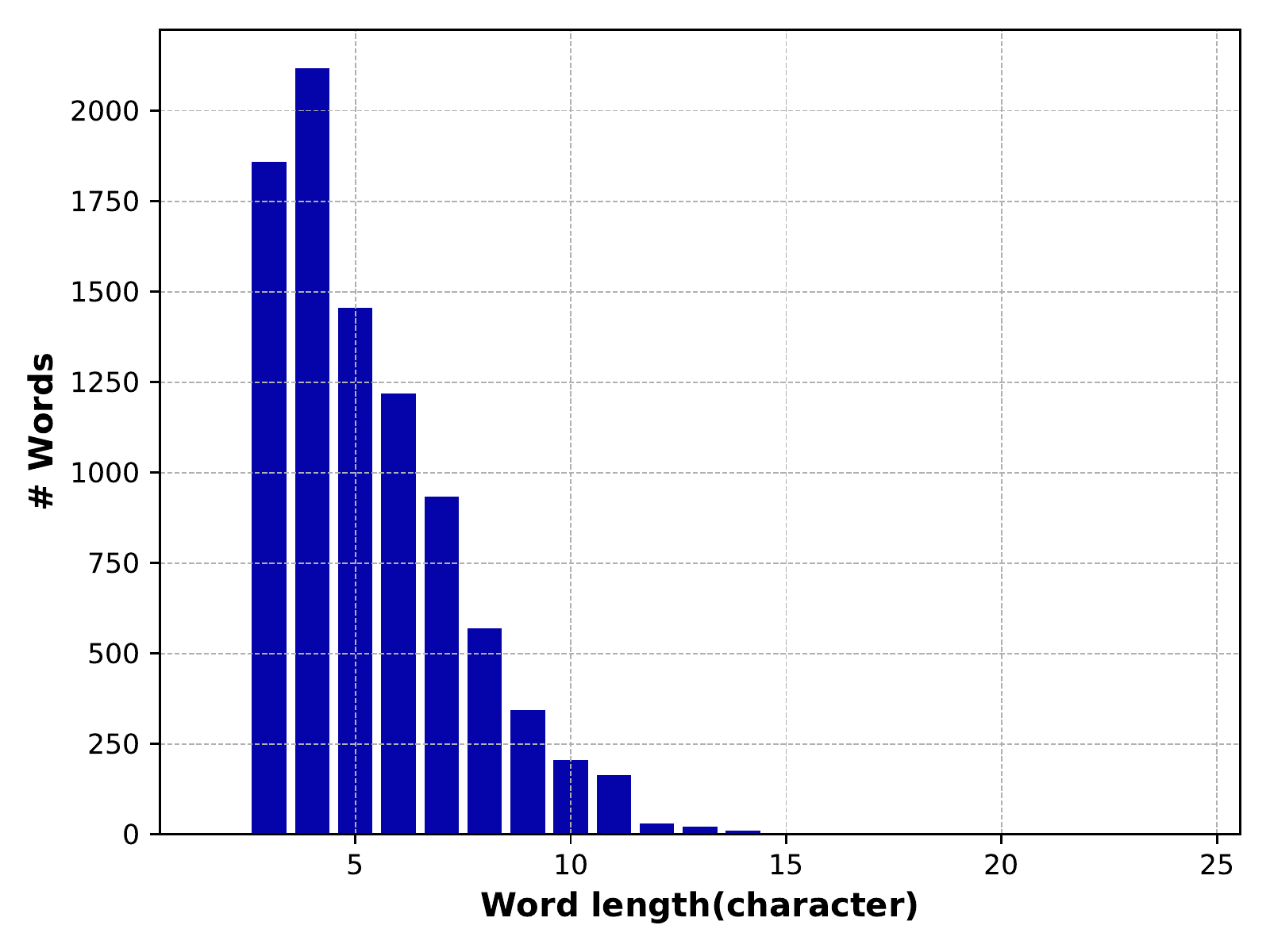}}\\
    (a) ICDAR13 & (b) ICDAR15 &  (c) COCO-Text   
    \end{tabular}}
    \caption{Statistics of word length in characters computed on (a) ICDAR13, (b) ICDAR15 and (c) COCO-Text recognition datasets.}
    \label{fig:wordLength_stat}
\end{figure*}

\begin{figure*}[th] 
    \centering
     \setlength{\tabcolsep}{2 pt}
    \begin{tabular}{ccc}
    \resizebox{0.33\linewidth}{!}{\includegraphics*{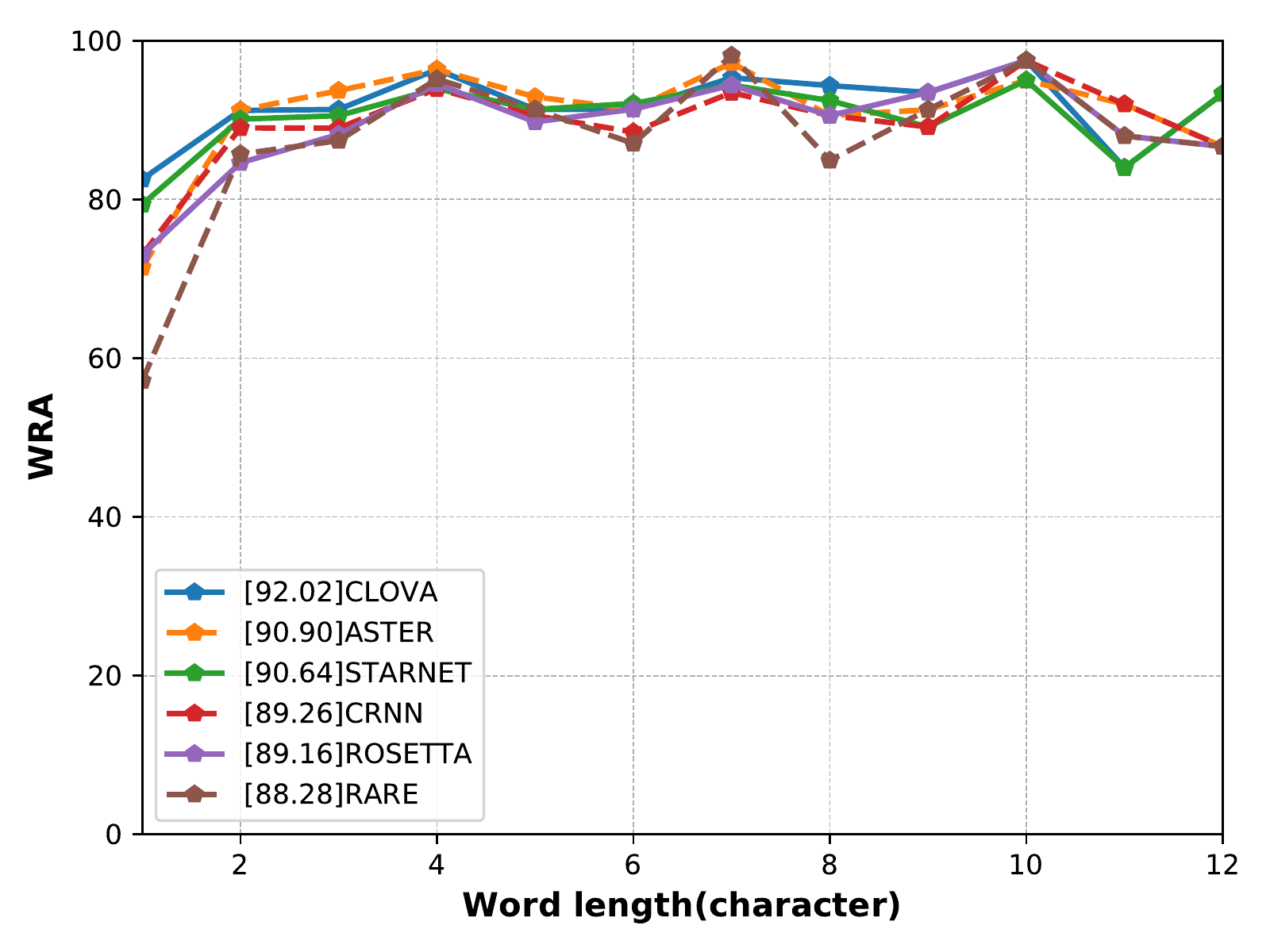}}&  
    \resizebox{0.33\linewidth}{!}{\includegraphics*{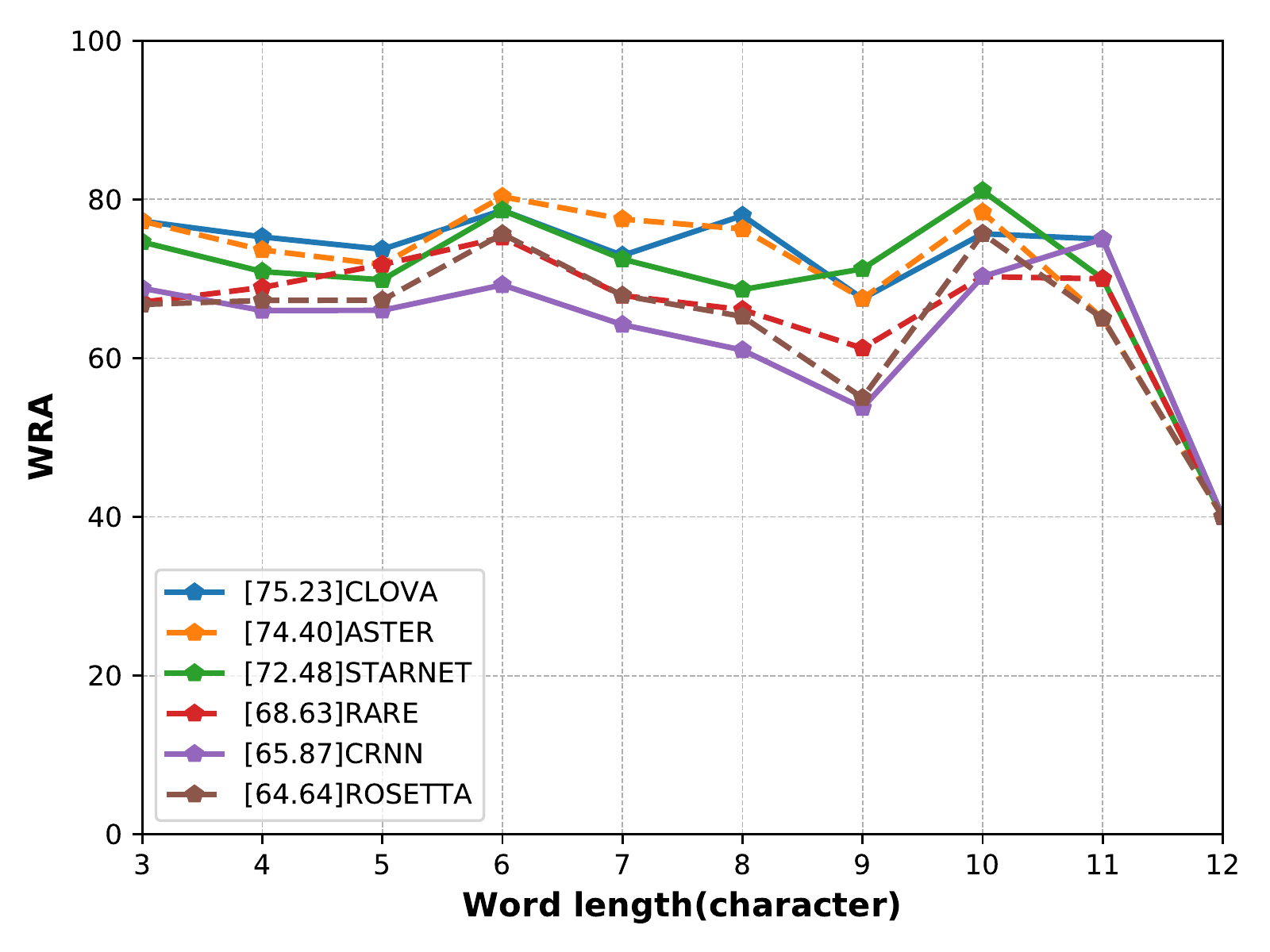}}&
    \resizebox{0.33\linewidth}{!}{\includegraphics*{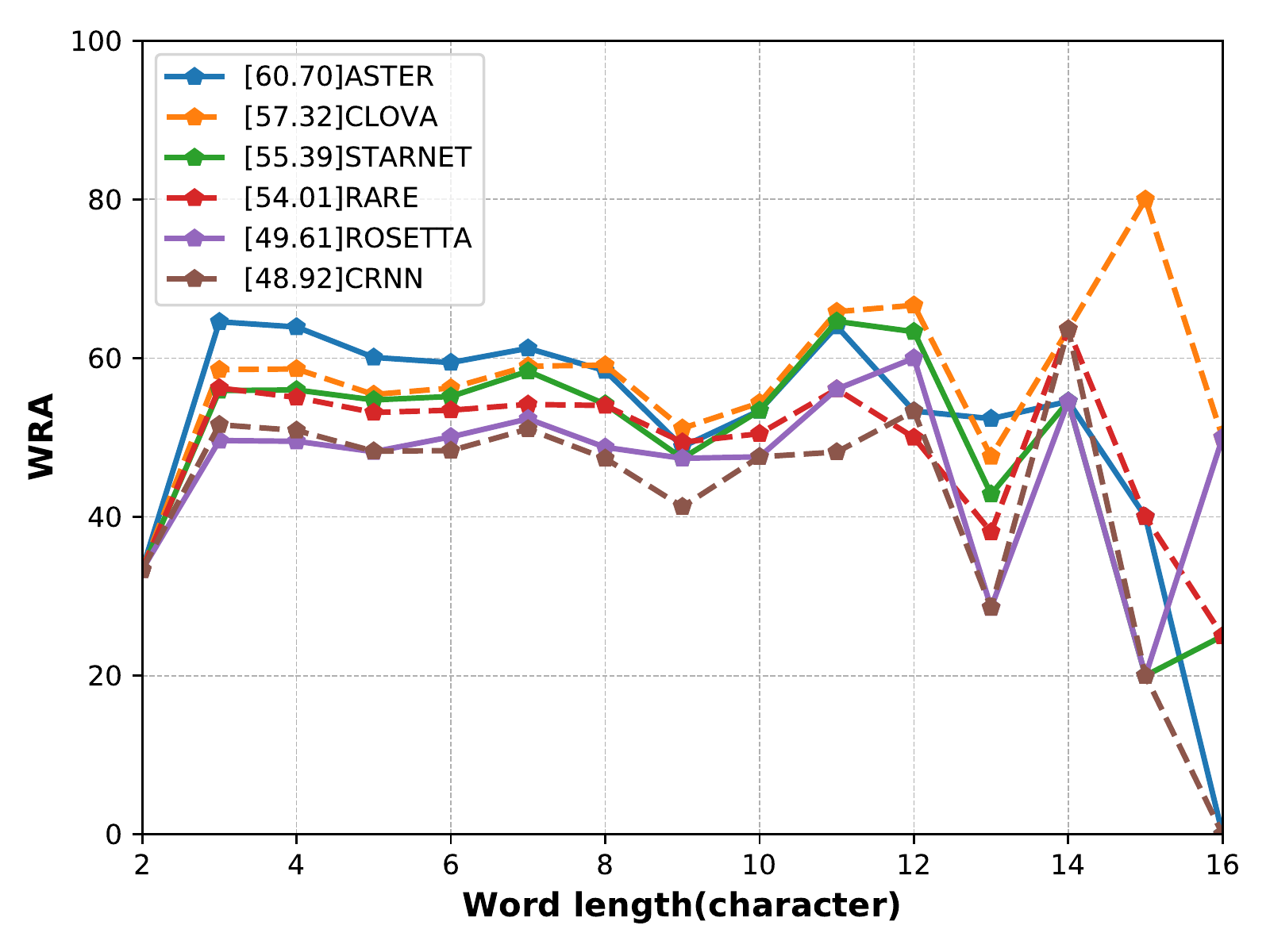}}\\
    (a) ICDAR13 & (b) ICDAR15 &  (c) COCO-Text   
    \end{tabular}
    \caption{Evaluation of the average WRA at different word length for ASTER \cite{shi2018aster}, CLOVA \cite{baek2019STR}, STARNET \cite{STARNet_2016}, RARE \cite{shi2016robust}, CRNN \cite{shi2016end}, and ROSETTA \cite{rosetta2018} computed on (a) ICDAR13, (b) ICDAR15, and (c) COCO-Text recognition datasets.}
    \label{fig:word_length_rec}
\end{figure*}

\paragraph{Word-length}
In this analysis, we first obtained the number of images with different word lengths for ICDAR13, ICDAR15 and COCO-Text datasets as shown in Fig. \ref{fig:wordLength_stat}.
As can be seen from Fig. \ref{fig:wordLength_stat}, most of the words have a word length between 2 to 7 characters, so we will focus this analysis on short and intermediate words. 
Fig. \ref{fig:word_length_rec} illustrates the accuracy of the text recognition methods at different word-lengths for ICDAR13, ICDAR15 and COCO-Text datasets. 

%
On ICDAR13 dataset, shown in Fig. \ref{fig:word_length_rec}(a), all the methods offered consistent accuracy values for words with length larger than 2 characters. 
This is because all the text instances in this dataset are horizontal and of high resolution. However, for  words with 2 characters, RARE offered the worst accuracy ($\sim$58\%), while CLOVA offered the best accuracy ($\sim$ 83\%).
On ICDAR15 dataset, the recognition accuracies of the methods follow a consistent trend similar to ICDAR13 \cite{karatzas2013icdar}. However, the recognition performance is generally lower than that obtained on ICDAR13, because this dataset has more blurry, low resolution, and rotated images than ICDAR13. 
%
On COCO-Text dataset, 
ASTER and CLOVA achieved the best, and the second-best ac curacies,
and overall, except some fluctuations at word length more than 12 characters, all the methods followed a similar trend.



\paragraph{Aspect-ratio}
In this experiment, we study the accuracy achieved by the studied methods on words with different aspect ratio (height/width). As can be seen from Fig. \ref{fig:ratio_stat} most of the word images in the considered datasets are of aspect ratios between 0.3 and 0.6.
Fig. \ref{fig:word_ratio_rec} shows the WRA values of the studied methods \cite{shi2016end,STARNet_2016,shi2016robust,rosetta2018,shi2018aster,baek2019STR} versus the word aspect ratio computed on ICDAR13, ICDAR15 and COCO-Text datasets. 
%
From this figure, for images with aspect-ratio $< 0.3$ the studied methods offer low WRA values on the three considered datasets. 
The main reason for this is that this range mostly include text of long words that face an assessment challenge of correctly predicting every character within a given word. 
For images within $0.3 \leq $ aspect-ratio $\leq 0.5$, which include 
images of medium word length (4-9 characters per word), it can be seen from Fig. \ref{fig:word_ratio_rec} that the highest WRA values are offered by the studied methods.
It can be observed from Fig. \ref{fig:word_ratio_rec} also that when evaluating the target state-of-the-art methods on images with aspect ratio $\geq 0.6$, all the methods have experienced a decline in the WRA value.
This is due to those images are mostly of words of short length and of low resolution. 

\begin{figure}
    \centering
    \includegraphics[width=\linewidth]{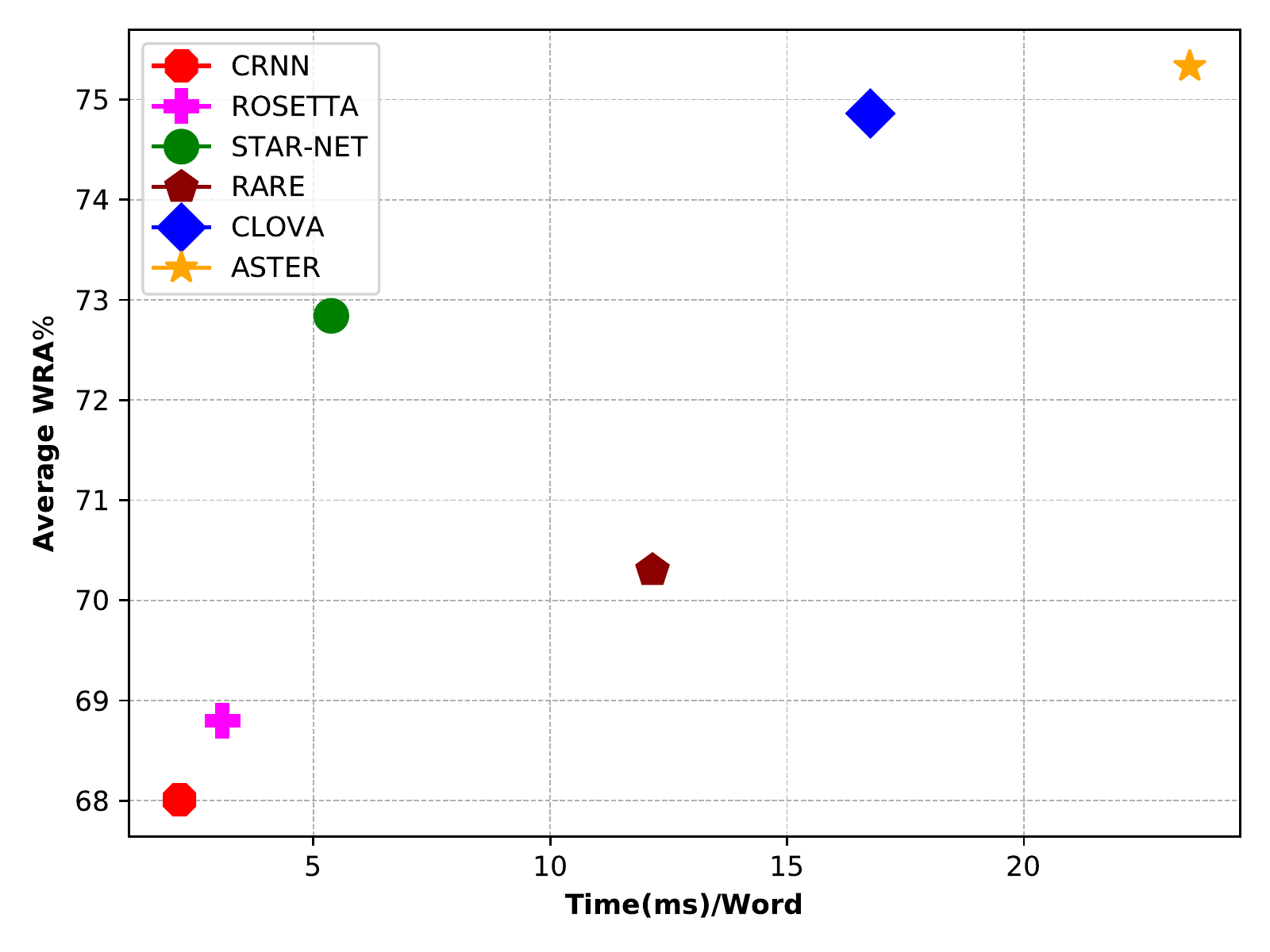}
    \caption{Average WRA versus average recognition time per word in milliseconds computed on ICDAR13 \cite{karatzas2013icdar}, ICDAR15 \cite{karatzas2015icdar}, and COCO-Text \cite{veit2016coco} datasets.}
    \label{fig:fps_rec}
\end{figure}

\paragraph{Recognition Time}
We also conducted an investigation to compare the recognition time versus the WRA for the considered state-of-the-art scene text recognition models \cite{shi2016end,STARNet_2016,shi2016robust,rosetta2018,shi2018aster,baek2019STR}.
Fig. \ref{fig:fps_rec} shows the inference time per word-image in milliseconds, when the test batch size is one, where the inference time could be reduced with using larger batch size. 
The fastest and the slowest methods are CRNN \cite{shi2016end} and ASTER \cite{shi2018aster} that achieve time/word of $\sim2.17$ msec and $\sim23.26$ msec, respectively, which illustrate the big gap in the computational requirements of these models.    
Although attention-based methods, ASTER \cite{shi2018aster} and CLOVA \cite{baek2019STR}, provide higher word recognition accuracy (WRA) than that of CTC-based methods, CRNN \cite{shi2016end}, ROSETTA \cite{rosetta2018} and STAR-Net \cite{STARNet_2016}, however, they are much slower compared to CTC-based methods. 
This slower speed of attention-based methods come back to the deeper feature extractor and rectification modules utilized in their architectures.
{\begin{figure*}[th]
    \centering
     \setlength{\tabcolsep}{2 pt}
          \scalebox{1}{
    \begin{tabular}{ccc}
    \resizebox{0.33\linewidth}{!}{\includegraphics*{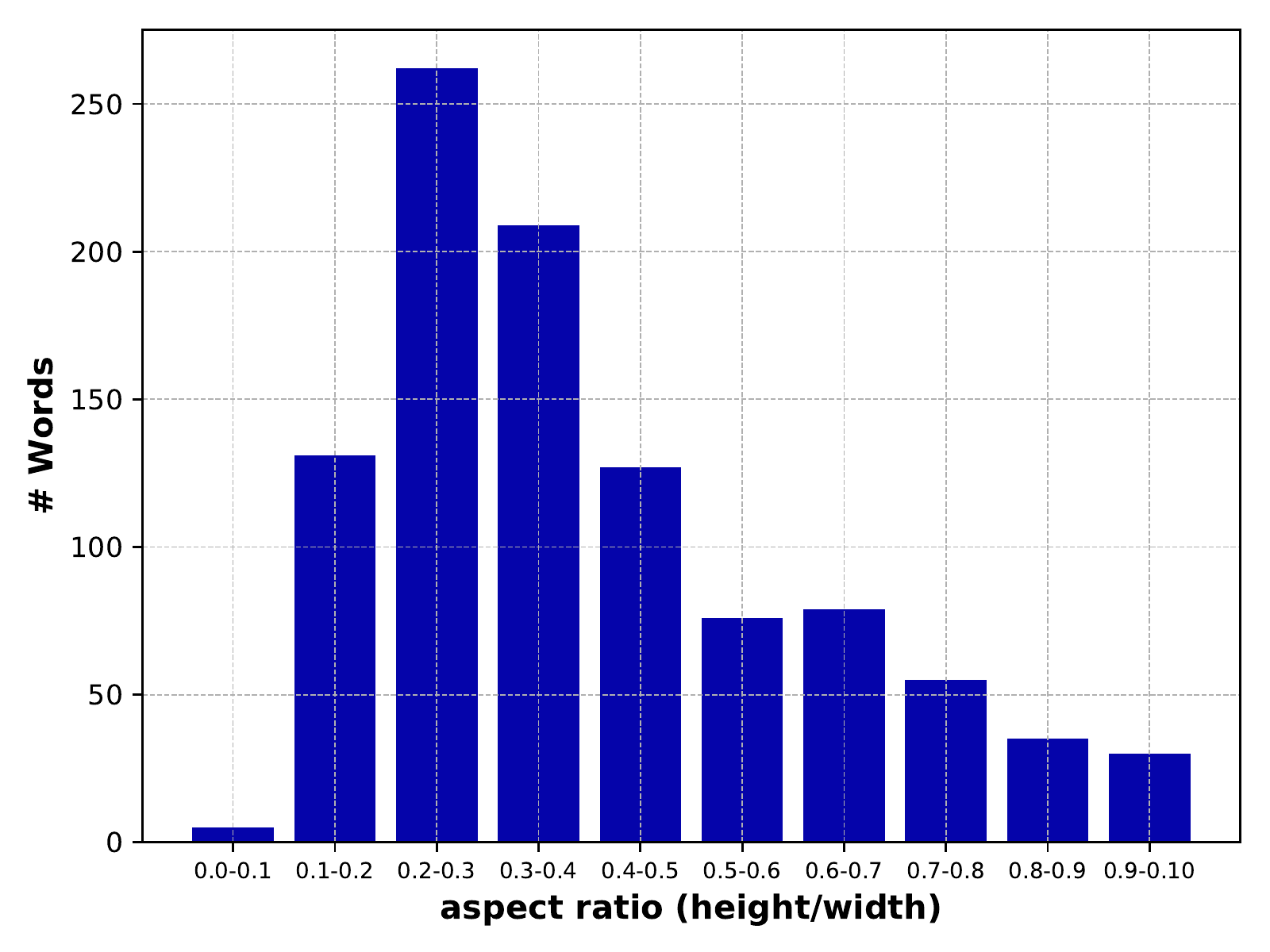}}&  
    \resizebox{0.33\linewidth}{!}{\includegraphics*{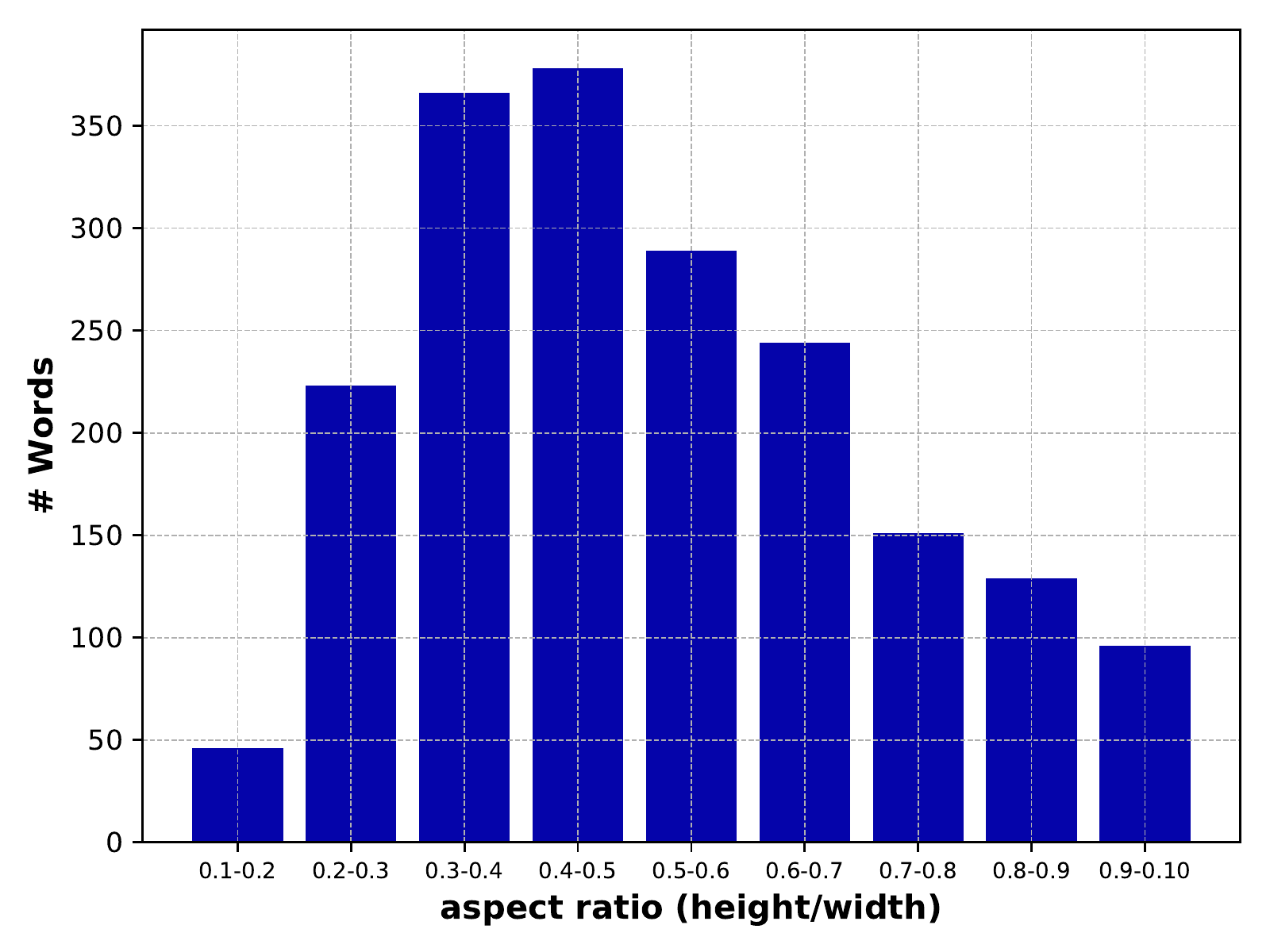}}&
    \resizebox{0.33\linewidth}{!}{\includegraphics*{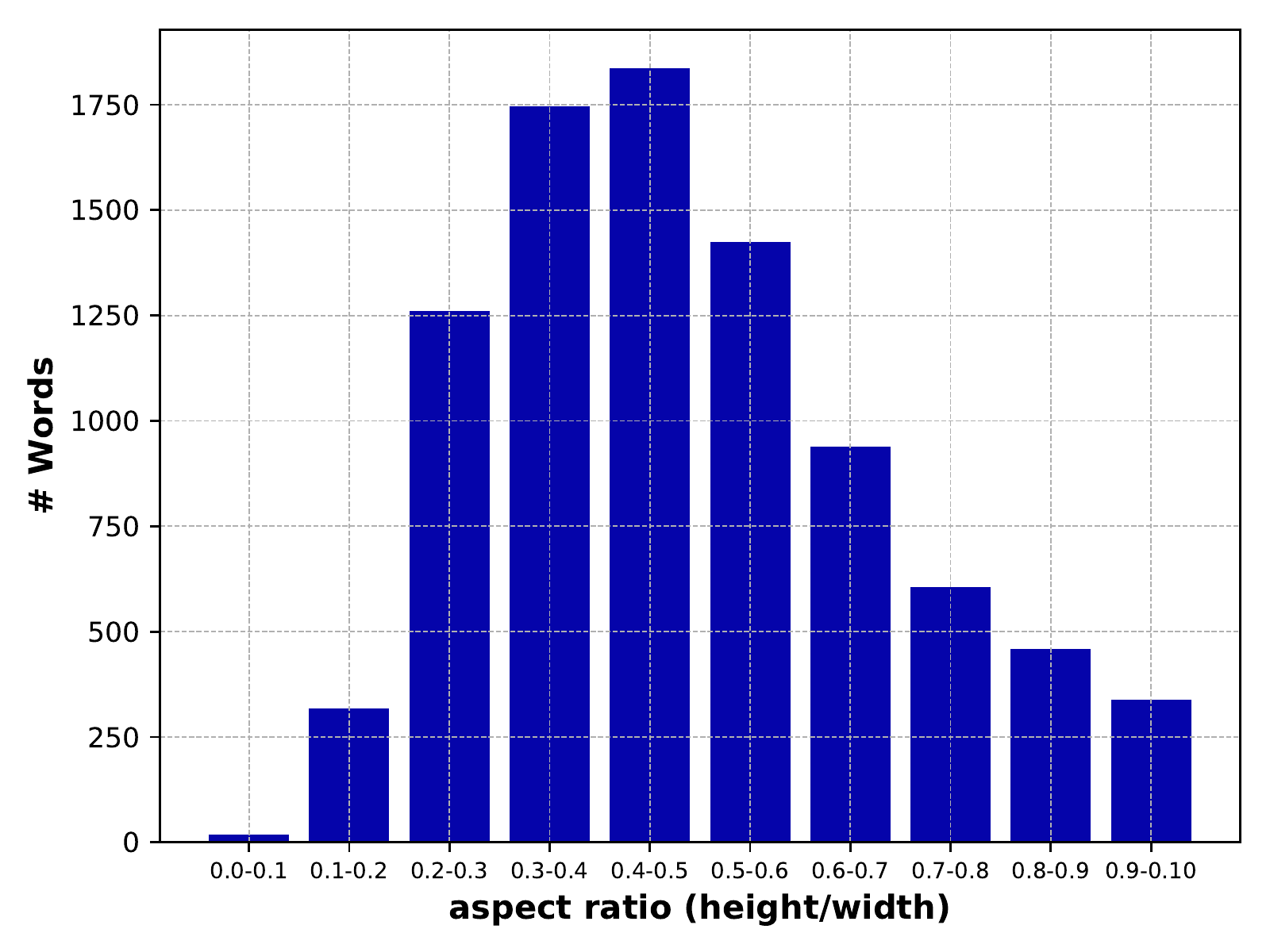}}\\
    (a) ICDAR13 & (b) ICDAR15 &  (c) COCO-Text   
    \end{tabular}}
    \caption{Statistics of word aspect-ratios computed on (a) ICDAR13, (b) ICDAR15 and (c) COCO-Text recognition datasets.}
    \label{fig:ratio_stat}
\end{figure*}
\begin{figure*}[th]
    \centering
     \setlength{\tabcolsep}{2 pt}
    \begin{tabular}{ccc}
    \resizebox{0.33\linewidth}{!}{\includegraphics*{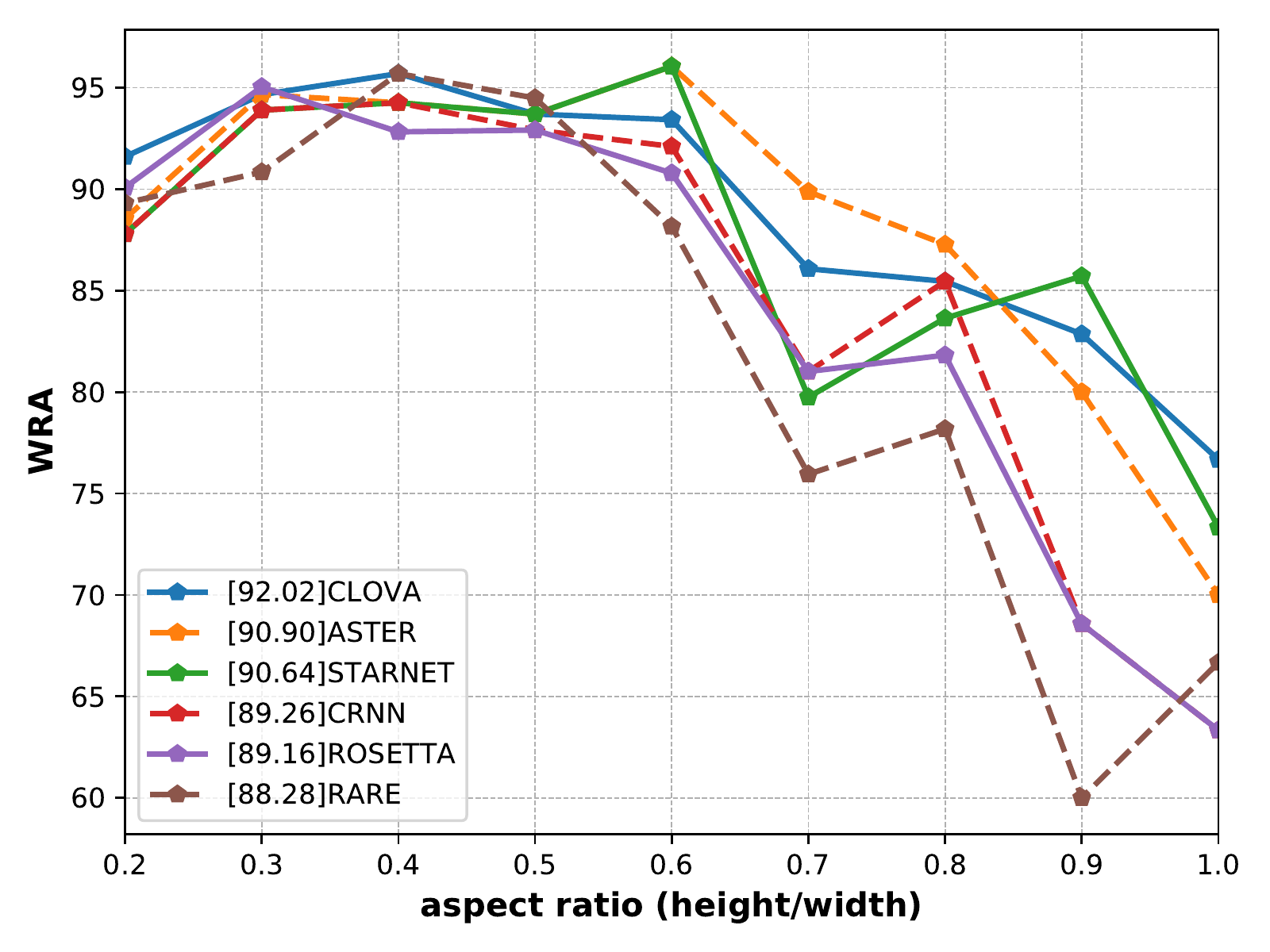}}&  
    \resizebox{0.33\linewidth}{!}{\includegraphics*{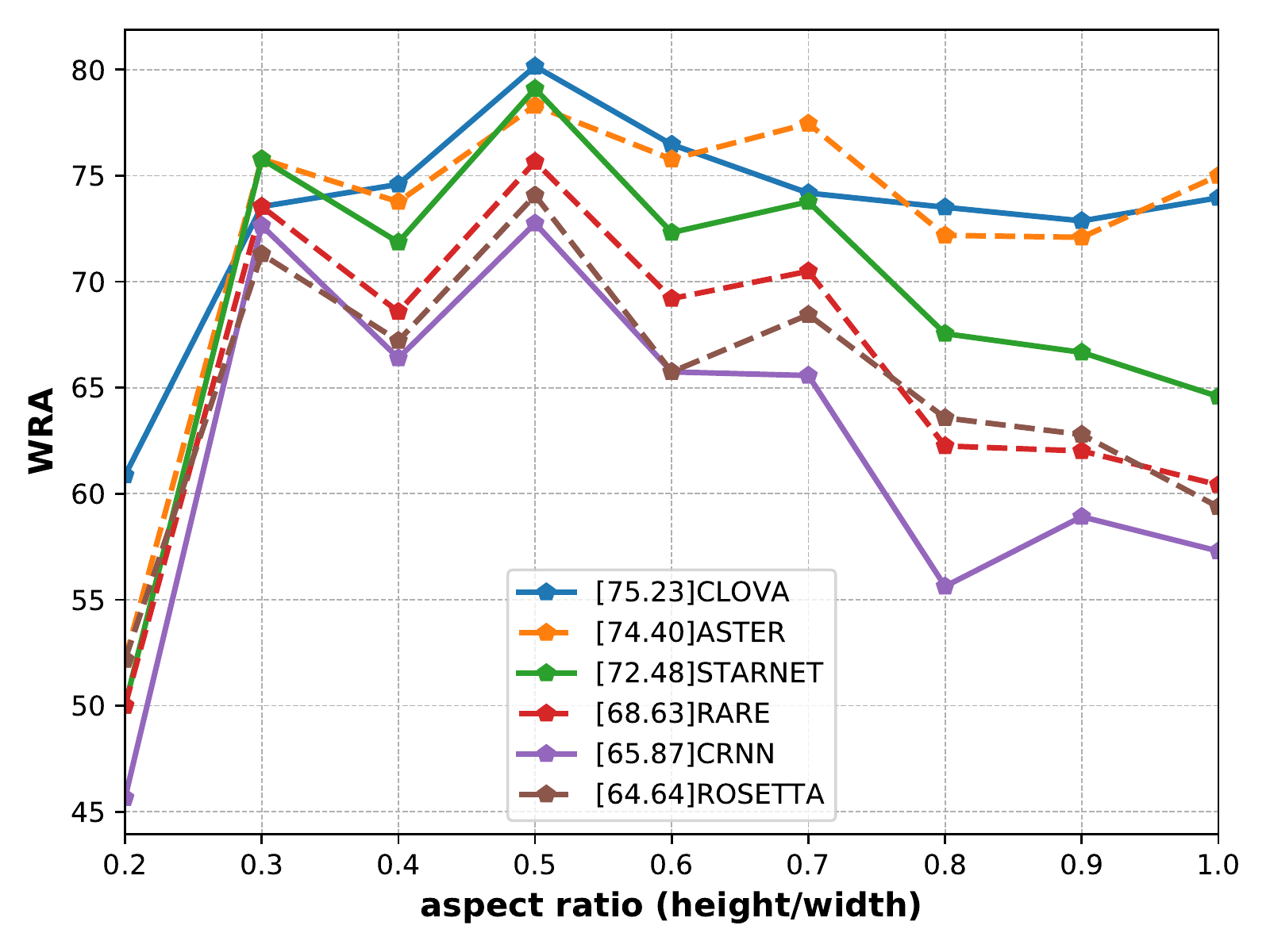}}&
    \resizebox{0.33\linewidth}{!}{\includegraphics*{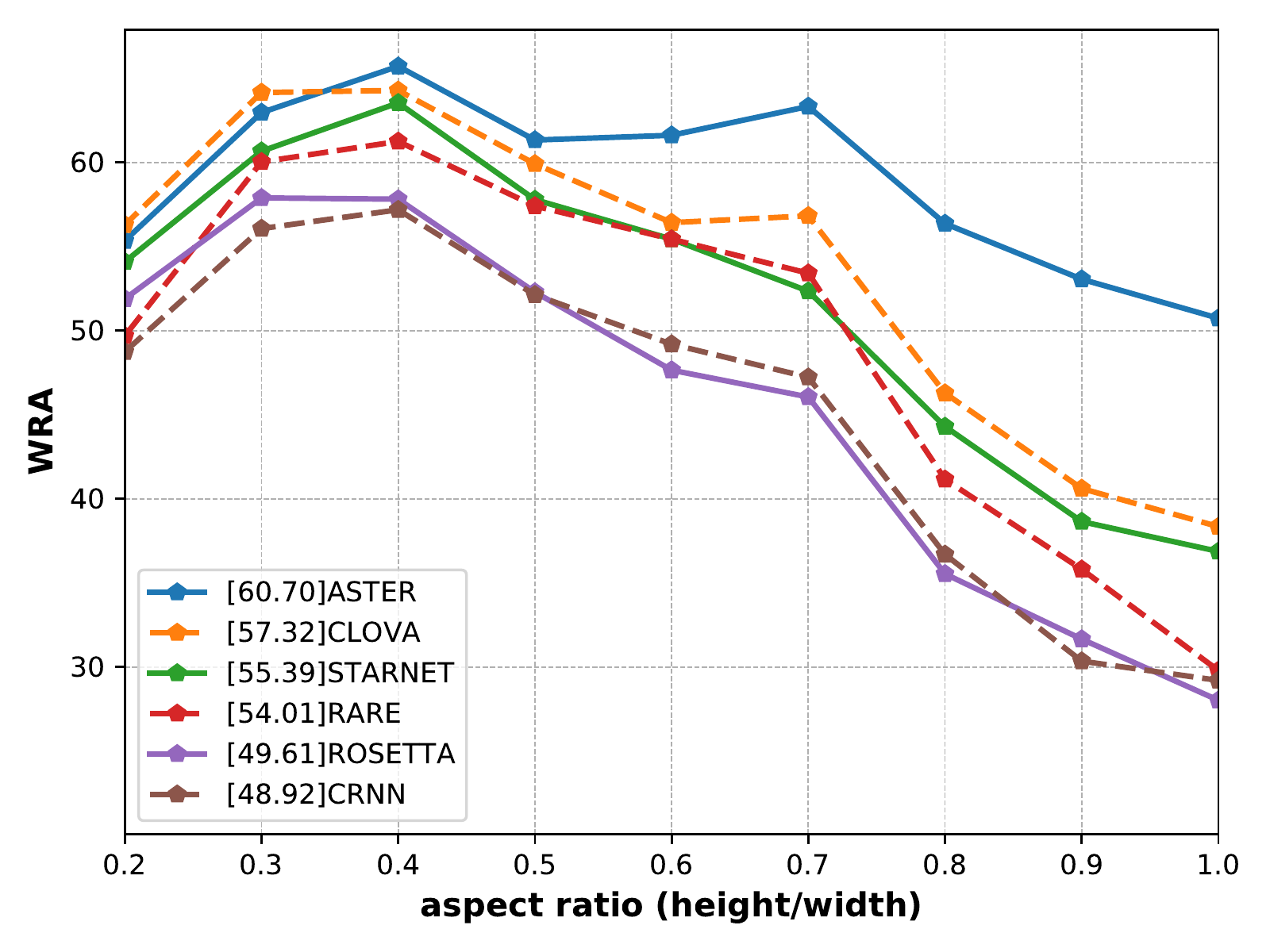}}\\
    (a) ICDAR13 & (b) ICDAR15 &  (c) COCO-Text   
    \end{tabular}
    \caption{Evaluation of average WRA at various word aspect-ratios for ASTER \cite{shi2018aster}, CLOVA \cite{baek2019STR}, STARNET \cite{STARNet_2016}, RARE \cite{shi2016robust}, CRNN \cite{shi2016end}, and ROSETTA \cite{rosetta2018} using (a) ICDAR13, (b) ICDAR15 and (c) COCO-Text recognition datasets.}
    \label{fig:word_ratio_rec}
\end{figure*}}

\subsection{Open Investigations for Scene Text Detection and Recognition}
Following the recent development in object detection and recognition problems, deep learning scene text detection and recognition frameworks have progressed rapidly such that the reported H-mean performance and recognition accuracy are about 80\% to 95\% for several benchmark datasets. 
However, as we discussed in the previous sections, there are still many open issues for future works. 

\subsubsection{Training Datasets}
Although the role of synthetic datasets can not be ignored in the training of recognition algorithms, detection methods still require more real-world datasets to fine-tune.
Therefore, using generative adversarial network \cite{GAN_goodfellow2014} based methods or 3D proposal based \cite{chen20153d} models that produce more realistic text images can be a better way of generating synthetic datasets for training text detectors.

\subsubsection{Richer Annotations}
For both detection and recognition, due to the annotation shortcomings for quantifying challenges in the wild images, existing methods have not explicitly evaluated on tackling such challenges. 
Therefore, future annotations of benchmark datasets should be supported by additional meta descriptors (e.g., orientation, illumination condition, aspect-ratio, word-length, font type, etc.) such that methods can be evaluated against those challenges, and thus it will help future researchers to design more robust and generalized algorithms.

\subsubsection{Novel Feature Extractors}
It is essential to have a better understanding of what type of features are useful for constructing improved text detection and recognition models. 
For example, a ResNet \cite{ResNet_He2015L} with higher number of layers will give better results \cite{baek2019craft,zhao2019object}, while it is not clear yet what can be an efficient feature extractor that allows differentiating text from other objects, and recognizing the various text characters as well. 
Therefore, a more thorough study of the dependents on different feature extraction architecture as the backbone in both detection and recognition is required.
\subsubsection{Occlusion Handling}
So far, existing methods in scene text recognition rely on the visibility of the target characters in images, however, text affected by heavy occlusion may significantly undermine the performance of these methods.
Designing a text recognition scheme based on a strong natural language processing model like, BERT \cite{BERT_devlin2018}, can help in predicting occluded characters in a given text.



\subsubsection{Complex Fonts and Special Characters}
Images in the wild can include text with a wide variety of complex fonts, such as calligraphic fonts, and/or colors. 
Overcoming those variabilities can be possible by generating images with more real-world like text using style transfer learning techniques \cite{gomez2019selective,karras2019analyzing} or improving the backbone of the feature extraction methods \cite{resnext_xie2017,capsnet_sabour2017}.
As we mentioned in Section \ref{sec:RecQualitativeResults}, special characters (e.g., \$, /, -, !,:, @ and \#) are also abundant in the wild images, but the research community has been ignoring them during training, which leads to incorrect recognition for those characters. 
Therefore, including images of special characters in training future scene text detection/recognition methods as well, will help in evaluating these models on detecting/recognizing these characters.

\section{Conclusions and Recommended Future Work}
\label{sec:DiscussionsandConlusions}

It has been noticed that in recent scene text detection and recognition surveys, despite the performance of the analyzed deep learning-based methods have been compared on multiple datasets, the reported results have been used for evaluation, which make the direct comparison among these methods difficult. This is due to the lack of a common experimental settings, ground-truth and/ or evaluation methodology.
In this survey, we have first presented a detailed review on the recent advancement in scene text detection and recognition fields with focus on deep learning based techniques and architectures. 
Next, we have conducted extensive experiments on challenging benchmark datasets for comparing the performance of a selected number of pre-trained scene text detection and recognition methods, which represent the recent state-of-the-art approaches, under adverse situations. 
More specifically, when evaluating the selected scene text detection schemes on ICDAR13, ICDAR15 and COCO-Text datasets we have noticed the following:
\begin{itemize}
    \item Segmentation-based methods, such as PixelLink, PSENET, and PAN, are more robust in predicting the location of irregular text. 
    \item  Hybrid regression and segmentation based methods, like PMTD, achieve the best H-mean values on all the three datasets, as they are able to handle better multi-oriented text.
    \item  Methods that detect text at the character level, as in CRAFT, can perform better in detecting irregular shape text. 
    \item In images with text affected by more than one challenge, all the studied methods performed weakly.
\end{itemize}

With respect to evaluating scene text recognition methods on challenging benchmark datasets, we have noticed the following:
\begin{itemize}
\item Scene text recognition methods that only use synthetic scene images for training have been able to recognize text in real-world images without fine-tuning their models.
    \item 
    In general, attention-based methods, as in ASTER and CLOVA, that benefit from a deep backbone for feature extraction and transformation network for rectification have performed better than that of CTC-based methods, as in CRNN, STARNET, and ROSETTA. 
\end{itemize}

It has been shown that there are several unsolved challenges for detecting or recognizing text in the wild images, such as in-plane-rotation, multi-oriented and multi-resolution text, perspective distortion, shadow and illumination reflection, image blurriness, partial occlusion, complex fonts and special characters, that we have discussed throughout this survey and which open more potential future research directions.
This study also highlights the importance of having more descriptive annotations for text instances to allow future detectors to be trained and evaluated against more challenging conditions.



\begin{acknowledgements}
The authors would like to thank the Ontario Centres of Excellence (OCE) - Voucher for Innovation and Productivity II (VIP II) - Canada program, and ATS Automation Tooling Systems Inc., Cambridge, ON Canada, for supporting this research work.
\end{acknowledgements}

%
%

{\small
\bibliographystyle{IEEEtran}
\bibliography{myref2}
}



\end{document}